\RequirePackage{etoolbox}
\documentclass[a4paper,fleqn]{cas-dc}
\usepackage{enumitem}
\usepackage[normalem]{ulem} 
\usepackage{soul} 
\usepackage{array}
\newcolumntype{?}{!{\vrule width 1pt}}
\usepackage{gensymb}
\usepackage{caption}
\usepackage[T1]{fontenc}
\usepackage{mathtools}
\usepackage{amsmath}
\usepackage{nccmath}
\usepackage{ltablex}
\usepackage{amssymb}
\usepackage{textcomp} 
\usepackage[nomath]{lmodern}
\usepackage{graphicx}
\graphicspath{{images/}}
\usepackage[sort&compress]{natbib}
\usepackage{flushend}
\usepackage{longtable}
\usepackage[flushleft]{threeparttable} 
\usepackage{tablefootnote}
\usepackage{cuted}
\usepackage{booktabs}
\usepackage{cellspace}
\usepackage{tabularx}
\usepackage{tabularray}
\usepackage{subfig}
\usepackage{multirow}
\NewColumnType{M}[1]{Q[m,c,#1]}  
\usepackage{makecell}
\usepackage{pifont}
\usepackage{microtype}
\usepackage[titletoc,title]{appendix}
\usepackage{imakeidx}
\usepackage{xcolor}
\def\tsc#1{\csdef{#1}{\textsc{\lowercase{#1}}\xspace}}
\tsc{WGM}
\tsc{QE}
\tsc{EP}
\tsc{PMS}
\tsc{BEC}
\tsc{DE}

\newcommand{\tabitem}{~~\llap{\textbullet}~~}

\usepackage{lipsum}

\usepackage{framed} 

\usepackage{multicol} 

\usepackage[intoc]{nomencl} 

\makenomenclature

\makeindex

\renewcommand*\nompreamble{\begin{multicols}{2}}

\renewcommand*\nompostamble{\end{multicols}}

\renewcommand\nomgroup[1]{%
  \item[\itshape
  \ifstrequal{#1}{A}{\textrm{Abbreviations}}{%
  \ifstrequal{#1}{B}{\textrm{Superscripts}}{}
  \ifstrequal{#1}{C}{\textrm{Subscripts}}{%
  \ifstrequal{#1}{D}{\textrm{Sets}}{}}%
  \ifstrequal{#1}{S}{\textrm{Symbols}}{}}%
]}

\makeatletter
\let\oldlt\longtable
\let\endoldlt\endlongtable
\def\longtable{\@ifnextchar[\longtable@i \longtable@ii}
\def\longtable@i[#1]{\begin{figure}[t]
\onecolumn
\begin{minipage}{0.5\textwidth}
\oldlt[#1]
}
\def\longtable@ii{\begin{figure}[t]
\onecolumn
\begin{minipage}{0.5\textwidth}
\oldlt
}
\def\endlongtable{\endoldlt
\end{minipage}
\twocolumn
\end{figure}}
\makeatother

\usepackage{subfig}

\usepackage{float}

\usepackage{fltrace}

\usepackage{titlesec}

\titleformat*{\section}{\fontsize{10}{12}\bfseries}
\titleformat*{\subsection}{\fontsize{10}{12}\itshape}
\titleformat*{\subsubsection}{\fontsize{10}{12}\itshape}

\usepackage{placeins}

\begin{document}

\captionsetup[figure]{labelfont={bf},textfont={rm},labelformat={default},labelsep=period,name={Fig.}}

\let\WriteBookmarks\relax
\def\floatpagepagefraction{1}
\def\textpagefraction{.001}
\shorttitle{\textrm{Comparative review of hybrid forecasting models for short-term prediction of building thermal load}}
\shortauthors{N. A. Efkarpidis et~al.}

\title [mode = title]{Comparative review of hybrid forecasting models for short-term prediction of building thermal load}                      

\author[1]{Nikolaos A. Efkarpidis}[type=editor,
orcid=0000-0001-8487-6824]
\fnref{fn1}
\credit{Conceptualization, Methodology, Software, Formal analysis, Investigation, Writing - original draft, Writing - review \& editing}

\address[1]{VIVAVIS Schweiz AG - 5405 Baden, Switzerland}

\author[2]{Despoina Kothona}[type=editor,
orcid=0000-0002-1759-6061]
\cormark[1]
\credit{Conceptualization, Methodology, Software, Formal analysis, Investigation, Writing - original draft, Writing - review \& editing}

\address[2]{Department of Electrical and Computer Engineering, University of Western Macedonia, 50100 Kozani, Greece}

\author[2]{Georgios C. Christoforidis}[type=editor,
orcid=0000-0001-5595-8409]
\credit{Resources, Writing - review \& editing, Supervision}

\cortext[cor1]{Corresponding author}
\fntext[fn1]{Tel.: +40 766070656\\
E-mail address: nikolaos.efkarpidis@vivavis.ch}

\begin{abstract}
In this paper, a comparative review of different hybrid models for short-term forecasting of building thermal demand is carried out. Particularly, the assessment tackles the comparison of data-driven models enhanced with other state-of-the-art techniques. At the first step, the existing techniques reported in the literature are analysed. It is concluded that Metaheuristics or a data-driven model are used to identify the parameters of the basic model. The qualitative evaluation includes for each method the input and output features, main advantages and drawbacks. At the second step, an existing dataset of historical thermal demand from Scottish households, as well as historical weather forecasts are utilized to assess additionally the performance of existing hybrid methods. From the assessment of 13 hybrid methods, the Empirical Modal Decomposition - long short-term memory - Markov (EMD-LSTM-Markov) model can predict with the highest accuracy the day-ahead power pattern of heating and domestic hot water (DHW) demands. Though local power peaks are also accurately predicted, high power swells and spikes are underestimated. Other methods, such as Support Vector Machine - Simulated Annealing (SVM-SA) and Random Forest - Improved Sparrow Search Algorithm - LSTM (RF-ISSA-LSTM) predict a smooth pattern of heating and DHW demand profiles with rapid changes underestimating most power peaks.
\end{abstract}


\maketitle

\section{Introduction}

\subsection{General problem and motivation}

\label{intr}

Nowadays, the increasing cost of energy and the promotion of low-carbon policies lead to the need for low-energy solutions. The building sector accounts for 40\% of the energy demand and 36\% of the greenhouse gas emission in the European Union (EU) \cite{Squire2024}. Heating Ventilation and Air Conditioning (HVAC) systems, responsible for space heating and cooling, emerge as the primary energy consumers, constituting about half of the energy consumption in the EU \cite{EU2024}. 

With a growing promotion by governmental authorities for building energy conservation and carbon emission reduction, building energy efficiency has become the main focus of both research and industrial activities. The latest energy efficiency guidelines in the building sector have also emphasized the need to reduce the building energy consumption. The decrease in energy consumption can be achieved through advanced building designs, by retrofitting the building envelope or improving the operation of HVAC system controllers \cite{Shaikh2014}. The latter activity is included in the design of suitable building energy management systems (BEMS) based on models that adequately capture the relevant thermal dynamics of a building. Particularly, BEMS technologies require accurate thermal models of building envelopes with the main goal to maintain the indoor air quality, thermal and visual comfort achieving energy savings \cite{Prívara2013}. Model-based control of buildings, such as Model predictive control (MPC), shows high dependency of the energy savings on the forecast of weather and indoor conditions, e.g., outdoor temperature and occupancy level, as well as on system constraints, e.g., HVAC power limit and thermal comfort criteria \cite{Oldewurtel2012}. 

One of the main challenges to achieve high BEMS performance for model-based control is the development of accurate dynamical thermal models, which are sufficiently adaptive to internal and external disturbances \cite{Oldewurtel2012}. In particular, the main prerequisite for optimal thermal energy management is the accurate forecast of building thermal load with a prediction horizon of up to a few days, referred to as short-term load forecasting in the literature \cite{Fux2013}. However, this task is complicated due to various factors influencing the thermal load, e.g., climate, building envelope, BEMS operation, occupant behaviour, and indoor environmental quality \cite{Oldewurtel2012}. 

\begin{figure*}
  \includegraphics[width=\textwidth]{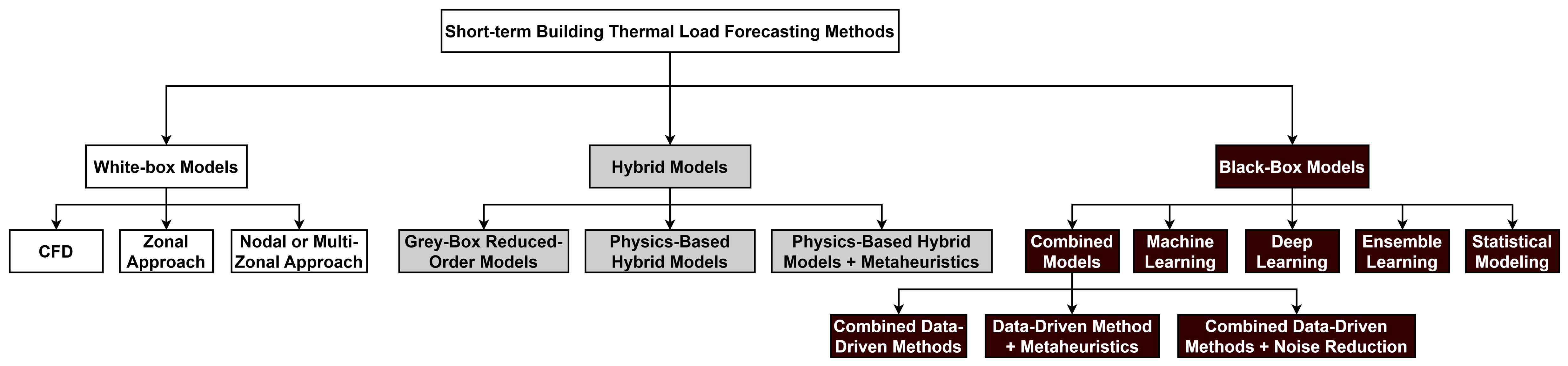}
  \caption{Classification of short-term building thermal load forecasting methods.}
   \label{fig:Classification_Diagram}
\end{figure*}

Numerous approaches are reported in the literature for short-term prediction of building thermal demand, and they can be distinctly classified into three macro-categories: (i) physical-based or white-box models, such as energy simulation computer tools, (ii) data-driven or black-box models, and (iii) a combination of the aforementioned techniques, such as grey-box models \cite{Chen2022}. The selection of the most suitable method requires a trade-off between model accuracy, computational efficiency, and interpretability \cite{Deb2017,Afroz2018,Foucquier2013,Li2014,Bourdeau2019}. White-box models are high-dimensional, nonlinear and often error-prone, since they require complete knowledge of physical phenomena, structural and thermal attributes of buildings. Albeit they are the most interpretable ones from the physical point of view, they have high computational cost due to the intricate, underlying system of partial differential equations to be solved \cite{Afroz2018,Foucquier2013,Li2014}. Black-box models are empirical models based on statistical data neglecting information on physical phenomena, structural and thermal parameters of buildings \cite{Deb2017,Afroz2018,Bourdeau2019}. They usually require a large training dataset collected over an exhaustive period of time, nevertheless, their generalization outside the training distribution may not be successful and they lack physical interpretability \cite{Foucquier2013,Bourdeau2019,Li2014,Afroz2018}. Hybrid models combine statistical information with physical phenomena, structural and thermal attributes of buildings \cite{Deb2017,Afroz2018,Bourdeau2019,Foucquier2013}. As a result, they require knowledge of both physical-based and data-driven models, however, they lack a unified design streamline. The identification of accurate hybrid models also remains a challenging task that involves the solution of complex non-convex optimization problems \cite{Foucquier2013,Bourdeau2019,Li2014,Afroz2018}. Besides that, a thorough analysis of parameter identifiability is required prior to the physical interpretation of building thermal dynamics \cite{Li2014}. The comparison of different building thermal load forecasting models is challenging, and shows that there is no clear outperformance of a specific category. 

\subsection{Literature review}
\label{lit}

The existing methods for short-term forecasts of building thermal load are classified into various categories, as illustrated in Fig.~\ref{fig:Classification_Diagram}. The white-box models can be based on three different concepts: (a) Computational Fluid Dynamics (CFD), (b) zonal and (c) nodal or multi-zonal approach \cite{Ajib2018,Atam2016}. The black-box models can be divided into five main categories: (a) Machine Learning (ML), (b) Deep Learning (DL), (c) Ensemble Learning (EL), statistical modeling and their combinations \cite{Sarker2021,Zhang2021}. The hybrid approaches are divided into the grey-box reduced order models, physics-based hybrid models and physic-based hybrid models combined with Metaheuristics. The grey-box comprises lumped element grey-box models, while the physics-based hybrid models combine the grey-box models with data-driven techniques. The physics-based hybrid models combined with Metaheuristics consist of different data-driven approaches supported further from heuristic or metaheuristic algorithms. The involved parameters of the RC-based models are calculated through the application of data-driven models or Metaheuristics. Particularly, the acquisition of building and occupants' attributes required for the definition of grey-box model parameters is usually prohibited due to data privacy restrictions imposed for the end-customers. As a result, the literature review is limited to combinations of black-box models, as another type of hybrid models that do not require any knowledge of the building characteristics. 

Various existing comparative assessments of thermal load forecasting approaches include different hybrid and black-box models that are not combined with additional techniques \cite{
Delcroix2021,Fux2013,Li2021,Potočnik2019,Wang2019,Cai2019,
Koschwitz2020,Cui2019,Maccarana2019,Kamel2020,Mustafaraj2010,Huchuk2021,Chinde2016,Steindl2019,Ferracuti2017,Ling2021,Pathak2019,Wang2020}. 
Their outcomes present diversity with respect to the models with clear outperformance. 

As for the ML-category, various methods are applied for short-term thermal load prediction in buildings, however, a single approach does not prevail over all references \cite{Delcroix2021,Fux2013,Li2021,Potočnik2019,Wang2019}. In \cite{Fux2013}, it is concluded that the least squares support vector machine (LS-SVM) surpasses neural networks (NNs) in building thermal demand forecasting. Delcroix et al. face the lack of building data, thus, black-box models are used showing a slight superiority of Nonlinear AutoRegressive network with eXogenous inputs (NARX) compared to autoregressive with external inputs (ARX) \cite{Delcroix2021}. In \cite{Li2021}, the authors evaluate the performance of different Support Vector Regression (SVR) types, i.e., Gaussian radial basis function kernel SVR (SVR-R), linear kernel SVR (SVR-L), and polynomial kernel SVR (SVR-P), alongside extreme gradient boosting (XGB), Random Forest (RF), ordinary least-squares linear regression (OLS), ridge regression (RR), least absolute shrinkage and selection operator (LASSO), elastic net (EN), and Artificial Neural Network (ANN). It is concluded that the SVR-P and SVR-R have the best accuracy in the individual building and whole stock levels, respectively. Another comparison of various forecasting models, particularly, ARX, NN with Levenberg-Marquardt (LM) or Bayesian regularization (BR), and Extreme Learning Machine (ELM), concludes on outperformance of NN with BR \cite{Potočnik2019}. Wang et al. also evaluate different data-driven and grey-box models in \cite{Wang2019}, showing that ANN achieves higher accuracy than RC and ARX ensuring global optimum owing to training with linear least squares. 

In terms of the DL methods, Cai et al. conclude that the 24h-Gated Convolutional Neural Network (GCNN) shows the highest accuracy compared to the 1h-GCNN, 1h-Gated Recurrent Neural Network (GRNN), 24h-GRNN and seasonal autoregressive integrated moving average with exogenous inputs (SARIMAX) \cite{Cai2019}. Koschwitz et al. present that NARX Recurrent Neural Networks (RNNs) can yield higher accuracy than $\epsilon$-Support Vector Regression ($\epsilon$-SVR), though general findings from the literature state that SVMs outperform ANNs \cite{Koschwitz2020}. 

Regarding the EL methods, their advantages and drawbacks are also highlighted in the literature \cite{Cui2019,Maccarana2019,Kamel2020}. Cui et al. indicate that the ANN and Gradient Boosting Trees (GBM) have relatively poor performance when compared with other nonlinear methods, while the tree-based XGB and RF yield the highest accuracies \cite{Cui2019}. The better performance of RF models compared to standard model-based approaches is also highlighted in \cite{Maccarana2019}, however, its disadvantage is the lack of robustness to changing conditions. In \cite{Kamel2020}, XGB provides the most accurate forecast from an extensive comparison of Filter method, Wrapper Backward Elimination (WBE), Wrapper Recursive Feature Elimination (WRFE), LASSO regression, linear regression (LR) and XGB. 

As for the statistical models, in \cite{Mustafaraj2010}, it is shown that the Box–Jenkins (BJ) provides better prediction results than both the ARX and autoregressive moving average with external inputs (ARMAX) models, nevertheless, the conclusion cannot be generalized. In another comprehensive comparison of data-driven thermal models for multi-hour forecasts, grey-box 1R1C, ARX, RF, LASSO and RR are assessed, showing that the LASSO and RR have the highest performance \cite{Huchuk2021}. 

Contradictory results about the performance of different models are also presented in various comparative studies \cite{Chinde2016,Steindl2019,Ferracuti2017,Ling2021,Pathak2019,Wang2020}. Particularly, the comparison of Gaussian Process (GP), ARX/ARMAX, semiparametric models and ANNs cannot also lead to a clear outperformance of an individual method \cite{Chinde2016}. Steindl et al. also concluded that the k-Nearest Neighbor (k-NN) regression does not outperform different lumped grey-box models (5R3C, 3R3C, 2R2C) in all examined periods \cite{Steindl2019}. \cite{Ferracuti2017} shows preference of ARX models for 1h-ahead prediction in comparison to lumped element grey-box and NARX models, while the NARX yields the lowest error for 1-step ahead forecasts. In another study \cite{Ling2021}, Ling et al. achieve lower average prediction errors with SVR models under the hourly correlation analysis, while the back propagation neural network (BPNN) surpasses under non-hourly correlation analysis. Pathak et al. show that the Bayesian Markov Chain Monte Carlo (MCMC) and Automatic Differentiation Variational Inference (ADVI) algorithms can provide estimations of the same order, however, both have strengths and drawbacks \cite{Pathak2019}. Particularly, ADVI is faster at the expense of providing only point estimates with uninformed priors, while MCMC is suited for constructing hierarchical models with multiple priors on model parameters. In \cite{Wang2020}, both shallow ML algorithms (LR, RR, LASSO, EN, SVM, RF, XGB) and DL methods (Vanilla Deep Neural Network (VDNN) and Long Short Term Memory (LSTM)) are evaluated, showing outperformance of LSTM and XGB for short-term and long-term predictions, respectively.


A further category comprises specific data-driven approaches enhanced with other data-driven techniques \cite{Rätz2019,Cen2024,Dong2022,Ren2022,Liu2023,Yan2023,Yan2024}. Rätz et al. propose an automated data-driven ML tool that combines any of the multilayer perceptron (MLP) ANN, ($\epsilon$-SVM-R), RF, GBM and LASSO leading to higher performance than each individual model \cite{Rätz2019}. In \cite{Cen2024}, a combination of different data-driven methods, particularly, the multi-task learning model incorporates patching operation, Temporal Convolutional Network and Time-Series Transformer (PatchTCN-TST) is applied. Dong et al. evaluate a hybrid prediction model based on decoupled weight decay Adam optimized Improved LSTM (ILSTM) neural network, named as DwdAdam-ILSTM \cite{Dong2022}. The weight decay term is introduced in the parameter update after the gradient moment estimation and the bias correction of the traditional Adam. Ren et al. also propose a hybrid thermal load forecasting model utilizing CNN, LSTM, and AR (CLSTM-AR) combined with the multi-dimensional feature fusion (MFFCLA) \cite{Ren2022}. Liu et al. combine a time-series shifting method based on the PCC with a Gate Recurrent Unit (GRU) DL model \cite{Liu2023}. After time-series shifting, the Kaiser-Meyer-Olkin (KMO) test is applied to assess the correlation of input with the output features. The SVR and RF models are also examined in the place of GRU. Yan et al. propose an improved Feature-Time Transformer Encoder-Bi-LSTM (FTTrans-E-BL) model, based on the Transformer (TST) architecture \cite{Yan2023}. Firstly, Windowed Time-Lagged Cross-Correlation (WTLCC) is employed, followed by the encoder, which consists of a modified TST model and the decoder, which involves a Bi-LSTM model. In \cite{Yan2024}, the short-term prediction model for integrated energy load aggregation combines clustering, prediction and ensemble learning methods. Particularly, Affinity Propagation (AP) clustering, prediction based on the Feature-Temporal Attention-enhanced Bi-directional Simple Recurrent Unit (FTA-Bi-SRU) and Light Gradient Boosting Machine (LightGBM) model with Bayesian optimization compose the hybrid forecasting model.

An extension of combined data-driven techniques includes noise reduction functionalities \cite{Wang2022,Song2024,Huang2024}. A hybrid model denoted as WTD–CNN–LSTM for the cooling load prediction is also proposed in \cite{Wang2022}. Initially, Wavelet Threshold Denoising (WTD) is applied for denoising the cooling load sequence, and next, a neural network based on the CNN and LSTM is established. Song et al. present a short-term cooling load forecasting framework, which also includes the stages of data adaptive noise reduction and hybrid model construction \cite{Song2024}. First, the complete ensemble empirical mode decomposition with adaptive noise (CEEMDAN), sample entropy analysis and WTD are utilized for mode decomposition. Next, the hybrid model is constructed based on a combination of Gate-controlled multi-head time convolution network (GMTCN) and an embedded bidirectional long short-term memory (BiLSTM) network with Sparse probabilistic self-attention (SPSA). In \cite{Huang2024}, the Empirical Modal Decomposition (EMD) is applied for data decomposition. Next, Pearson’s correlation coefficient (PCC) and Gradient Boosting Decision are utilized for extracting the input features of the hybrid EMD-LSTM-Markov model. 

\begin{figure*}
  \includegraphics[width=\textwidth]{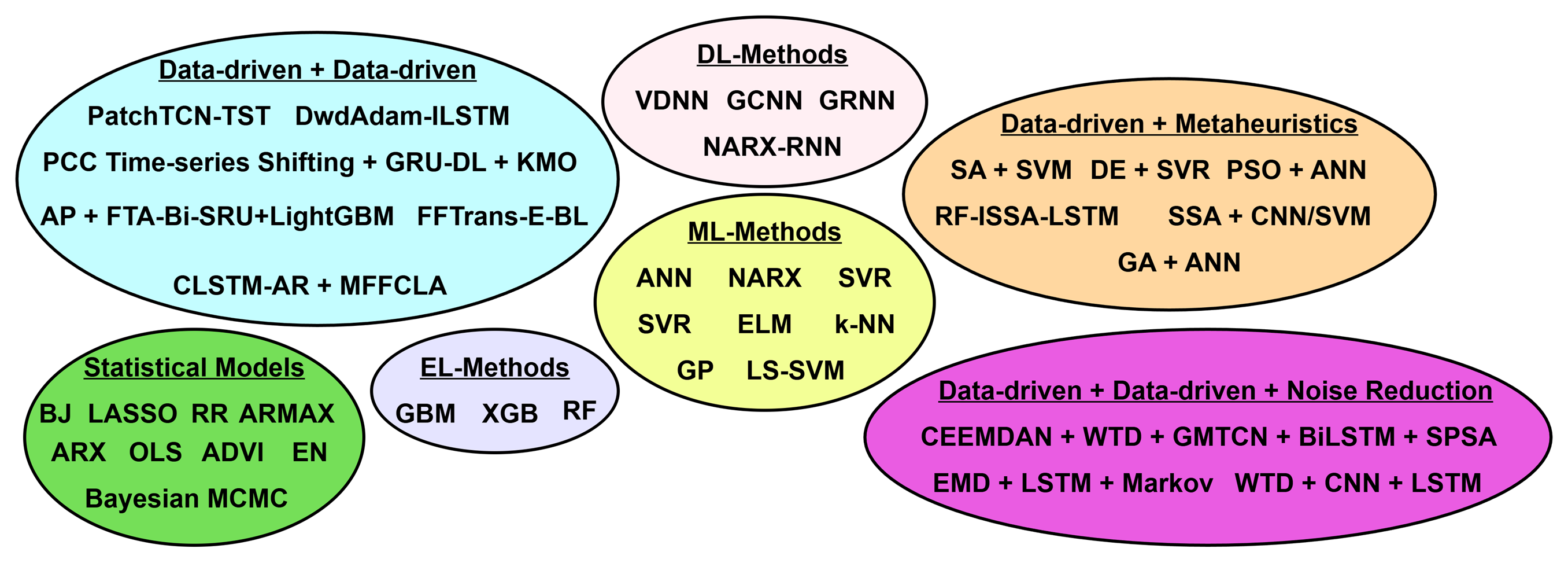}
  \caption{Classification of the techniques applied for short-term thermal load forecasting in buildings.}
   \label{fig:Classification_techniques}
\end{figure*}


Further studies combine data-driven methods with metaheuristics, which optimize the hyperparameters of the data-driven method \cite{Godinho2021,Li2015,An2024,Liu2023b}. As highlighted in \cite{Godinho2021}, hybrid approaches can enhance black-box models, i.e., simulated annealing (SA) as metaheuristic method is adopted in SVM leading to higher accuracy compared to conventional SVM and ANN. Particle Swarm Optimization (PSO) and Genetic Algorithms (GAs) combined with ANNs can also improve the ANN accuracy in \cite{Li2015}. From the results, it is concluded that the MLP combined with the PSO Grey Wolf Optimization (GWO) yields the highest accuracy. In \cite{An2024}, the authors also propose a hybrid forecasting model that combines ML models with Metaheuristics. Particularly, the sparrow search algorithm (SSA) is used to optimize the CNN and SVM in order to predict the secondary supply temperature. Authors in \cite{Liu2023b} propose the hybrid RF-ISSA-LSTM model based on RF, Improved Sparrow Search Algorithm (ISSA) and LSTM for cooling load forecasting. Firstly, the RF is applied to select high-impact features reducing the dimensionality of the original input data. Secondly, the ISSA metaheuristic algorithm identifies the optimal combination of LSTM hyperparameters. 

From the literature review, it is clear that there is a trend to combination of different methods. Particularly, the combined black-box models, as well as the hybrid or data-driven methods enhanced with Metaheuristics present the highest share, as shown in Fig.~\ref{fig:Classification_techniques}. Next, ML-based methods and statistical models are also frequently applied for short-term thermal demand forecasting in buildings. DL-methods and physics-based hybrid models are partially used, followed by the EL-techniques and the combined data-driven methods, which are reinforced with noise reduction. The reported techniques obtain diverse results without a clear preference of an individual method. Several factors hinder the outperformance of a specific approach, particularly, complex building attributes and non-linear building systems, unpredictable occupants' behavior, weather conditions, bad or missing data, as well as data produced by the control systems. While an individual model outperforms other methods in the context of a comparative study, in other analyses the respective model may be surpassed by other techniques due to the aforesaid reasons. Therefore, it is also very challenging to develop a scalable, low-cost approach to simulate accurate models of different buildings’ thermal envelopes that can: (a) learn both the envelope dynamics and disturbance patterns in short training time, (b) be adaptable to future changes in building architecture and demand and, (c) provide stable extrapolation based on the existing available data. 

From the above analysis, it is also clear that there exists no comprehensive study that aims to define and compare effective data-driven and hybrid techniques. To the best of the authors’ knowledge, the comparison of various effective hybrid approaches based on different state-of-the-art methods is missing in the literature. All previous surveys either assess state-of-the-art techniques, or develop a hybrid technique based on state-of-the-art models and compare it with conventional methods. All previous surveys either assess state-of-the-art techniques \cite{
Delcroix2021,Fux2013,Li2021,Potočnik2019,Wang2019,Cai2019,
Koschwitz2020,Cui2019,Maccarana2019,Kamel2020,Mustafaraj2010,Huchuk2021,Chinde2016,Steindl2019,Ferracuti2017,Ling2021,Pathak2019,Wang2020}, or develop a hybrid technique based on state-of-the-art models and compare it with conventional methods \cite{Rätz2019,Bashir2020,Cen2024,Dong2022,Ren2022,Liu2023,Yan2023,Yan2024,Wang2022,Song2024,Huang2024,Godinho2021,Li2015,An2024,Liu2023b}. The present study addresses this gap by focusing on the subsequent comparative assessment on the latter group of enhanced hybrid models.  

\subsection{Approach and main contributions}
\label{cont}

The main purpose of this paper is to carry out a comparative analysis of different short-term hybrid forecasting models for building thermal demand. Though multiple relevant studies are reported in the literature, this work tackles the comparison of data-driven models enhanced with other state-of-the-art techniques. Hence, the authors do not focus on the extensively reported physics-based or lumped element grey-box models, but on hybrid methods consisted of different state-of-the-art data-driven approaches. Particularly, a comprehensive assessment of the combined black-box models, data-driven methods enhanced with metaheuristics and data-driven methods combined with noise reduction models. The main reason of this decision is that the enhanced hybrid techniques show clear advantages compared to the conventional data-driven models. Consequently, the comparison of different enhanced hybrid methods can lead to clearer outcomes about their outperformance. 

At the second step, the qualitative comparison of the existing techniques is carried out with the main goal to address their main strengths and weaknesses. In particular, a brief introduction about the involved models and applied procedure, as well as their main requirements for inputs, outputs and specific prerequisites or conditions are provided. The main advantages and drawbacks of each approach are reported, highlighting its out- or under-performance against other models, as concluded in comparative studies. Since the pre-assessment at the second step includes hybrid approaches from multiple references, the selected methods may be based on different state-of-art data-driven models. Furthermore, the selected methods can belong to different categories of Fig.~\ref{fig:Classification_techniques}. 

The forecasting models of most references are usually evaluated under different conditions including building envelope, historical weather and thermal demand data. Hence, a further assessment with common inputs is required for the selected hybrid methods. Within the context of this evaluation, an open-source database with historical thermal energy consumption data from various residential buildings in United Kingdom (UK) is used \cite{IDEAL}. The assessment includes the computation of different accuracy indicators for the properties with the highest coverage of historical data. The goal of this additional evaluation is to determine hybrid combinations that show outperformance under the same conditions. In this manner, the preselected techniques from different studies can be further constrained providing clearer outcomes about the most effective hybrid methods for short-term building thermal load forecasts.

The contributions of this comparative analysis, are summarized, as follows:
\begin{itemize}
    \item Comprehensive analysis of various hybrid thermal load forecasting models considering input requirements, strengths and weaknesses. 
    \item Comparison of various hybrid techniques, which consist of different state-of-the-art data-driven models. 
    \item Further evaluation of various hybrid methods using a real dataset of thermal energy demand from various residential buildings. 
\end{itemize}

\subsection{Structure of the paper}
\label{str}

The structure of the paper is organized as follows: Section~\ref{hybrid} compares the existing hybrid data-driven models considering their input requirements, strengths, and weaknesses. Section~\ref{inp_ass} describes the utilized dataset of residential energy consumption and historical weather forecasts, as well as simplifications and assumptions considered for the simulated buildings. Next, the results from the assessment of selected hybrid approaches are analyzed including a discussion with limitations and future works in Section~\ref{res}. Finally, Section~\ref{sum} provides the main conclusions of this study.

\section{Hybrid data-driven models}
\label{hybrid}

In this section, existing hybrid data-driven models for short-term prediction of building thermal load are reviewed. In particular, three categories are considered: combined black-box models, data-driven methods enhanced with Metaheuristics, and data-driven methods enhanced with noise reduction. The input and output features of the aforementioned methods, as well as their advantages and drawbacks are presented. 

\subsection{Existing approaches}

As for the combined data-driven methods, a plethora of approaches show outperformance when compared to other state-of-the-art methods. Particularly, the existing literature includes combination of black-box models \cite{Cen2024,Dong2022,Ren2022,Liu2023,Yan2023,Yan2024}, hybrid data-driven techniques enhanced with noise reduction \cite{Wang2022,Song2024,Huang2024}, and data-driven techniques combined with Metaheuristics \cite{Godinho2021,Li2015,An2024,Liu2023b}.

Cen et al. present a PatchTCN-TST method that consists of instance normalization and de-normalization, patching operation, TCN embedding and a TST-based stacked encoder structure \cite{Cen2024}. The multi-tasking approach can be applied for the prediction of day-ahead DHW and heating load profiles. First, the univariate load data is normalized by mean and standard deviation, and next segmented into several sub-series. In the embedding phase, the TCN block extracts temporal attributes and transforms each sub-series to align with the desired model dimensions. Then, positional features are integrated into the embedded sequence to facilitate the model in the examination of the relative positions. Subsequently, stacked TST encoders and a linear layer are applied for the prediction of the normalized single-load profile. Finally, multiple individual univariate data are denormalized and concatenated to generate the multivariate output profiles.

In \cite{Dong2022}, the hybrid DwdAdam-ILSTM model is based on an improved LSTM neural network and improved Adam optimizer. The most relevant parameters are determined by the cross-correlation analysis. The data set used for training and testing is firstly normalized. Next, a grid search method is used to determine the batch, number of hidden layers, and number of hidden units of the ILSTM. Then, the DwdAdam-optimizer is used to optimize the ILSTM parameters that minimize the loss function. 

In \cite{Ren2022}, Ren et al. apply AR-based time-series feature extraction, CLSTM-based spatial-temporal feature extraction, LSTM-based environment feature extraction, and multidimensional feature fusion (MFFCLA). For the spatial-temporal correlation of the data, the auto-correlation is used for the load data, and the spatial coupling between the data is evaluated through the PCC method. Within the MFFCLA module, the extracted multi-dimensional abstract features are fused using Concatenate and Add layers, and finally, load prediction is generated by fully connected layers.

Liu et al. apply PCC-based time-series shifting to the outdoor dry bulb temperature and HVAC system energy consumption \cite{Liu2023}. Next, the KMO test is applied to evaluate the factor correlation of all input variables with the output variables. Apart from the GRU, as a LSTM variant, the SVR and RF are also evaluated. Based on manual tuning process, the most optimal hyperparameters are selected for each prediction model. 

In \cite{Yan2023}, authors propose the improved Feature-Time TST Encoder-Bi-LSTM (FTTrans-E-BL) model which consists of three major stages: data preprocessing, correlation analysis, and prediction. The data preprocessing stage includes data reading, missing values filling, and outliers processing. The correlation analysis stage is carried out using WTLCC, where the input components and input steps to the model are selected based on the correlation coefficients. The prediction stage is initiated with normalization, followed by the encoder, which consists of a modified TST model and the decoder that involves a Bi-LSTM model. The output part comprises a linear layer for deterministic prediction, and a likelihood layer for uncertainty prediction. Pareto optimization is applied to update the weights of different losses and equalize the errors of each load prediction. The proposed model is also extended for probabilistic prediction. The performance of the proposed model is compared with various forecasting models, particularly, RNN, GRU, Bi-LSTM, CNN Bidirectional GRU Attention (CNN-Bi-GRU-ATT), CNN Bidirectional LSTM Attention (CNN-Bi-LSTM-ATT), Transformer and Transformer-based Encoder-Bidirectional LSTM Memory (TST-E-Bi-LSTM). 

Yan et al. propose a three-stage approach which consists of clustering, prediction and integration \cite{Yan2024}. In the clustering stage, energy data preprocessing is carried out, while dimensionality of individual user energy data is reduced through the Largest-Triangle-Three-Buckets (LTTB) algorithm. Next, cluster analysis is performed using the multi-dimensional similarity and Affinity Propagation (AP). In the prediction stage, first, the inputs for the cluster-specific prediction models are determined through the transfer entropy analysis. Next, the FTA-Bi-SRU model is applied for load prediction based on the Bi-SRU architecture incorporating feature attention. Moreover, loss equilibrium is considered to balance the prediction accuracy. The Multi-Term Adam (MTAdam)-Dynamic Weight Average (DWA) algorithm is applied to facilitate a more precise adjustment of model parameters in order to enhance the training effectiveness of multi-task learning. In the integration stage, the BO-LightGBM ensemble learning framework is applied for the predicted results of individual clusters to generate the forecasts for the integrated energy load aggregation.

Wang et al. propose the hybrid WTD–CNN–LSTM cooling load forecasting model \cite{Wang2022}. First, WTD is applied for denoising the cooling load data selecting the soft threshold function as the threshold. Next, the prediction model consists of the upper CNN and the lower LSTM. The CNN includes an input layer, two convolutional layers and a pooling layer. The LSTM receives the time information of important features for the cooling load, as extracted by the CNN. The CNN output is transmitted to the gate unit, while the LSTM preserves the long-term memory by merging and updating the previously hidden memory units. Finally, the fully connected layer connects the LSTM output, which represents the load forecast. 

\begin{table*}
\caption{\textrm{Input and output features of the selected hybrid data-driven forecasting models.}}
\label{tab:hybrid-input-output}
\begin{tblr}{colspec={@{}m{0.6cm}|m{11.5cm}|m{4.4cm}@{}}, rows={valign=t}}
\hline\hline
\textrm{Ref.} & \textrm{Inputs} & \textrm{Outputs} \\
\hline

\textrm{\cite{Cen2024}} &
\tabitem \textbf{\textrm{Historical outputs:}} \textrm{Electricity demand of lighting, HVAC units and plug loads for each of the 33 building zones.}
\par
\tabitem \textbf{\textrm{Building and indoor environment:}} \textrm{Historical indoor ambient light (IAL), indoor ambient temperature (IAT), and indoor relative humidity (RH).} &
\tabitem \textrm{Electricity demand of lighting, HVAC units and plug loads.}
\par
\tabitem \textrm{Indoor ambient light.}
\par
\tabitem \textrm{Indoor ambient temperature.}
\par
\tabitem \textrm{Indoor relative humidity.}
\\
\hline

\textrm{\cite{Dong2022}} &
\tabitem \textbf{\textrm{Weather:}} \textrm{Solar radiation, outdoor temperature, outdoor humidity.}
\par
\tabitem \textbf{\textrm{Historical outputs:}} \textrm{Cooling load during the previous six hours.} &
\tabitem \textrm{Hourly cooling load.}
\\
\hline

\textrm{\cite{Ren2022}} &
\tabitem \textbf{\textrm{Weather:}} \textrm{Outdoor temperature, dew point temperature, humidity, air pressure, wind speed.}
\par
\tabitem \textbf{\textrm{Calendar:}} \textrm{Working day type.}
\par
\tabitem \textbf{\textrm{Historical outputs:}} \textrm{Electricity, heating and cooling loads (previous 4 h and previous 7 h), together with the corresponding values at $T$, $T-1$ and $T+1$ of the previous week.} &
\tabitem \textrm{Electricity load.}
\par
\tabitem \textrm{Heating load.}
\par
\tabitem \textrm{Cooling load.}
\\
\hline

\textrm{\cite{Liu2023}} &
\tabitem \textbf{\textrm{Weather:}} \textrm{Outdoor dry-bulb temperature, outdoor relative humidity, solar radiation intensity (time-shifted)}.
\par
\tabitem \textbf{\textrm{Building and indoor environment:}} \textrm{Indoor CO$_2$ concentration, indoor dry-bulb temperature, indoor relative humidity (time-shifted)}.
\par
\tabitem \textbf{\textrm{Historical outputs:}} \textrm{HVAC energy consumption (1 h lag).} &
\tabitem \textrm{HVAC energy consumption.}
\\
\hline

\textrm{\cite{Yan2023}} &
\tabitem \textbf{\textrm{Weather:}} \textrm{Air pressure, dew point temperature, air temperature, wind speed, wind direction (4 h lag).}
\par
\tabitem \textbf{\textrm{Calendar:}} \textrm{Month, date, day of week, day type, hour (4 h lag).}
\par
\tabitem \textbf{\textrm{Historical outputs:}} \textrm{Electricity, cooling and heating loads (1--24 h historical values).} &
\tabitem \textrm{Electricity load.}
\par
\tabitem \textrm{Cooling load.}
\par
\tabitem \textrm{Heating load.}
\\
\hline

\textrm{\cite{Yan2024}} &
\tabitem \textbf{\textrm{Weather:}} \textrm{Outdoor temperature, relative humidity, wind speed, wind direction, global horizontal radiation, direct normal radiation, diffuse horizontal radiation.}
\par
\tabitem \textbf{\textrm{Calendar:}} \textrm{Month, day, hour, day of week, day type.}
\par
\tabitem \textbf{\textrm{Historical outputs:}} \textrm{Electricity, cooling, heating and gas loads (1--24 h historical values).} &
\tabitem \textrm{Electricity load.}
\par
\tabitem \textrm{Cooling load.}
\par
\tabitem \textrm{Heating load.}
\par
\tabitem \textrm{Gas load.}
\\
\hline

\textrm{\cite{Wang2022}} &
\tabitem \textbf{\textrm{Weather:}} \textrm{Temperature, humidity.}
\par
\tabitem \textbf{\textrm{Building and HVAC system:}} \textrm{Number of CAC chillers.}
\par
\tabitem \textbf{\textrm{Historical outputs:}} \textrm{Cooling load.} &
\tabitem \textrm{Cooling load.}
\\
\hline

\textrm{\cite{Song2024}} &
\tabitem \textbf{\textrm{Weather:}} \textrm{Temperature, atmospheric pressure, dew point temperature, relative humidity, solar radiation, diffuse radiation, wind speed.}
\par
\tabitem \textbf{\textrm{Calendar:}} \textrm{Day of year, season, month, day of week, hour, day type.}
\par
\tabitem \textbf{\textrm{Historical outputs:}} \textrm{Cooling load (previous 24 h).} &
\tabitem \textrm{Cooling load.}
\\
\hline

\textrm{\cite{Huang2024}} &
\tabitem \textbf{\textrm{Weather:}} \textrm{Ambient temperature, outdoor relative humidity, maximum and minimum daily temperature.}
\par
\tabitem \textbf{\textrm{Historical outputs:}} \textrm{Cooling load (previous two timesteps).} &
\tabitem \textrm{Cooling load.}
\\
\hline

\textrm{\cite{Godinho2021}} &
\tabitem \textbf{\textrm{Weather:}} \textrm{Relative humidity, outdoor air temperature, wind speed, direct normal solar radiation, diffuse horizontal radiation.}
\par
\tabitem \textbf{\textrm{Building and HVAC system:}} \textrm{Estimated occupancy profile.}
\par
\tabitem \textbf{\textrm{Historical outputs:}} \textrm{Cooling or heating load of the same day during the previous one and two weeks.} &
\tabitem \textrm{Cooling load.}
\par
\tabitem \textrm{Heating load.}
\\
\hline

\textrm{\cite{Li2015}} &
\tabitem \textbf{\textrm{Weather:}} \textrm{Outdoor dry-bulb temperature, solar radiation.}
\par
\tabitem \textbf{\textrm{Calendar:}} \textrm{Hour of day.}
\par
\tabitem \textbf{\textrm{Building and HVAC system:}} \textrm{Occupancy level.}
\par
\tabitem \textbf{\textrm{Historical outputs:}} \textrm{Building electricity consumption (previous two hours).} &
\tabitem \textrm{Electricity load.}
\\
\hline

\textrm{\cite{An2024}} &
\tabitem \textbf{\textrm{Weather:}} \textrm{Outdoor temperature.}
\par
\tabitem \textbf{\textrm{Building and HVAC system:}} \textrm{Indoor temperature, primary supply temperature, primary return temperature, secondary return temperature.} &
\tabitem \textrm{Secondary supply temperature of the heating station.}
\\
\hline

\textrm{\cite{Liu2023b}} &
\tabitem \textbf{\textrm{Weather:}} \textrm{Outdoor dry-bulb temperature, outdoor relative humidity, solar radiation.}
\par
\tabitem \textbf{\textrm{Building and HVAC system:}} \textrm{Indoor human flow, lighting time.}
\par
\tabitem \textbf{\textrm{Historical outputs:}} \textrm{Cooling load (previous timestep).} &
\tabitem \textrm{Cooling load.}
\\

\hline\hline
\end{tblr}
\end{table*}

\begin{table*}[t!]
\small
\caption{\textrm{Advantages and drawbacks of selected hybrid data-driven load forecasting models with references}}
\label{tab:hybrid-advantages-drawbacks}
\begin{tblr}{
  colspec={@{}m{2.0cm}|m{9.5cm}|m{4.3cm}@{}},
  abovesep=2pt,
  belowsep=2pt,
  cell{2-Z}{2-3}={halign=l} 
}
\hline \hline
\textrm{Model} & \textrm{Advantages} & \textrm{Drawbacks} \\
\hline

\textrm{PatchTCN--TST \cite{Cen2024}} &
\tabitem \textrm{Channel-independent forecasting enables scalable multivariate prediction.} \par
\tabitem \textrm{Patching improves local dependency extraction and reduces attention complexity.} \par
\tabitem \textrm{TCN enhances temporal feature representation.} \par
\tabitem \textrm{Superior accuracy across multiple metrics and scenarios.}
&
\tabitem \textrm{Inter-channel dependencies (e.g., temperature and AC power correlation) are neglected.} \\
\hline

\textrm{DwdAdam--ILSTM \cite{Dong2022}} &
\tabitem \textrm{Captures long-term dependencies while avoiding gradient issues.} \par
\tabitem \textrm{Superior convergence, stability, and accuracy over SVR, BPNN, and LSTM variants.} \par
\tabitem \textrm{Strong generalization with limited training samples.}
&
\tabitem \textrm{Generalization validated mainly for commercial buildings.} \\
\hline

\textrm{CLSTM--AR--MFFCLA \cite{Ren2022}} &
\tabitem \textrm{Combines linear (AR) and nonlinear (LSTM/CLSTM) modeling.} \par
\tabitem \textrm{Multi-dimensional feature fusion improves spatial--temporal learning.} \par
\tabitem \textrm{Stable performance for peak and non-workday loads.}
&
\tabitem \textrm{Increased architectural and computational complexity.} \\
\hline

\textrm{PCC--GRU \cite{Liu2023}} &
\tabitem \textrm{PCC-based time shifting enhances feature correlation and model stability.} \par
\tabitem \textrm{GRU models nonlinear dynamics and peak-shaving behavior.} \par
\tabitem \textrm{Mitigates energy storage interference.}
&
\tabitem \textrm{Limited performance improvement for SVR-based models.} \\
\hline

\textrm{FTTrans--E--BL \cite{Yan2023}} &
\tabitem \textrm{Feature-time attention captures temporal and global dependencies.} \par
\tabitem \textrm{Pareto-based multi-task learning improves adaptability and stability.} \par
\tabitem \textrm{Probabilistic forecasting quantifies uncertainty.}
&
\tabitem \textrm{Higher computational complexity despite reduced training time.} \\
\hline

\textrm{TE--BiSRU--MTL \cite{Yan2024}} &
\tabitem \textrm{Transfer entropy clustering captures nonlinear multi-feature dependencies.} \par
\tabitem \textrm{Temporal attention and Bi-SRU enhance accuracy and efficiency.} \par
\tabitem \textrm{Multi-task learning improves generalization across horizons.}
&
\tabitem \textrm{Increased complexity due to ensemble and multi-task frameworks.} \\
\hline

\textrm{WTD--CNN--LSTM \cite{Wang2022}} &
\tabitem \textrm{Wavelet denoising removes noise preserving data characteristics.} \par
\tabitem \textrm{Effectively captures spatial--temporal correlations.} \par
\tabitem \textrm{Outperforms multiple benchmark models across horizons.}
&
\tabitem \textrm{Higher computational cost due to multi-stage processing.} \\
\hline

\textrm{aTCN--LSTM \cite{Song2024}} &
\tabitem \textrm{Strong short- and long-term forecasting accuracy.} \par
\tabitem \textrm{CNN and attention mechanisms enhance feature extraction.} \par
\tabitem \textrm{Fast convergence and robust performance across timesteps.}
&
\tabitem \textrm{Increased model complexity from multiple optimization mechanisms.} \\
\hline

\textrm{EMD--LSTM--Markov \cite{Huang2024}} &
\tabitem \textrm{EMD improves responsiveness to short- and long-term trends.} \par
\tabitem \textrm{Markov correction mitigates delay and hysteresis.} \par
\tabitem \textrm{Improved prediction accuracy over benchmark models.}
&
\tabitem \textrm{Generalization evaluated only for office buildings.} \\
\hline

\textrm{SVM--SA \cite{Godinho2021}} &
\tabitem \textrm{Sensitivity analysis improves cooling demand prediction.} \par
\tabitem \textrm{Outperforms SVM and ANN for cooling load forecasting.}
&
\tabitem \textrm{ANN performs better for heating demand prediction.} \\
\hline

\textrm{GA/iPSO--ANN \cite{Li2015}} &
\tabitem \textrm{Intelligent optimization improves ANN accuracy.} \par
\tabitem \textrm{iPSO enables faster training and online prediction.}
&
\tabitem \textrm{GA-based optimization is computationally costly.} \\
\hline

\textrm{SSA--CNN--SVM \cite{An2024}} &
\tabitem \textrm{Enhanced feature representation and parameter optimization.} \par
\tabitem \textrm{High prediction stability across datasets.}
&
\tabitem \textrm{High computational complexity and limited standardization.} \\
\hline 

\textrm{RF--ISSA--LSTM \cite{Liu2023b}} &
\tabitem \textrm{RF reduces input dimensionality and training time.} \par
\tabitem \textrm{ISSA improves convergence and optimization performance.} \par
\tabitem \textrm{Strong accuracy with reduced training samples.}
&
\tabitem \textrm{Generalization tested only for library and office buildings.} 
\\
\hline \hline
\end{tblr}
\end{table*}

In \cite{Song2024}, the adaptive noise reduction layer combines the advantages of CEEMDAN, sample entropy analysis and WTD considering the adaptive and Gaussian noise sensitivity. Next, the filtered data are considered as inputs of a deep neural network model, which is based on a combination of GMTCN and an embedded BiLSTM with SPSA. GMTCN extracts crucial features and temporal patterns for the cooling load at different scales using causal dilated convolutions with varying kernel sizes. Next, gate and self-attention mechanisms are applied with focus on essential TCN blocks. BiLSTM defines both global non-linearity and partial long-term data dependencies. Subsequently, the SPSA pays attention to hidden states at critical, historical moments in order to enhance the prediction accuracy. 

Huang et al. propose the EMD-LSTM-Markov for cooling load forecasting \cite{Huang2024}. The PCC and Gradient Boosting Decision Tree (GBDT) are used to extract the input features of the prediction model. The EMD technique analyses data characteristics by decomposing the original signal into multiple Intrinsic Mode Functions (IMFs) and a residual component. In this manner, the impact of data fluctuation on the cooling load prediction forecast is mitigated. Subsequently, a clustering algorithm is used to streamline computation by aggregating components and residuals. Next, the results from the EMD-based clustering are integrated into the LSTM model to generated the cooling load prediction. Finally, the cooling load forecast are adjusted through the Markov chain to refine the model accuracy. 

In \cite{Godinho2021}, the authors compare the state-of-the-art SVM and ANNs with the SVM enhanced with simulated annealing (SA). The Radial Basis Function (RBF) is the kernel function used for the SVM. The SA is applied every hour to optimize the SVM hyperparameters, since for each individual next hour a single combination of SVM with SA is determined.


In \cite{Li2015}, the hybrid iPSO-ANN model is proposed for building energy forecasting. A three-layer feedforward network with back-propagation training algorithm is constructed for ANN model initialization. Next, the improved PSO (iPSO) is applied to adjust the ANN weights and threshold values. PCA is applied to eliminate redundant information and determine the most relevant modeling inputs. For performance comparison, ANN and the hybrid GA-ANN are also constructed.

An et al. propose the hybrid SSA-CNN-SVM model adopting the Sparrow search algorithm (SSA) in CNN and SVM \cite{An2024}. PCC is applied to evaluate the correlation between the output temperature and various influencing factors. Firstly, the main features are extracted from the original data applying CNN. Next, the selected features are integrated into the SVM for the regression prediction. Finally, the hyperparameters of the CNN-SVM are optimized and adjusted through the SSA with the goal to enhance the regression performance of the hybrid model.

Liu et al. evaluate the performance of a hybrid model based on RF, improved sparrow search algorithm (SSA) and LSTM \cite{Liu2023b}. First, RF is applied to rank the high-impact features for building cooling load and reduce the dimensionality of the original input data. Next, three strategies are introduced to improve the location update in different stages of the ISSA. Lastly, the ISSA-optimized LSTM forecasting model is generated. 

\subsection{Input and output features}
\label{inp_out}

Table~\ref{tab:hybrid-input-output} summarizes the input and output variables reported in the literature for the aforementioned hybrid models. Four types of input features, particularly, weather, calendar, building and HVAC system-oriented, as well as historical data of output features, are mainly used. In terms of the weather factors, either historical data or forecasts are considered including solar radiation components, different types of temperature, air pressure, humidity, wind speed and direction. Regarding the calendar parameters, type of day, month, date, day of week, hour and season are used as input features. Concerning the building and HVAC system, indoor building parameters, such as relative humidity, occupancy, ambient light, CO\textsubscript{2} concentration and HVAC system parameters, e.g., supply and return temperatures, are used. Historical data of the outputs, such as loads of cooling, heating, electricity, and gas, as well as HVAC energy consumption and secondary supply temperature of heating station are also selected. Finally, as for the output features, the indoor air temperature, secondary supply temperature of heating station, and demand-related variables are considered by the existing forecasting models. 

\subsection{Advantages and drawbacks}
\label{adv_draw}

The advantages and drawbacks of the selected hybrid forecasting models are summarized in Table~\ref{tab:hybrid-advantages-drawbacks} and are described below. 

Various studies employ preprocessing methods to improve input data quality before prediction, including WTD \cite{Wang2022}, EMD \cite{Huang2024}, PCC-based time-series shifting \cite{Liu2023}, transfer entropy-based feature selection \cite{Yan2024}, and RF-based feature selection \cite{Liu2023b}. These approaches effectively remove noise, identify the most informative variables, or strengthen correlations between predictors and target variables. Thereby, forecasting accuracy is improved, while reducing the influence of redundant information.

CNNs and TCNs are frequently combined with recurrent networks such as LSTM, GRU, or SRU to exploit their respective strengths in feature extraction and sequential learning. More recent studies further incorporate Transformer-based attention mechanisms \cite{Cen2024,Yan2023,Yan2024}, which improve the modeling of long-range dependencies and interactions among multiple input variables. Moreover, several models employ multi-task learning to simultaneously predict multiple energy carriers or building variables, allowing shared representations to improve overall forecasting performance.

Metaheuristic optimization algorithms, including DwdAdam \cite{Dong2022}, iPSO \cite{Li2015}, SSA \cite{An2024}, and ISSA \cite{Liu2023b}, are frequently used to optimize network parameters and hyperparameters. These methods generally improve convergence speed, training stability, and prediction accuracy compared with conventional optimization strategies.

Despite these advantages, Table~\ref{tab:hybrid-advantages-drawbacks} also reveals several recurring limitations. First, the integration of multiple preprocessing, optimization, and forecasting modules inevitably increases architectural complexity and computational requirements. Models such as FTTrans--E--BL \cite{Yan2023}, TE--BiSRU--MTL \cite{Yan2024}, WTD--CNN--LSTM \cite{Wang2022}, aTCN--LSTM \cite{Song2024}, and SSA--CNN--SVM \cite{An2024} achieve excellent predictive performance but require more sophisticated training procedures and greater computational resources, potentially limiting their deployment in real-time building energy management systems.

Second, although various hybrid models demonstrate excellent predictive accuracy, their generalization capability remains insufficiently investigated. Several studies validate their approaches using datasets from a single building category, including commercial buildings \cite{Dong2022}, office buildings \cite{Huang2024}, and library or office buildings \cite{Liu2023b}. Consequently, their robustness under different building types, occupancy patterns, climatic conditions, and operational strategies has yet to be comprehensively assessed.

Finally, certain hybrid architectures improve computational efficiency at the expense of modeling important physical relationships. For example, the channel-independent strategy adopted by PatchTCN--TST \cite{Cen2024} reduces computational complexity and facilitates multivariate forecasting but ignores inter-variable dependencies, such as the coupling between indoor temperature and HVAC electricity consumption. Similarly, while feature selection techniques effectively reduce input dimensionality, they may exclude variables that become important under changing operating conditions.

\section{Inputs and assumptions}
\label{inp_ass}

\subsection{Thermal load dataset}
\label{dataset}

The IDEAL household energy dataset is used for the short-term prediction of building thermal load \cite{IDEAL}. A sample of 255 houses from Edinburgh and the regions of the Lothians and south Fife, in Scotland, UK are considered for the data collection from August 2016 up to the end of June 2018. The dataset includes measurements from different sensors and meters including electricity apparent power, gas usage, 12-second temperature, humidity, light, and 12-second boiler pipe temperatures for central heating and hot water pipes. The focus is given on the measurements of 12-second boiler pipe temperatures for central heating and hot water pipes. Furthermore, the measurements are averaged on the time resolution of 15 minutes in order to facilitate their evaluation. From the data analysis of the sample, it is concluded that 9 out of the 255 households include continuous measurements without missing data for a period of more than one year. This prerequisite is significant for a sufficient dataset, where the first year can be considered for model training and the remaining sample will be utilized for testing. Therefore, the selected 9 households will be utilized for further assessment of the thermal load forecasts when applying the hybrid forecasting models.

From the temperatures of central heating flow and return, the heating demand Q\textsubscript{heat}($t$) (kW) at timestep $t$ is calculated by:
\begin{equation}
\label{eq:thermal_heating}
\textrm{Q}_{\textrm{heat}}(t) = \frac{\dot{\textrm{m}}_{\textrm{heat}} \cdot \textrm{c}_\textrm{p} \cdot \mathrm{\Delta T}(t)} {1000} 
\end{equation}
where the mass flow rate of the fluid $\dot{\textrm{m}}_{\textrm{heat}}$, and the specific heat capacity of the fluid c\textsubscript{p} are assumed to be 0.5 kg/s and 4180 J/kg·K, respectively \cite{Bergman2017}.  

\begin{figure}
    \centering
    \subfloat[]{
        \includegraphics[width=\columnwidth]{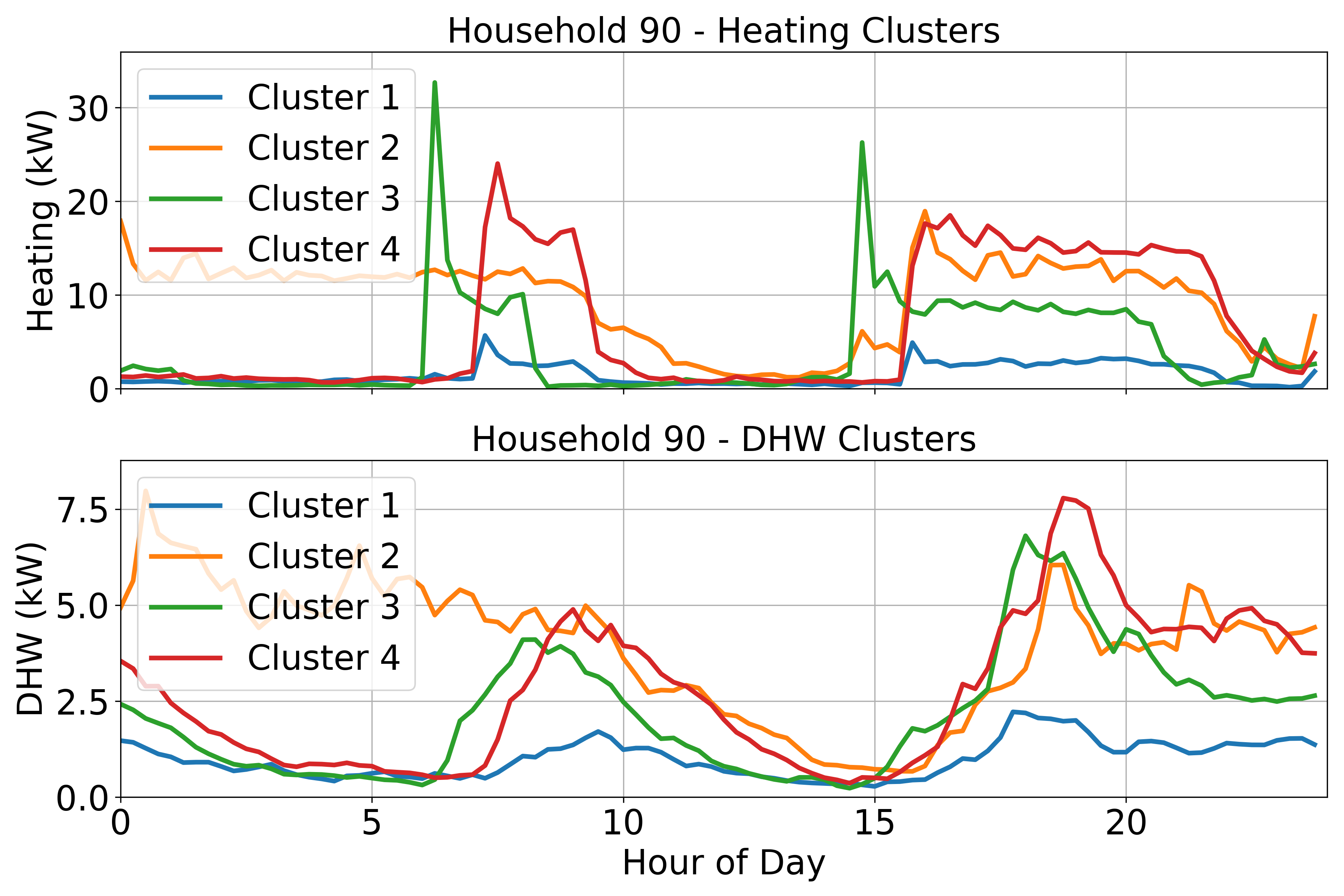}
        \label{fig:home90_clusters}
    }\\
    \subfloat[]{
        \includegraphics[width=\columnwidth]{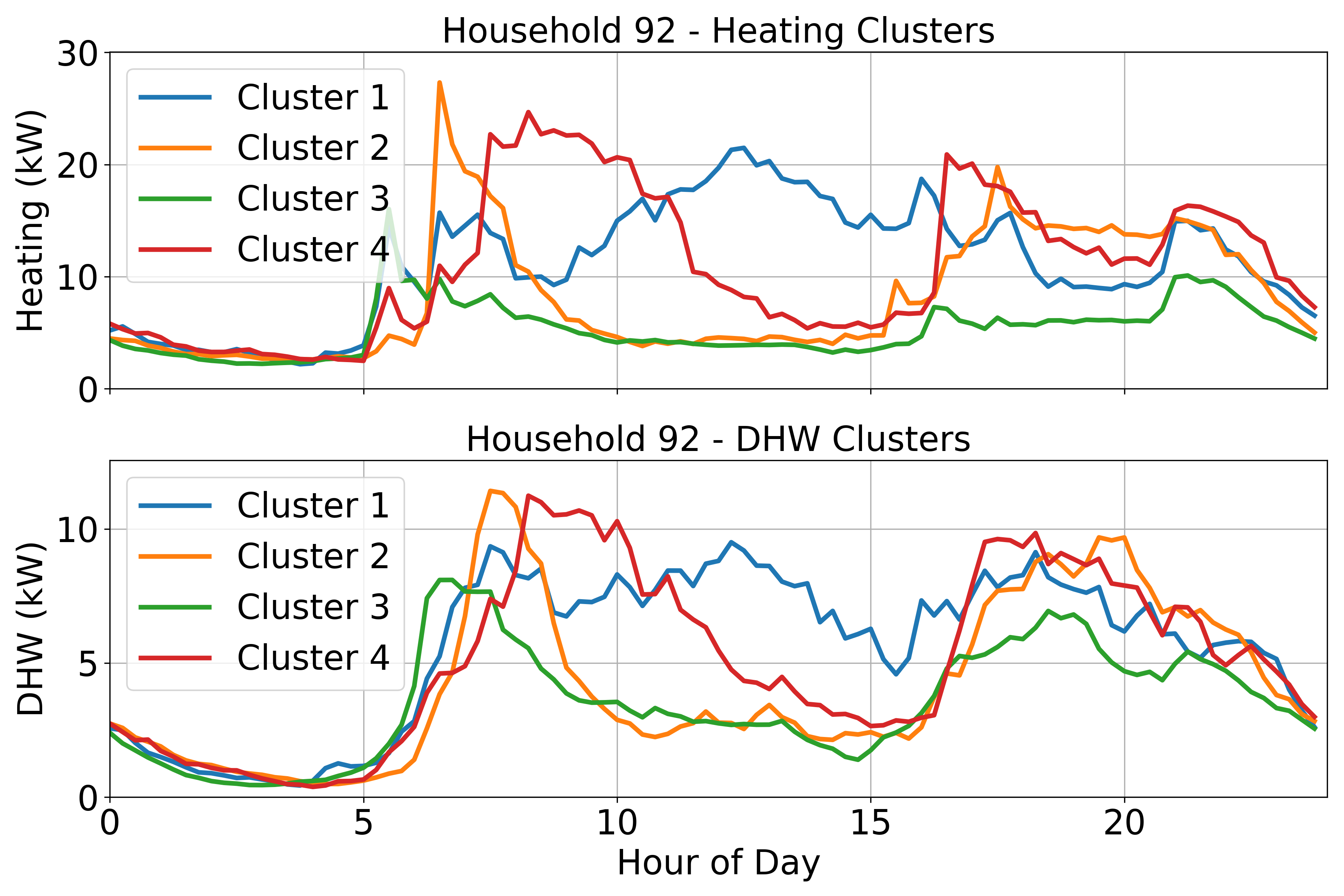}
        \label{fig:home92_clusters}
    }
    \caption{\textrm{Typical clusters of DHW and heating demand for two households: (a) household 90. (b) household 92.}}
    \label{fig:cluster_comparison}
\end{figure}

From the temperatures of hot water hot and cold pipes, the domestic hot water (DHW) demand Q\textsubscript{dhw}($t$) (kW) is calculated in the same manner to Q\textsubscript{heat}($t$):
\begin{equation}
\label{eq:dhw}
\textrm{Q}_{\textrm{dhw}}(t) = \frac{\dot{\textrm{m}}_{\textrm{dhw}} \cdot \textrm{c}_\textrm{p} \cdot \mathrm{\Delta T}(t)} {1000} 
\end{equation}
where the mass flow rate of the fluid $\dot{\textrm{m}}_{\textrm{dhw}}$ depends on each activity. In this study, a mean rate of 8 L/min (0.1333 kg/s) is assumed, since no details are given for the DHW activities.

\begin{figure*}
    \centering
    \includegraphics[width=\textwidth]{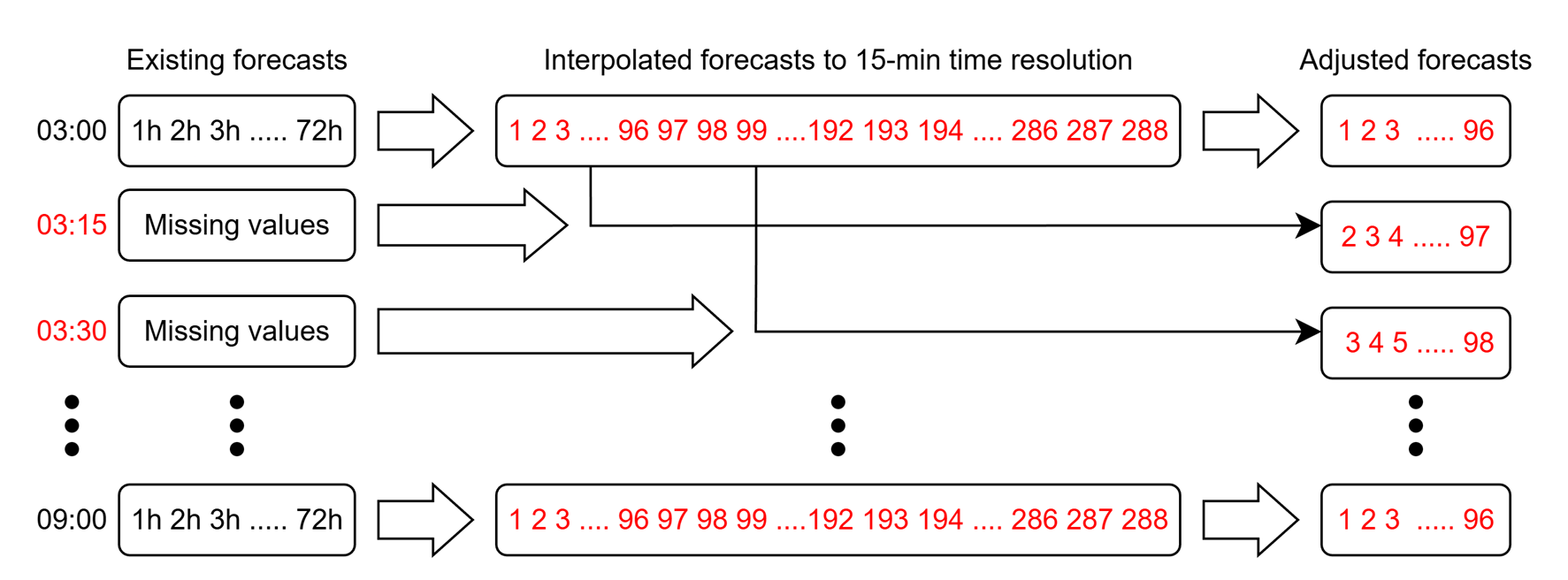}
    \caption{\textrm{Adjustment of provided weather forecasts to the prediction horizon of next day with time resolution of 15 minutes.}}
    \label{fig:Adjusted_Forecasts}
\end{figure*}

By applying the k-means clustering method separately to the daily DHW and heating demand profiles, four clusters are generated for the daily DHW and heating patterns. Figure~\ref{fig:home90_clusters} displays four clusters of heating and DHW demand for the household 90. As can be noticed in the patterns, the demand for heating and DHW consists of two cycles that start in the early morning and early evening. During noon, it is clear that the demand for heating and DHW is low. From a similar evaluation for the other households, it is concluded that not all of them present easily predicted demand patterns. For instance, in Figure~\ref{fig:home92_clusters}, it is clear that the heating and DHW profiles of cluster 1 do not show specific patterns, as the load presents various fluctuations. Nevertheless, the diversity of demand profiles among the investigated households enables the assessment of forecasting models under various conditions ensuring their robustness.

\subsection{Weather forecasts}
\label{wea_for}

As for the weather forecasts, the inputs from the provider Meteonorm are used \cite{meteonorm}. Particularly, the wind speed, two-meter air temperature, two-meter dew point temperature, relative humidity, global radiation, diffuse radiation, clear sky radiation, and direct normal radiation are predicted. For each parameter, predictions are generated four times per day at 03:00, 09:00, 15:00 and 21:00 for the next 72 hours. 

From the assessment of weather forecasts generated four times per day, there may be missing data in the 72 values. The existing data is divided into the four seasons (autumn, winter, spring and summer). In each season, the empirical distribution of existing weather forecasts is generated at each timestep, and according to this distribution the missing values are filled in. 

In this study, the short-term prediction horizon is restricted to the next day, and the time resolution of forecasts is selected as 15 minutes so as to be consistent with the adjusted historical thermal demand time-series. Therefore, the existing forecasts are interpolated on 15-min basis generating 288 values for the provided four times. Moreover, additional forecasts are generated every 15 minutes based on the 288 values interpolated for the times of 03:00, 09:00, 15:00 and 21:00. Particularly, the 96 forecasts for the missing times within the existing times, e.g., 03:15, 03:30, ..., 08:45, are based on the 288 values at the existing time, e.g, 03:00, by shifting the prediction window of 96 values by one timestep. In Fig.~\ref{fig:Adjusted_Forecasts}, the update of weather forecasts is illustrated, indicating with red color the new values. The interpolated forecasts on 15-min basis at 03:00 are utilized for the generation of adjusted forecasts at 03:00, 03:15 and so on. As can be noticed, at 03:00 the forecasts of the first 96 timesteps are assigned, and at the next 15 minutes, the window of 96 forecasts is shifting per one timestep. In this manner, at the next timestamps after 03:00, we use the forecasts produced at 03:00 and shifted by respective timesteps. As a result, it is expected that their accuracy may be lower than the forecasts that could be generated at the timestamps of 03:15, 03:30 and so on. However, in our study, we do not focus on achieving the highest accuracy with the evaluated methods, but, on comparing them under identical conditions. Consequently, the adjusted forecasts are considered as sufficient for the comparative analysis.

\subsection{Hardware and software requirements}
\label{sim_ass}

For training and validating the hybrid forecasting models, a workstation equipped an NVIDIA GeForce RTX 5070 Ti graphics card with 16 GB of RAM is used. In this manner, high-performance parallel computation is enabled reducing significantly the simulation time. The machine is powered by an AMD Ryzen Threadripper PRO 7955WX processor with 16 CPU cores and 16 GB of system RAM, providing sufficient computational resources for data processing and coordination with the GPU. The environment uses CUDA version 12.2 to enable GPU computing capabilities. The software stack runs on Windows Subsystem for Linux with an Ubuntu Linux distribution, allowing a Linux-based development environment within Windows while maintaining GPU support. GPU-accelerated data science libraries are installed through RAPIDS (rapids-nightly) inside the WSL environment. The Python scripts are executed using Python version 3.11.14, ensuring compatibility with the RAPIDS ecosystem and CUDA-enabled libraries for efficient GPU-accelerated computation. In terms of the forecasting models, their parameters are provided in the Appendix~\ref{ap:Hyperpar}.

\subsection{Assessment criteria}
\label{kpis}

Forecasting performance is evaluated using common error metrics that capture different aspects of prediction accuracy including the MAE, RMSE and the 95th Percentile Absolute Error (P95AE). These metrics quantify the discrepancy between the predicted and observed values of the time series. The y\textsubscript{$t$} denotes the observed value at time $t$, while $\hat{\textrm{y}}_t$ is the corresponding predicted value, and n represents the total number of observations. The forecast error is defined as $\textrm{e}_t = y_t - \hat{\textrm{y}}_t$.

The MAE measures the average magnitude of the forecasting errors regardless of their direction and is defined as:

\begin{equation}
\textrm{MAE} = \frac{1}{\textrm{n}}\sum_{\textrm{t}=1}^{\textrm{n}} |\textrm{y}_t - \hat{\textrm{y}}_t|
\end{equation}

The RMSE penalizes larger errors more strongly by squaring the errors before averaging and then taking the square root:

\begin{equation}
\textrm{RMSE} = \sqrt{\frac{1}{\textrm{n}}\sum_{\textrm{t}=1}^{n}(\textrm{y}_t - \hat{\textrm{y}}_t)^2}
\end{equation}

\begin{figure*}[t]
    \centering
    \subfloat[]{
        \includegraphics[width=0.47\textwidth]{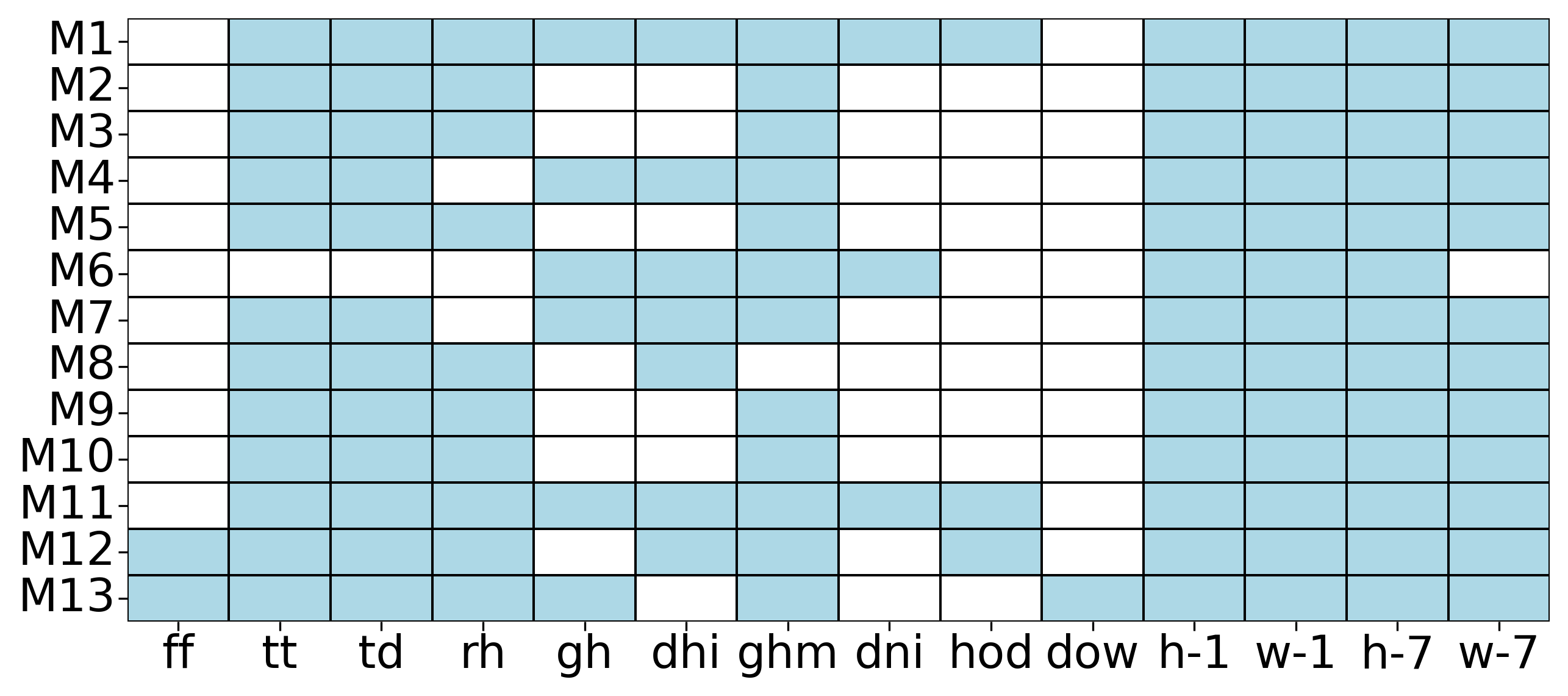}
        \label{fig:home68_heatmap}
    }
    \subfloat[]{
        \includegraphics[width=0.47\textwidth]{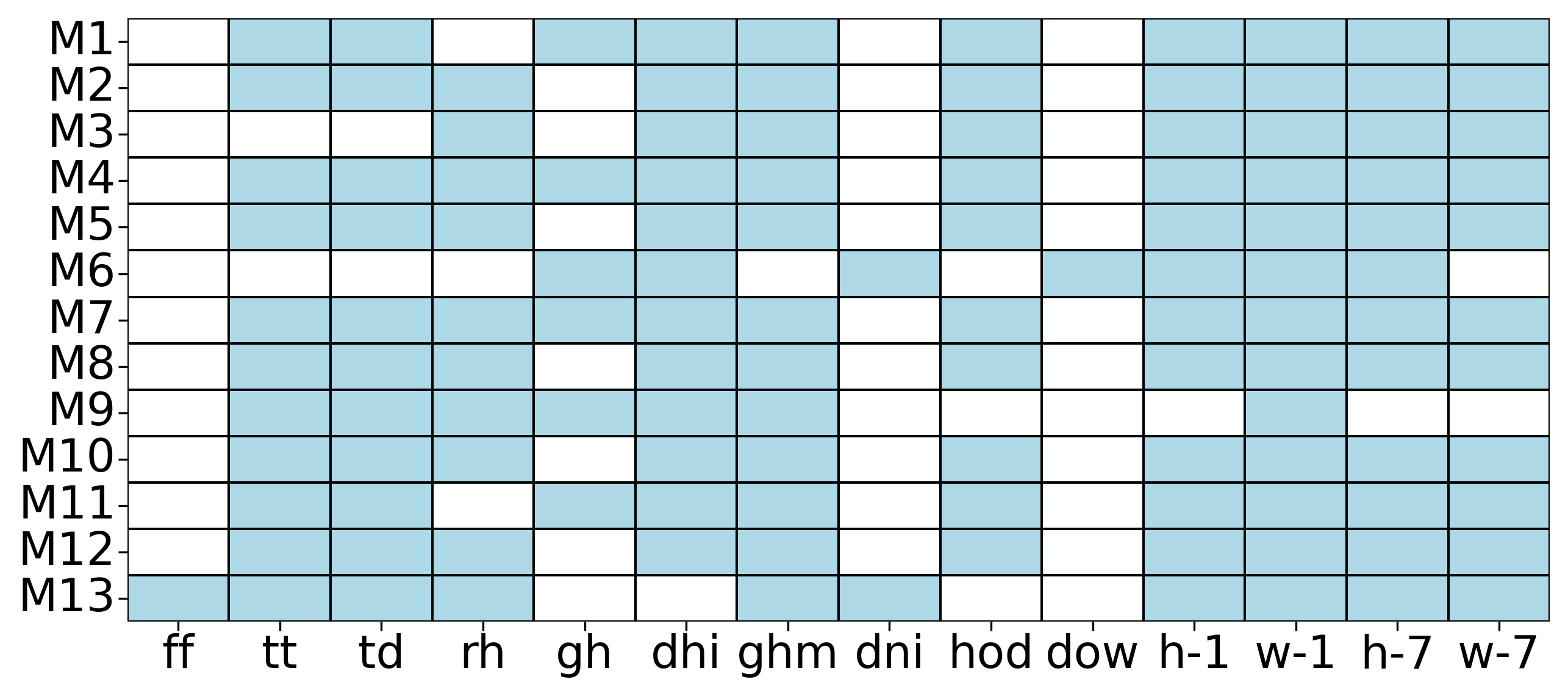}
        \label{fig:home81_heatmap}
    }\\
    \subfloat[]{
        \includegraphics[width=0.47\textwidth]{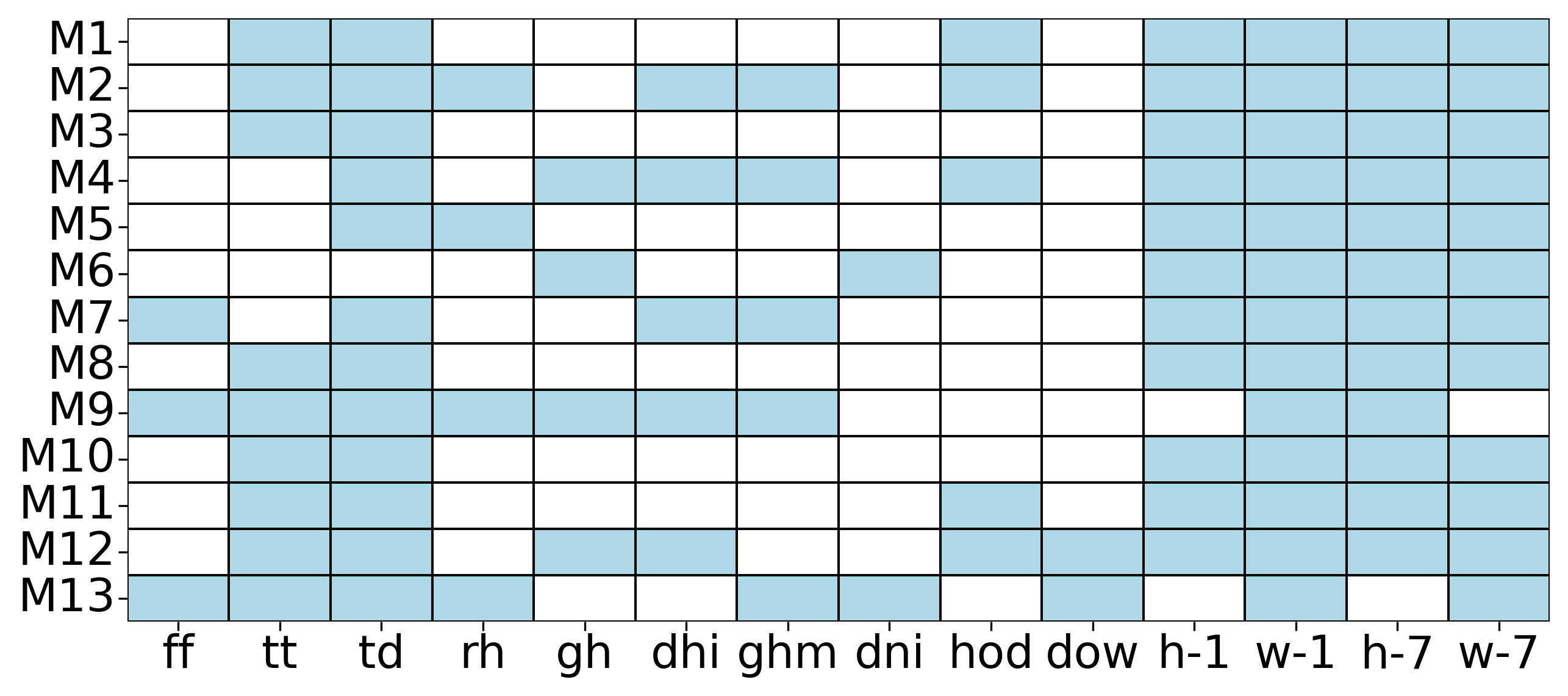}
        \label{fig:home82_heatmap}
    }
    \subfloat[]{
        \includegraphics[width=0.47\textwidth]{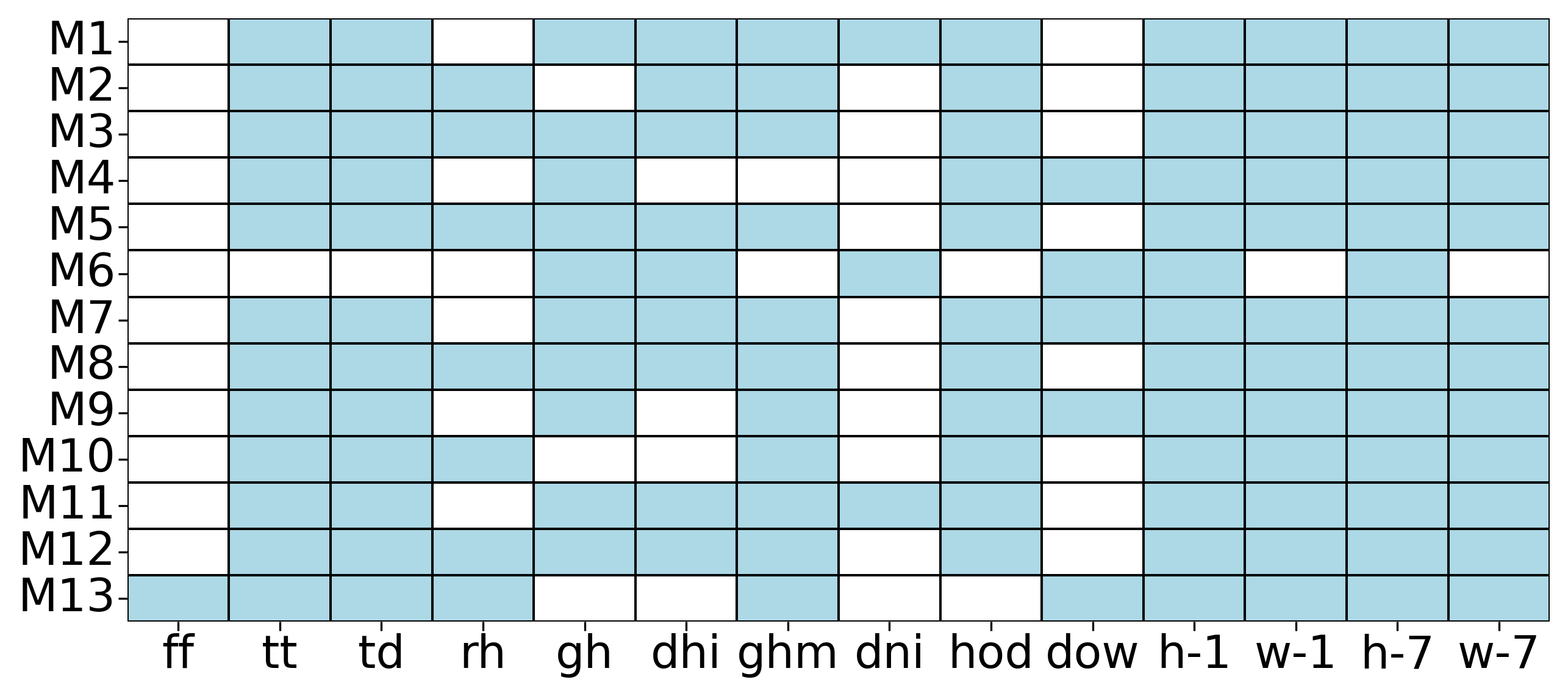}
        \label{fig:home88_heatmap}
    }\\
    \subfloat[]{
        \includegraphics[width=0.47\textwidth]{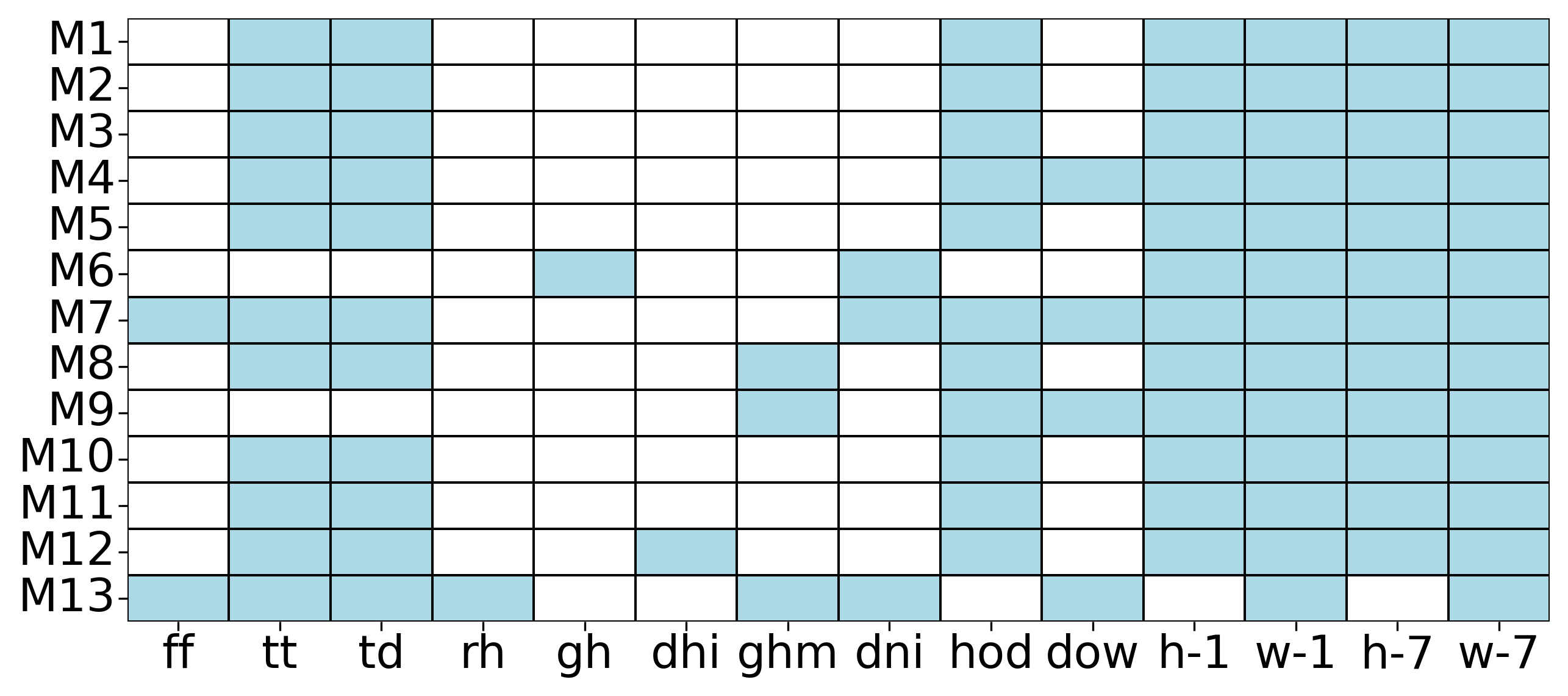}
        \label{fig:home90_heatmap}
    }
    \subfloat[]{
        \includegraphics[width=0.47\textwidth]{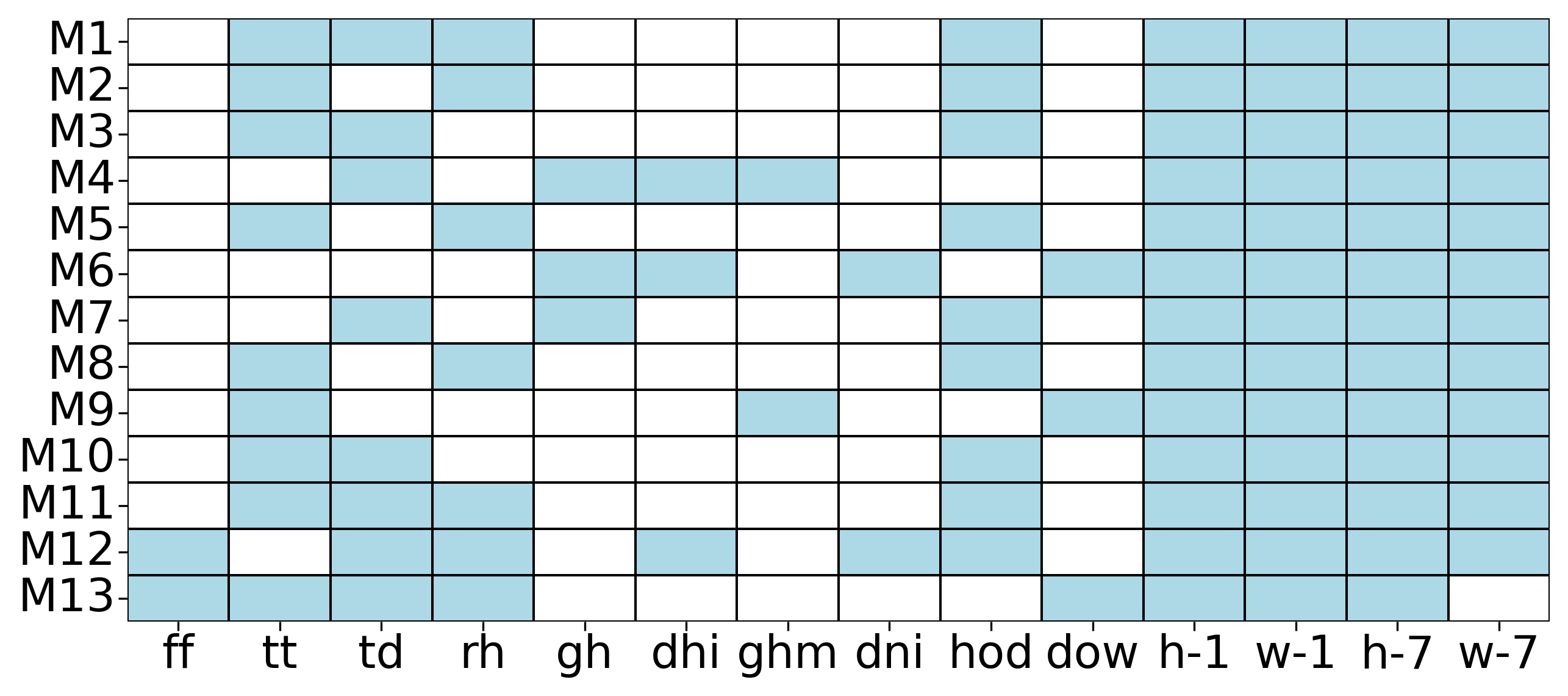}
        \label{fig:home92_heatmap}
    }\\
    \subfloat[]{
        \includegraphics[width=0.47\textwidth]{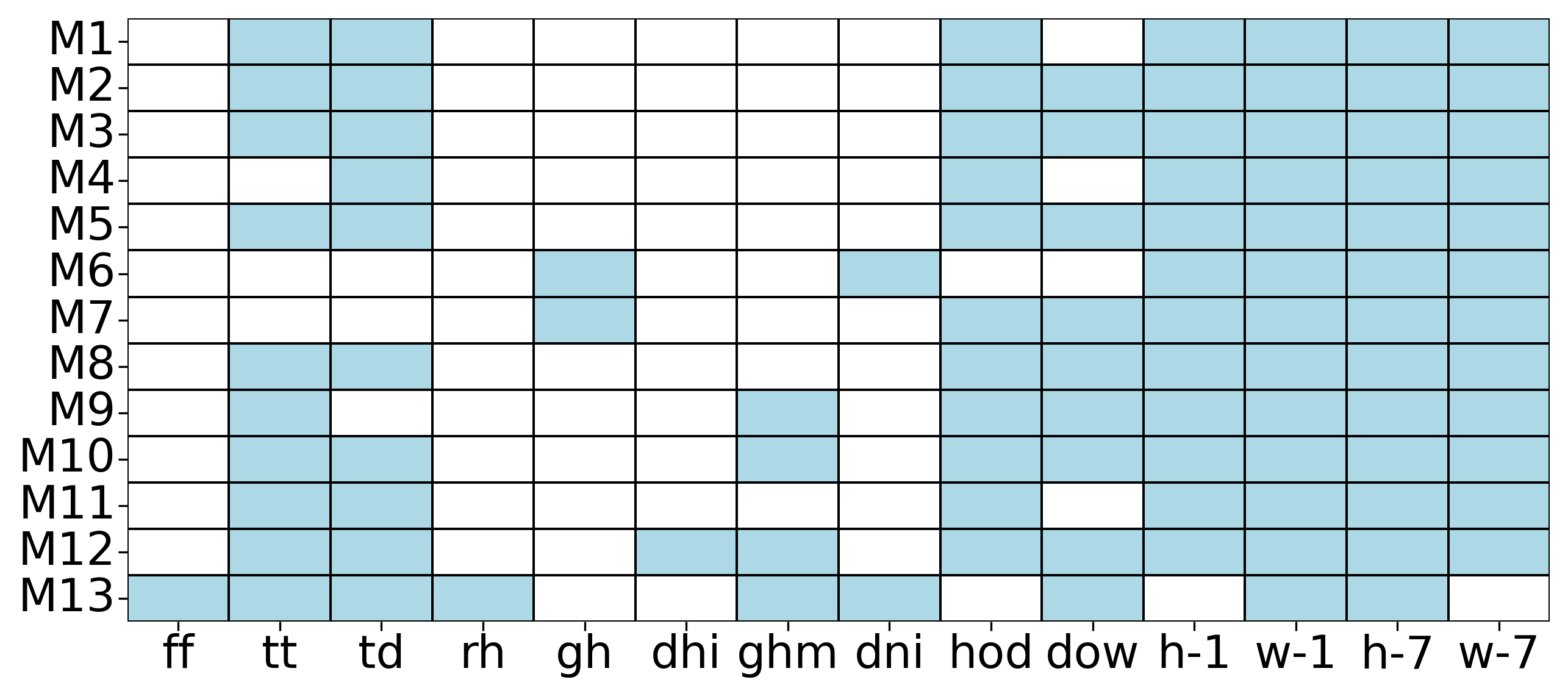}
        \label{fig:home101_heatmap}
    }
    \subfloat[]{
        \includegraphics[width=0.47\textwidth]{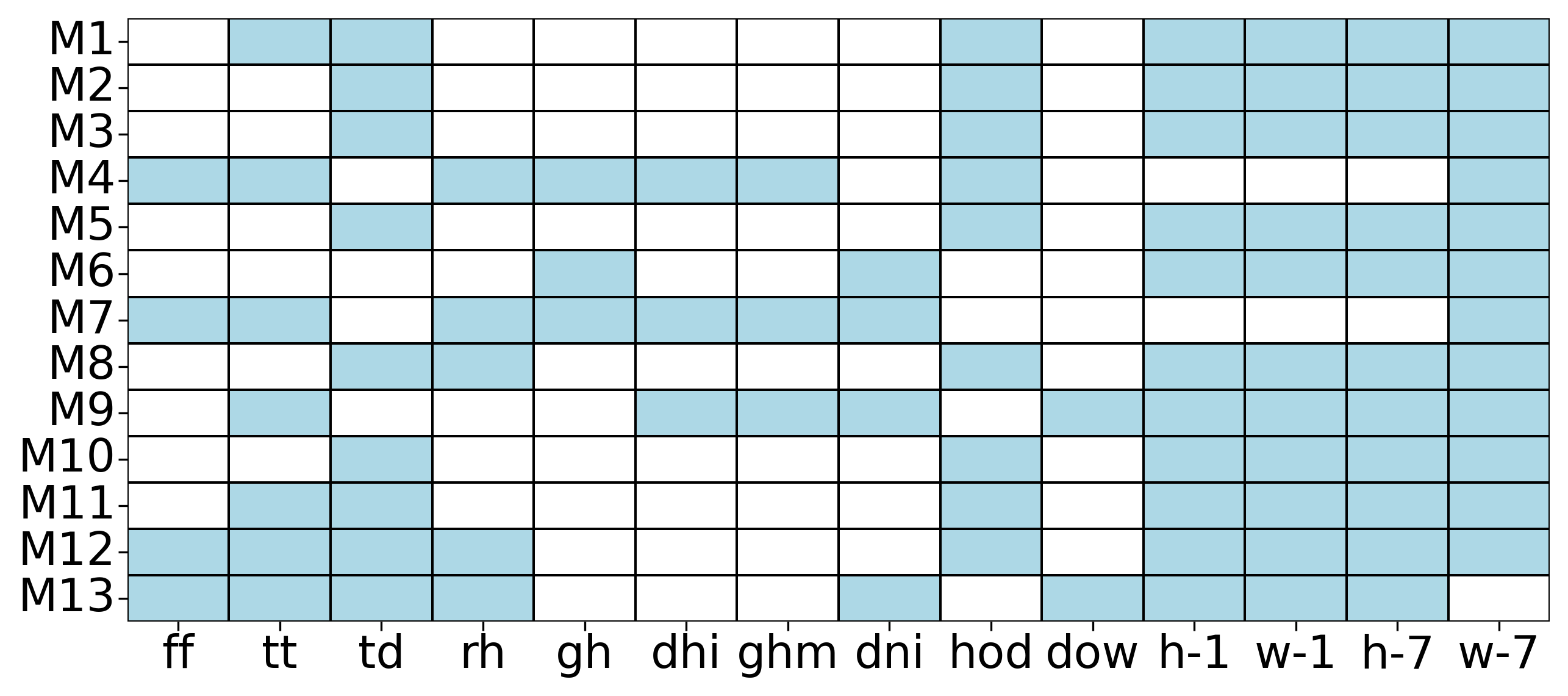}
        \label{fig:home107_heatmap}
    }\\
    \subfloat[]{
        \includegraphics[width=0.47\textwidth]{home101_heatmap}
        \label{fig:home109_heatmap}
    }
    \subfloat[]{
        \includegraphics[width=0.47\textwidth]{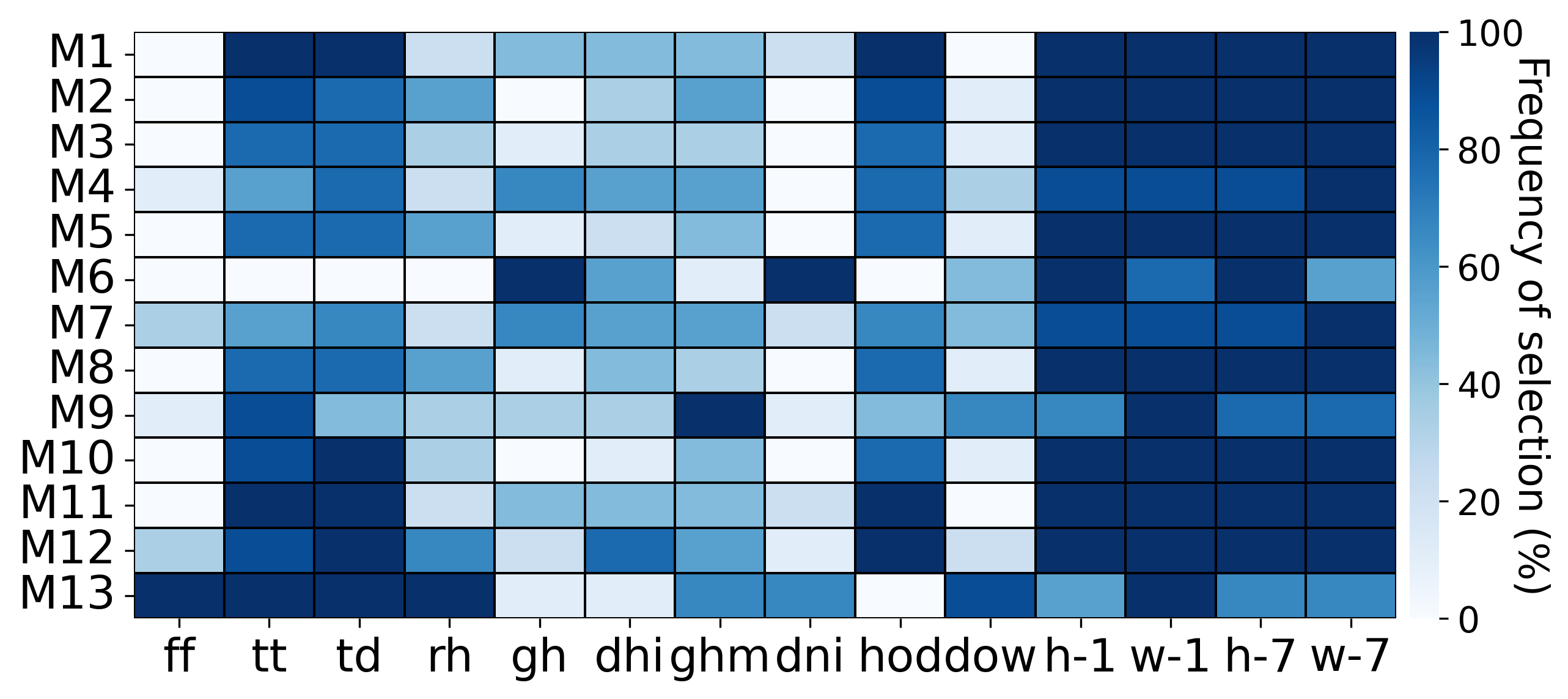}
        \label{fig:frequency_heatmap}
    }
    \caption{\textrm{Inputs (blue) for home: (a) 68, (b) 81, (c) 82, (d) 88, (e) 90, (f) 92, (g) 101, (h) 107, (i) 109 and (j) all collectively.}}
    \label{fig:heatmap_comparison}
\end{figure*}

Finally, the 95th Percentile Absolute Error (P95AE) measures the upper tail of the error distribution and reflects extreme forecast deviations:

\begin{equation}
\textrm{P95AE} = \textrm{Q}_{0.95}(|\textrm{y}_t - \hat{\textrm{y}}_t|)
\end{equation}
where $\textrm{Q}_{0.95}(\cdot)$ denotes the 95th percentile of the absolute error distribution.

\section{Results and analysis}
\label{res}

In this section, first, the selected input features for each home and hybrid forecasting model are presented. Next, the performance of the prediction models is evaluated for both DHW and heating forecasts, followed by a further discussion of the study limitations and future work. In the remainder, for simplicity, the forecasting models are renamed as: M1--PatchTCN-TST, M2--DwdAdam-ILSTM, M3--CLSTM-AR-MFFCLA, M4--PCC-GRU, M5--FTTrans-E-BL, M6--TE-BiSRU-MTL, M7--WTD-CNN-LSTM, M8--aTCN-LSTM, M9--EMD-LSTM-Markov, M10--SVM-SA, M11--GA/iPSO-ANN, M12--SSA-CNN-SVM, and M13--RF-ISSA-LSTM.

\subsection{Feature selection}
\label{feat_sel}

Depending on the hybrid forecasting model, either all or specific input features are selected through different methodologies. The most suitable inputs are determined from a set of available weather parameter forecasts and calendar parameters. Based on the available weather parameter forecasts, wind speed (ff), two-meter air temperature (tt), two-meter dew point temperature (td), relative humidity (rh), global radiation (gh), diffuse radiation (dhi), clear sky radiation (ghm), and direct normal radiation (dni) are considered as candidate input features. Regarding the calendar parameters, the hour of day (hod) and day of week (dow) are evaluated. In addition, the historical heating and DHW demand of the respective timestep of the previous day (h-1, w-1), as well as the historical demands of the respective timestep of the same day of the previous week (h-7, w-7) are considered as candidate input features. Fig.~\ref{fig:heatmap_comparison} displays the selected input features of the DHW and heat forecasting models for each home.

The M1- and M11-models select the input features based on the combination of scores collected by Mutual Information (MI), Pearson and Spearman correlations. The parameters tt, td, hod, h-1, h-7, w-1 and w-7 are selected at the models of all households. The parameters dhi, gh, ghm and dni are selected at 3 households, while rh and dni are used at two buildings. The remaining factors ff and dow are not considered at any household.

The M2-model applies the cross-correlation for the extraction of the top 6 input features with the highest correlation degree for the forecasting models. While ff, gh and dni are not selected at any home, h-1, h-7, w-1 and w-7 are used by all buildings. In addition, tt, td and hod are used at 8 households. The remaining parameters rh, dhi, ghm and dow have variable correlation degrees among the households, therefore, there is no clear trend about their influence. 

The M3-model applies the MFFCLA for feature extraction by selecting the top 6 input features with the highest correlation degree. The historical loads h-1, h-7, w-1 and w-7 are selected at all households. Besides them, td shows comparable correlation degree, since it is used at the model of 8 buildings, followed by tt and hod, which are selected at 7 households. The parameters rh, gh, dhi, ghm and dow show low correlation with the loads, while ff and dni are not used at any building. 

As for the M4-model, the top 6 features with the highest Pearson correlation are selected as input features. Particularly, the historical dhw load w-7 is used at all households, while td, h-1, h-7 and w-1 are selected by 8 models. Lower correlation is observed for tt, gh, dhi, ghm and hod, as they are considered at 5-7 buildings, while ff, rh and dow are used at 1-2 models. Finally, dni is not selected at any building.

The M5-model applies the WTLCC for identifying the top six influential factors for the forecasting models of each home. The historical loads h-1, h-7, w-1 and w-7 present the highest correlation, as they are selected at all households. Comparable correlation is observed for td, tt and hod, which are used at 7-8 buildings. In addition, rh, dhi, ghm are considered by 3-5 models, while gh and dow show the lowest selection rate of one and two households, respectively. Moreover, the parameters ff and dni are not used at any building.  

As for the M6-model, the top 6 input features are selected using a copula-based feature extraction method. From the analysis, the weather factors gh, dni and the historical loads h-1, h-7 are selected at all buildings. Comparable correlation is observed for the historical load w-1 that is considered at 8 households. Besides them, the historical load w-7 is used at 6 buildings. The parameters dhi, dow and ghm present lower correlation with usage at 1-4 households. The remaining factors ff, tt, td, rh and hod are not used at any home.  

The M7-model selects the top 6 input features with the highest absolute average Pearson correlation. The historical loads show the highest influence, as w-7 is selected at 9 households, and h-1, h-7 and w-1 are used at 8 buildings. Among the other parameters, gh is considered at 7 households, while td and hod are used at 6 buildings. Lower correlation is observed for the parameters tt, dhi and ghm with usage at 5 households. The remaining parameters ff, rh, dni, and dow present the lowest impact with selection at 2-4 buildings. 

The M8-model calculates the absolute Pearson correlation coefficient and the top 6 features are selected. The historical loads h-1, h-7, w-1 and w-7 are selected at all households. The parameters tt, td and hod show comparable correlation with usage at 7-8 buildings. Lower correlation is observed for rh with selection at 5 households. Besides them, gh, dhi, ghm and dow present low influence, as they are considered at 1-3 buildings. Finally, ff and dni are not used at any household. 

The M9-model applies first the absolute Pearson correlation and GBDT, and their scores are combined for ranking and selecting the top 6 input features. The historical load w-1 and ghm are used at all houses, while high correlation is also computed for tt and h-7 that is considered at 8 buildings. The other historical loads h-1 and w-7 are selected at 7 households. Moreover, dow is used by 6 buildings, while td, rh, gh, dhi and hod show lower influence with use at 3-4 households. Finally, ff and dni present the lowest correlation, since they are selected at a single building. 

As regards the M10-model, the top 6 input features are selected by ranking them based on their absolute average Pearson correlation. The historical loads h-1, h-7, w-1, w-7 and the weather parameter td are used at all households, while hod and tt are selected at 7 and 8 buildings, respectively. Lower correlation is observed for rh and ghm that are selected at 3 and 5 households, correspondingly. Besides them, dhi is used at a building, and the remaining parameters ff, gh and dni are not considered at any household.

\begin{figure*}[t]
    \centering
    \subfloat[]{
        \includegraphics[width=0.47\textwidth]{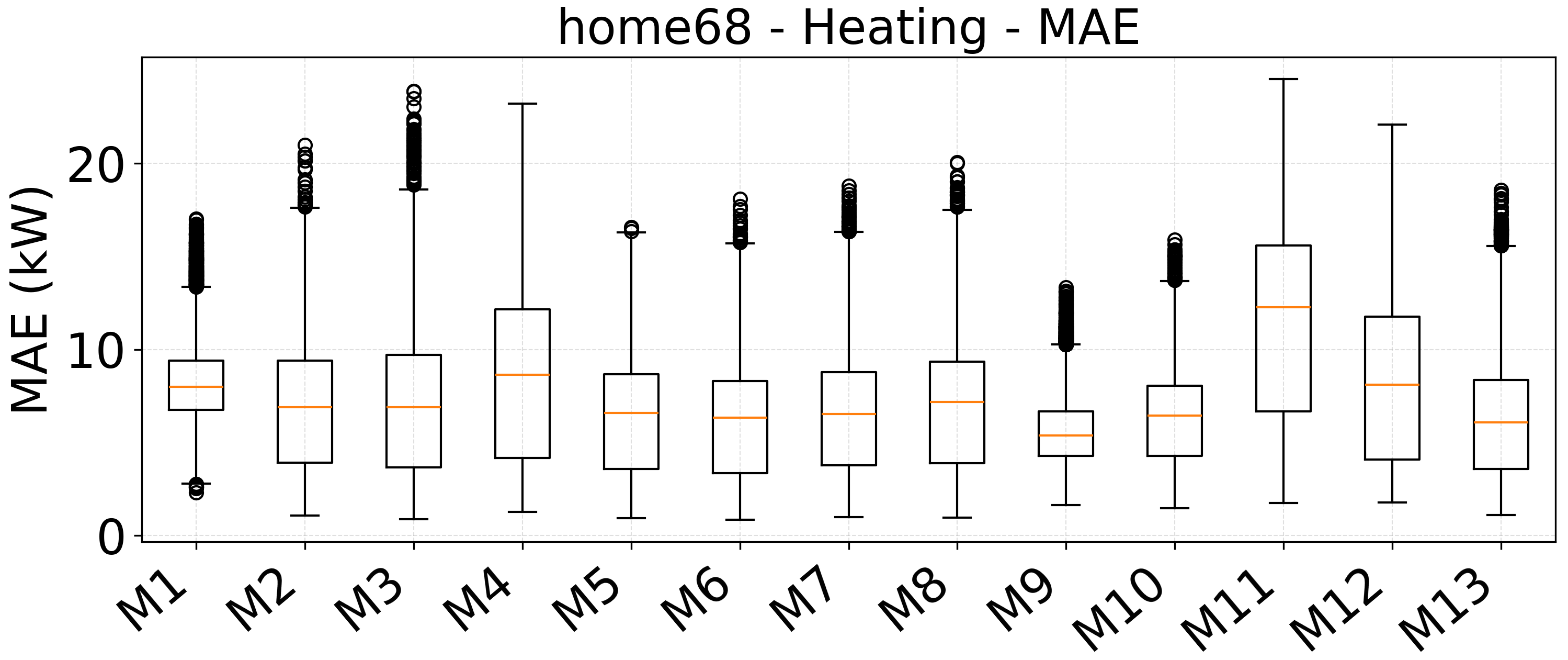}
        \label{fig:home68_MAE_Heating}
    }
    \subfloat[]{
        \includegraphics[width=0.47\textwidth]{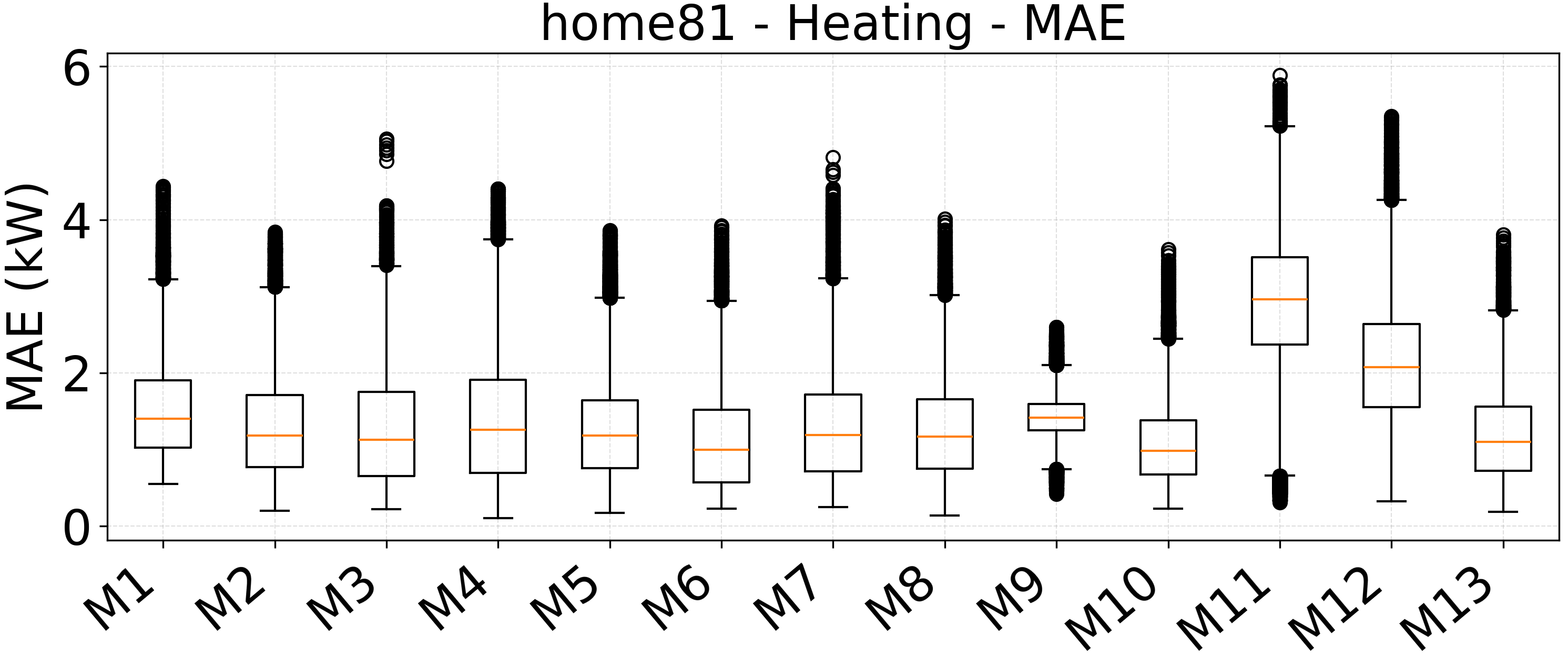}
        \label{fig:home81_MAE_Heating}
    }\\
    \subfloat[]{
        \includegraphics[width=0.47\textwidth]{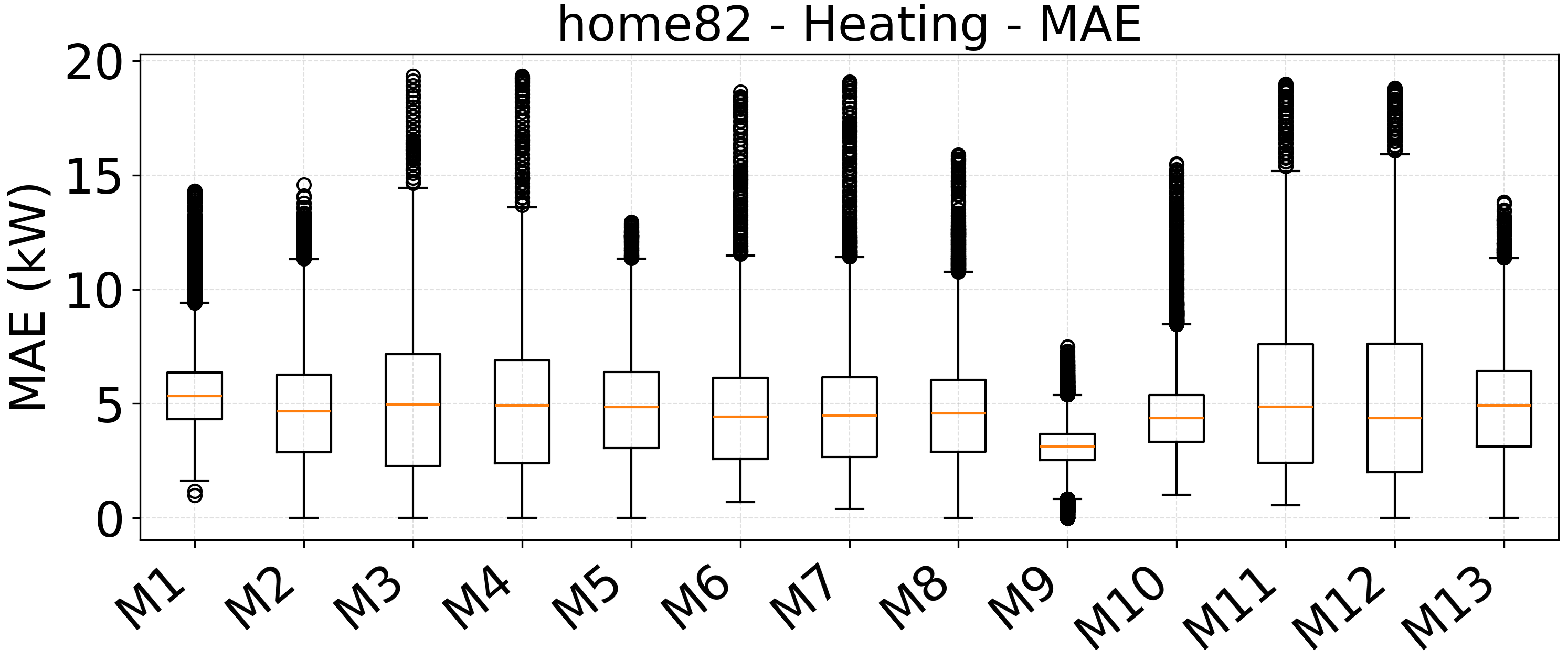}
        \label{fig:home82_MAE_Heating}
    }
    \subfloat[]{
        \includegraphics[width=0.47\textwidth]{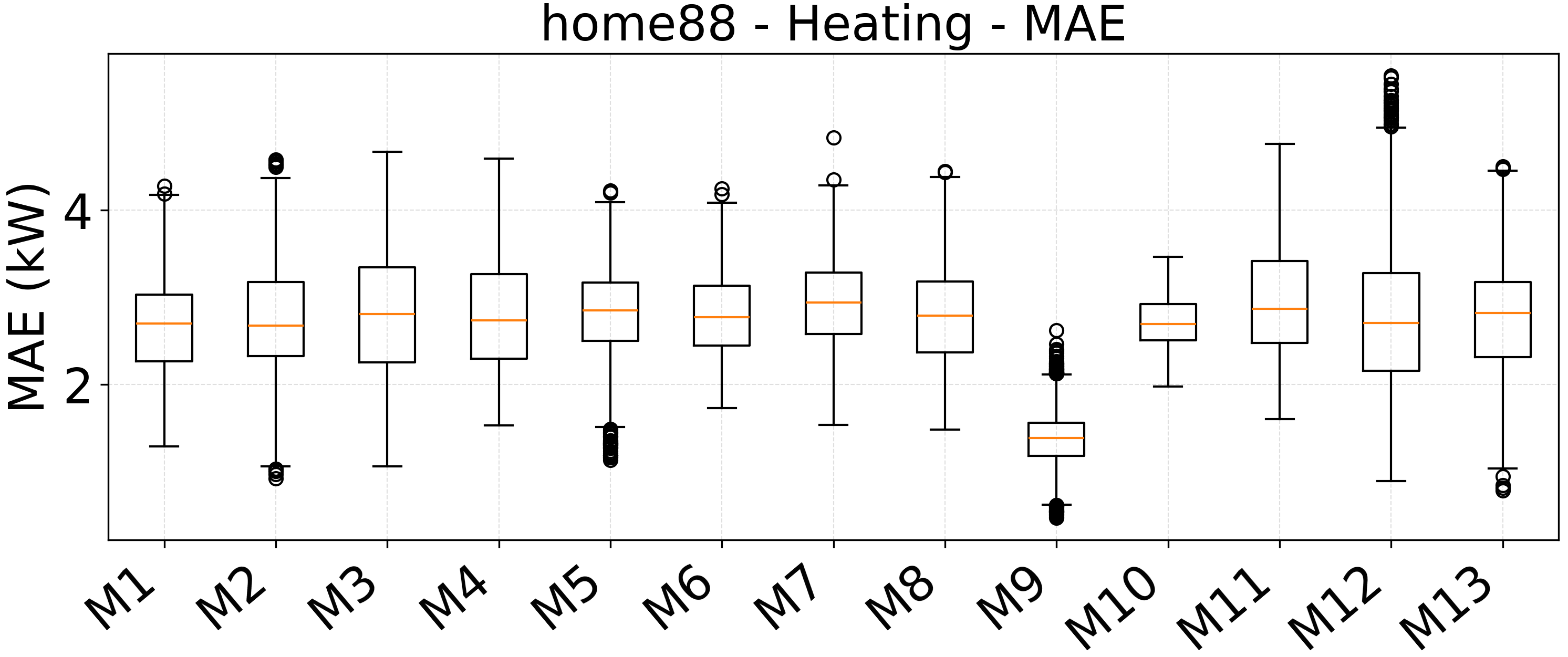}
        \label{fig:home88_MAE_Heating}
    }\\
    \subfloat[]{
        \includegraphics[width=0.47\textwidth]{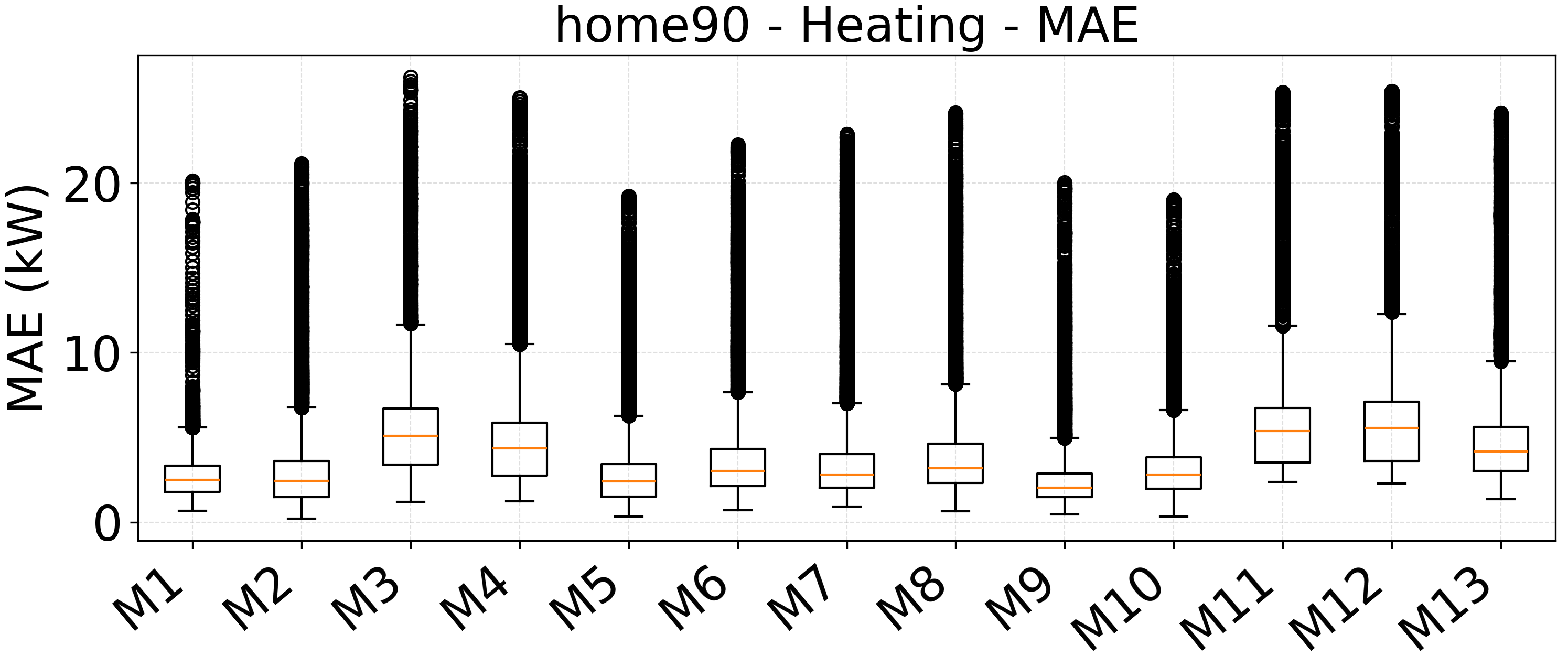}
        \label{fig:home90_MAE_Heating}
    }
    \subfloat[]{
        \includegraphics[width=0.47\textwidth]{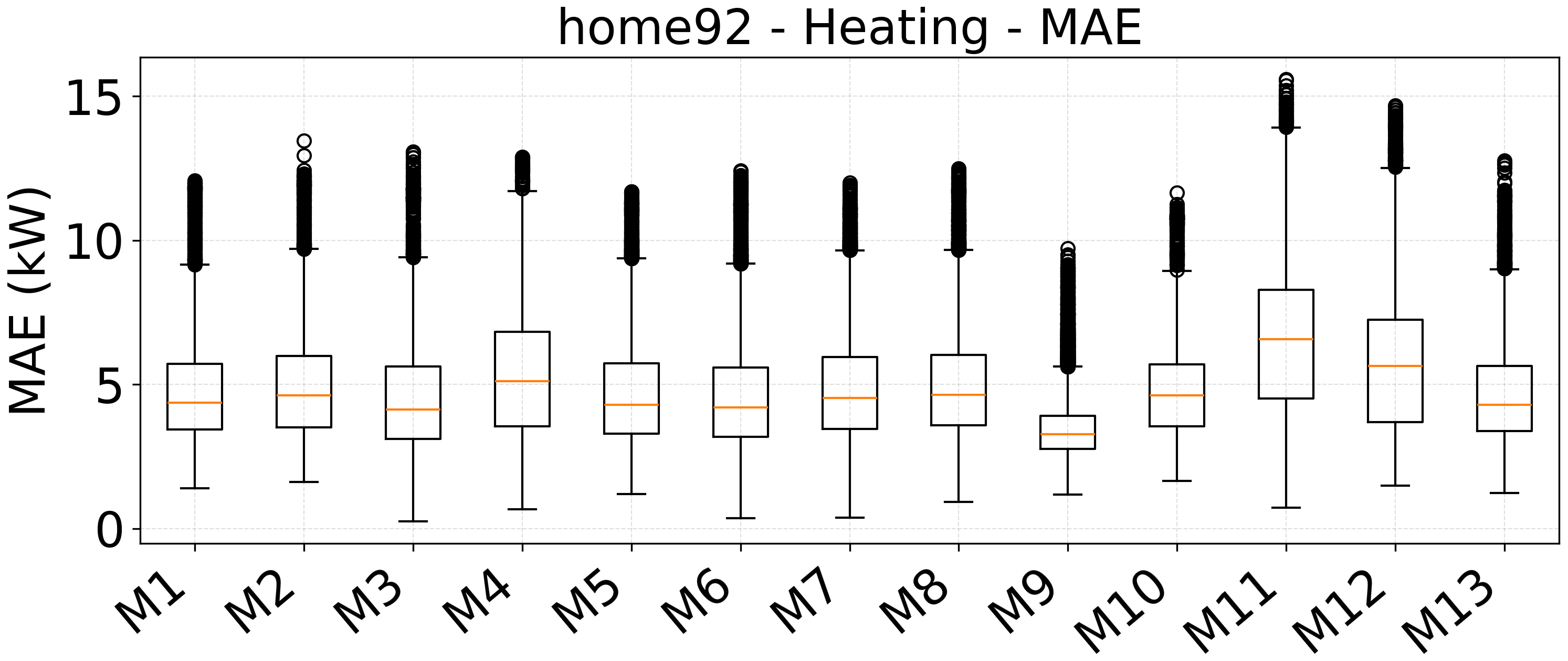}
        \label{fig:home92_MAE_Heating}
    }\\
    \subfloat[]{
        \includegraphics[width=0.47\textwidth]{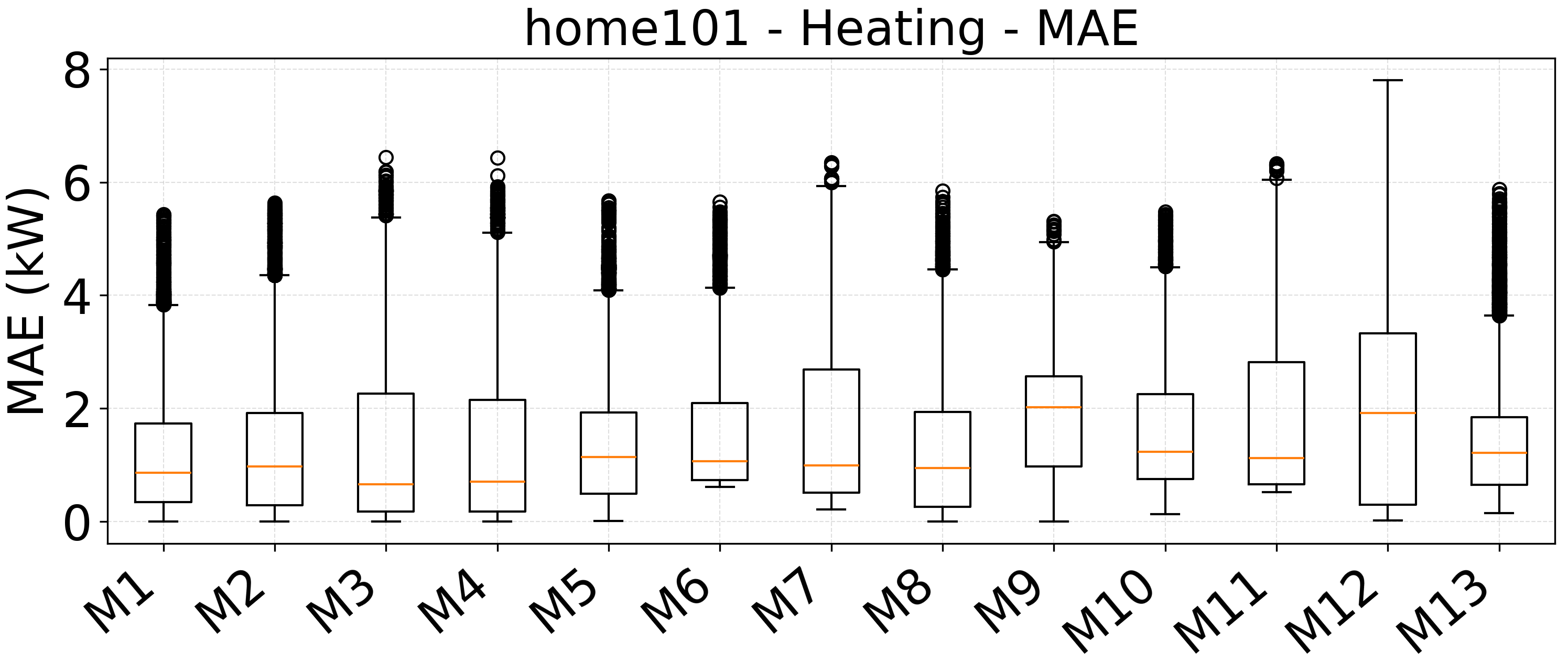}
        \label{fig:home101_MAE_Heating}
    }
    \subfloat[]{
        \includegraphics[width=0.47\textwidth]{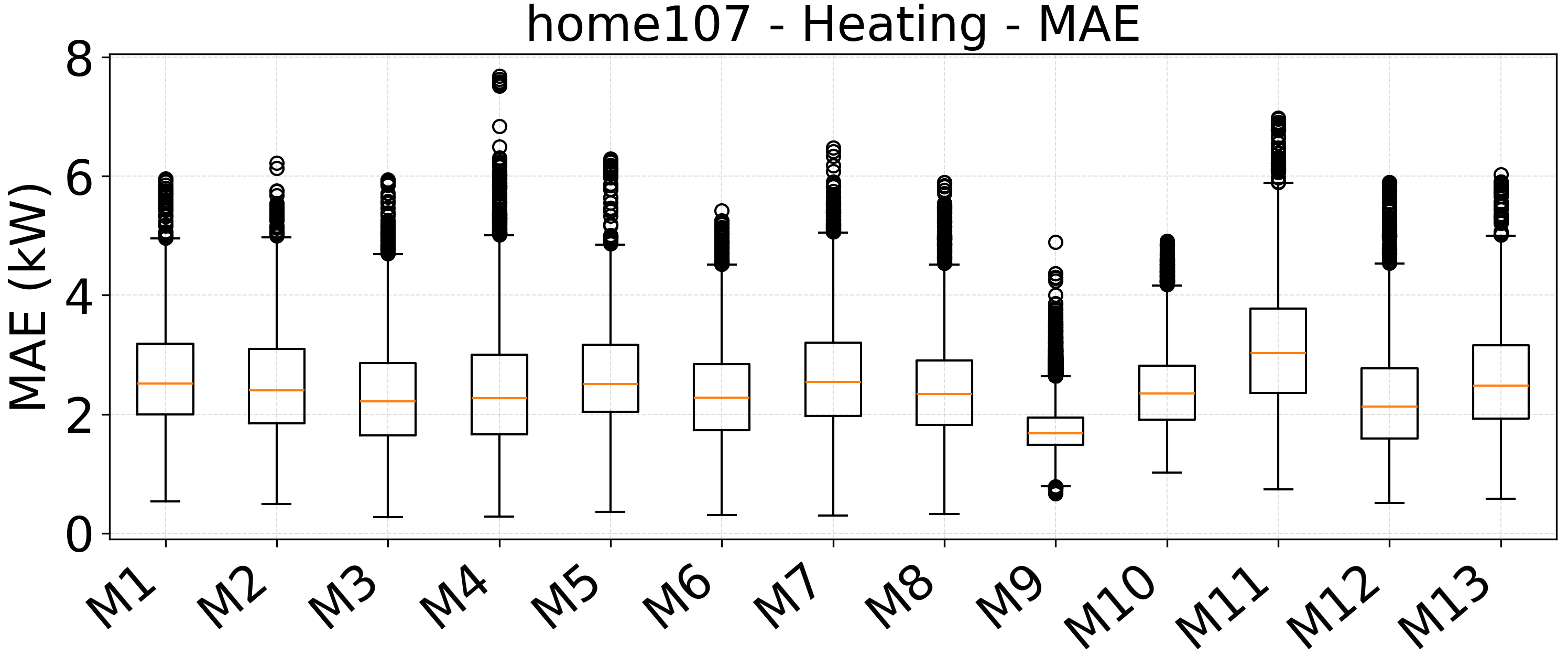}
        \label{fig:home107_MAE_Heating}
    }\\
    \subfloat[]{
        \includegraphics[width=0.47\textwidth]{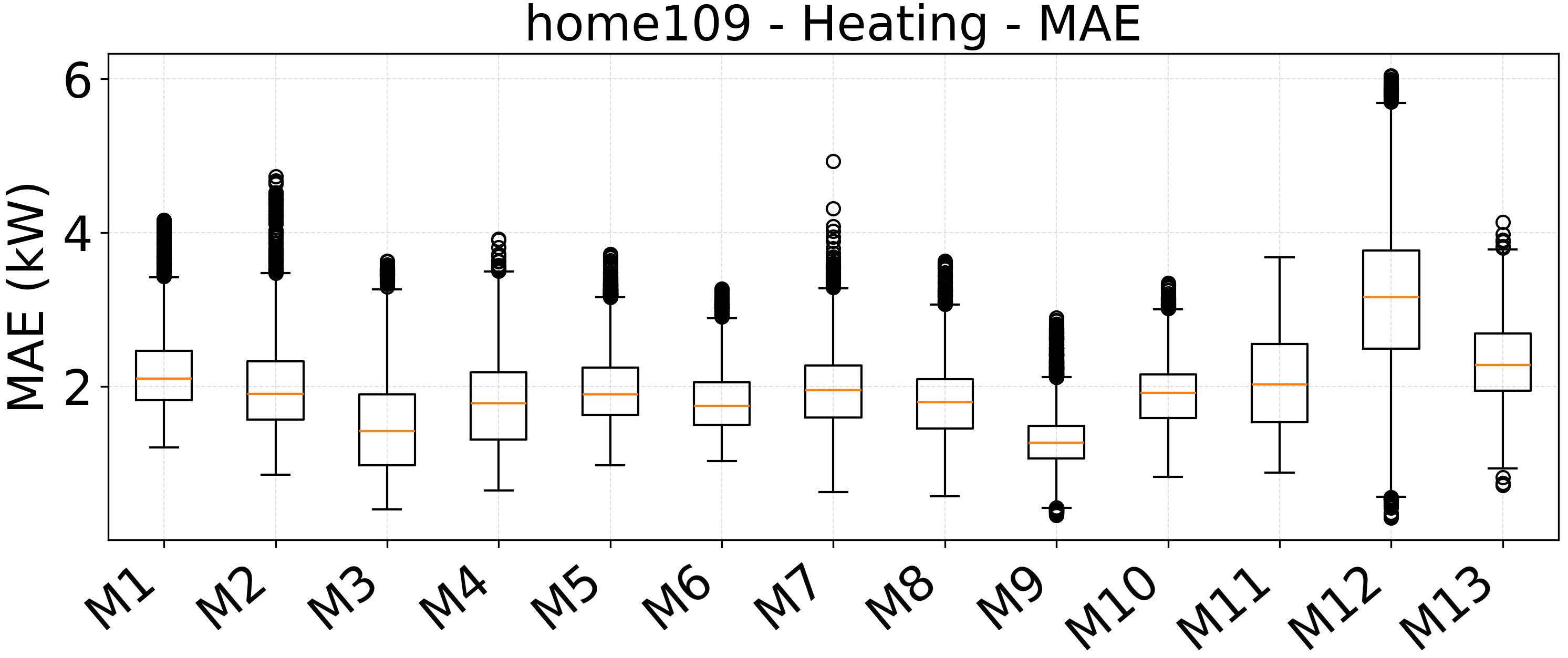}
        \label{fig:home109_MAE_Heating}
    }
    \caption{\textrm{MAE boxplots from the application of all heat forecasting models at home: (a) 68, (b) 81, (c) 82, (d) 88, (e) 90, (f) 92, (g) 101, (h) 107, and (i) 109.}}
    \label{fig:MAE_heating_comparison}
\end{figure*}

\begin{figure*}[t]
    \centering
    \subfloat[]{
        \includegraphics[width=0.47\textwidth]{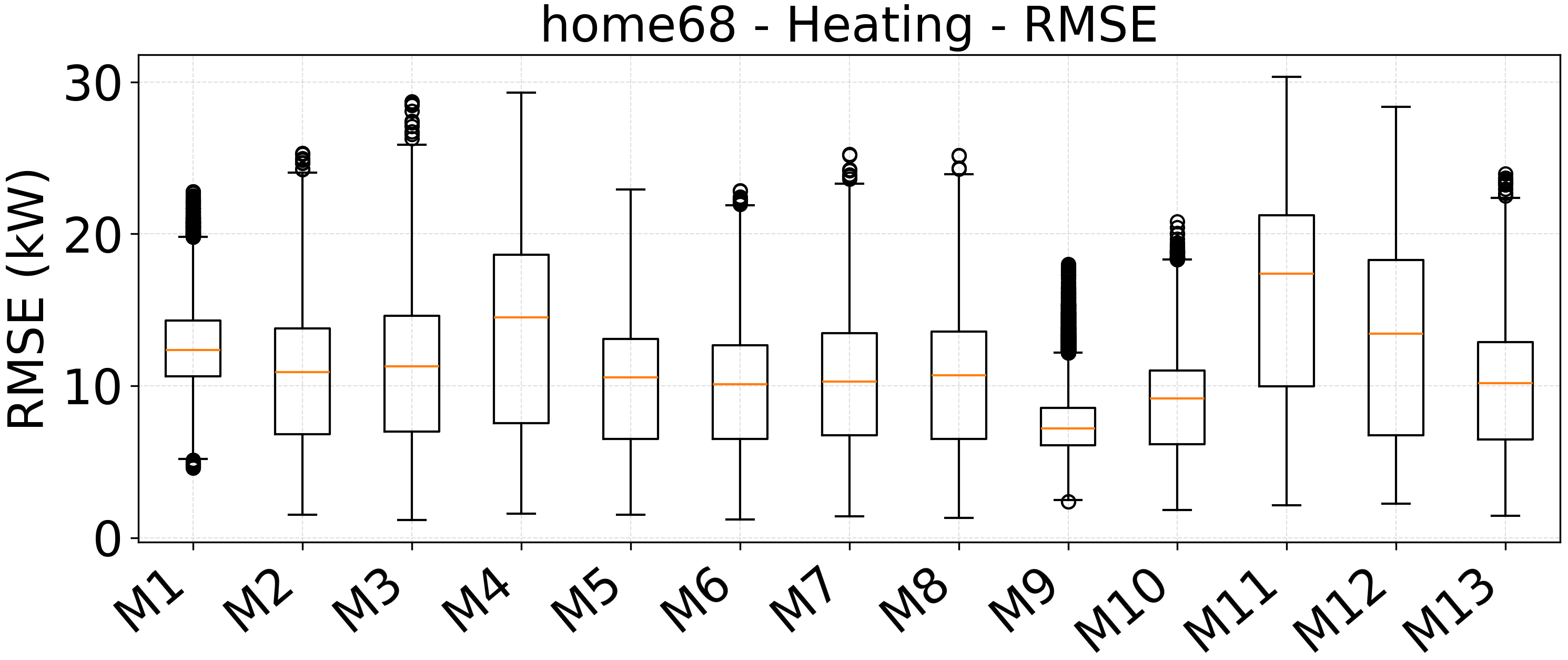}
        \label{fig:home68_RMSE_Heating}
    }
    \subfloat[]{
        \includegraphics[width=0.47\textwidth]{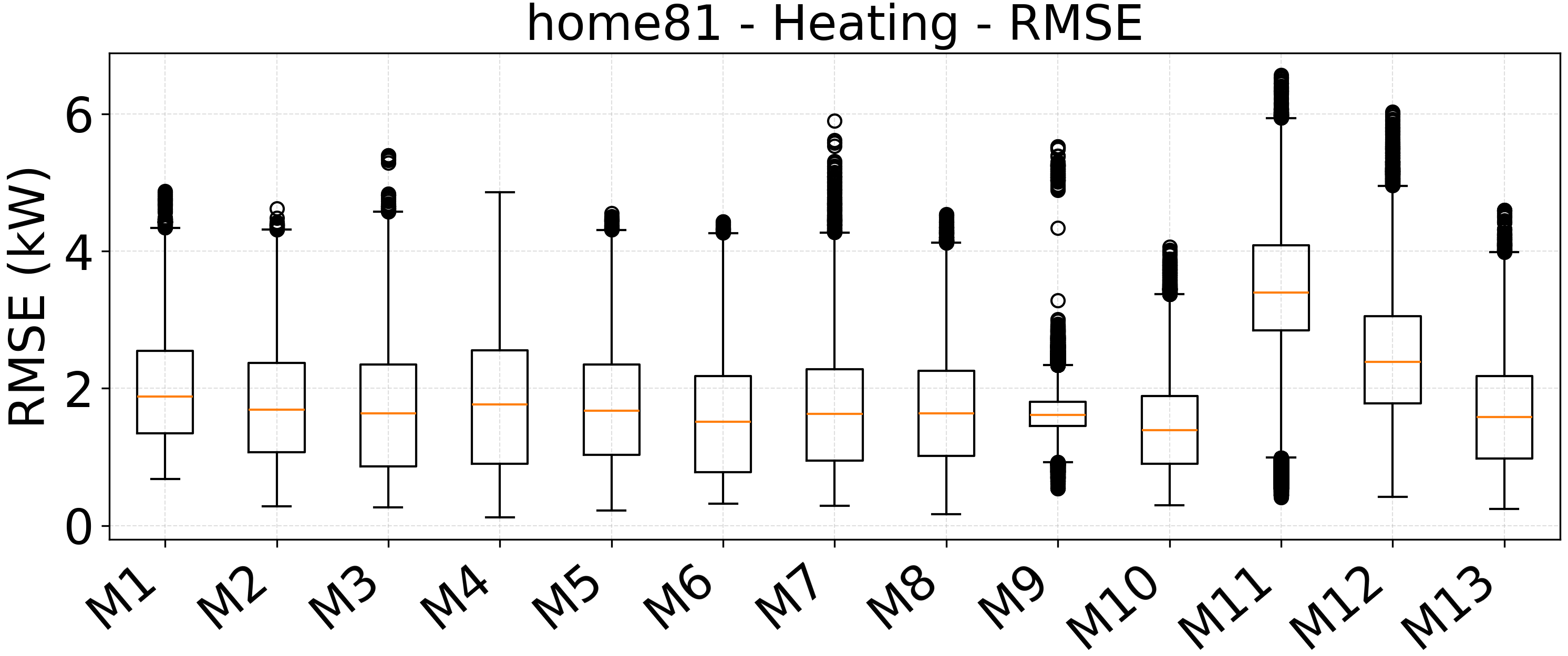}
        \label{fig:home81_RMSE_Heating}
    }\\
    \subfloat[]{
        \includegraphics[width=0.47\textwidth]{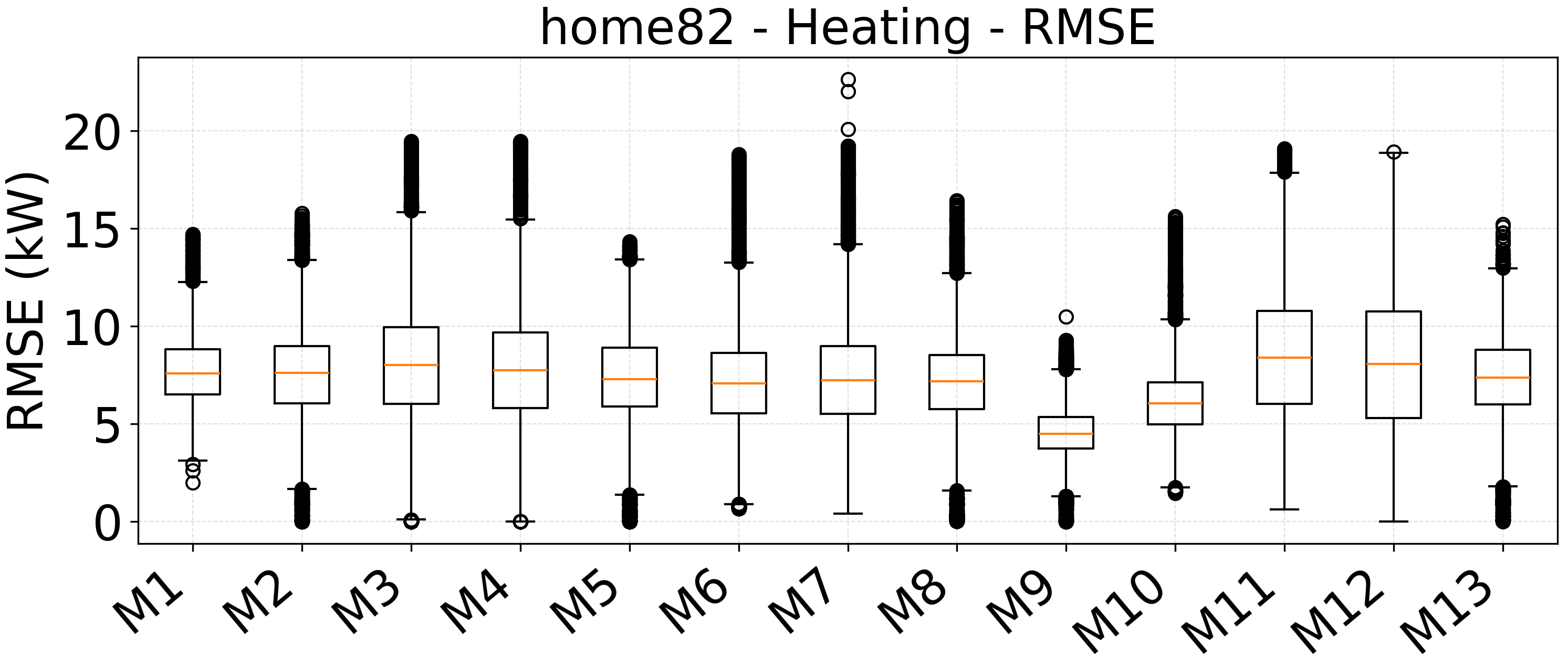}
        \label{fig:home82_RMSE_Heating}
    }
    \subfloat[]{
        \includegraphics[width=0.47\textwidth]{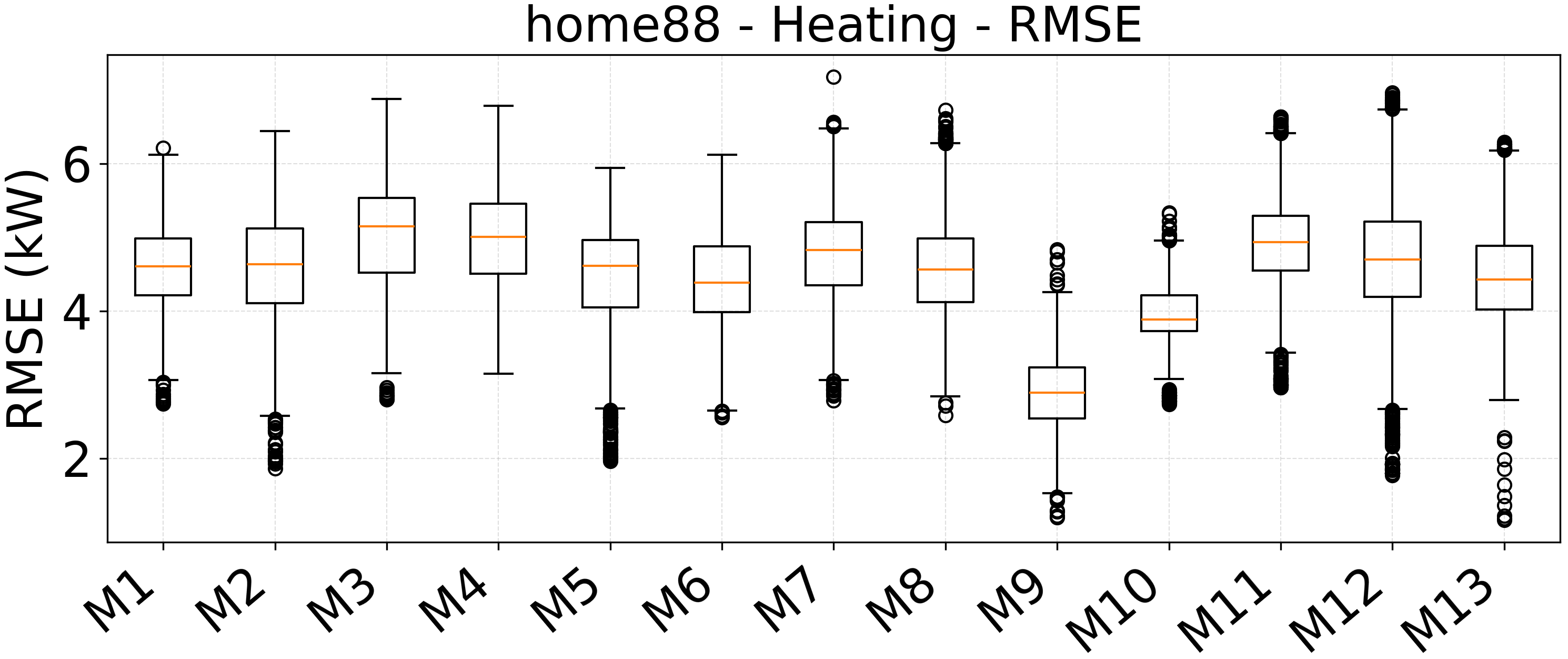}
        \label{fig:home88_RMSE_Heating}
    }\\
    \subfloat[]{
        \includegraphics[width=0.47\textwidth]{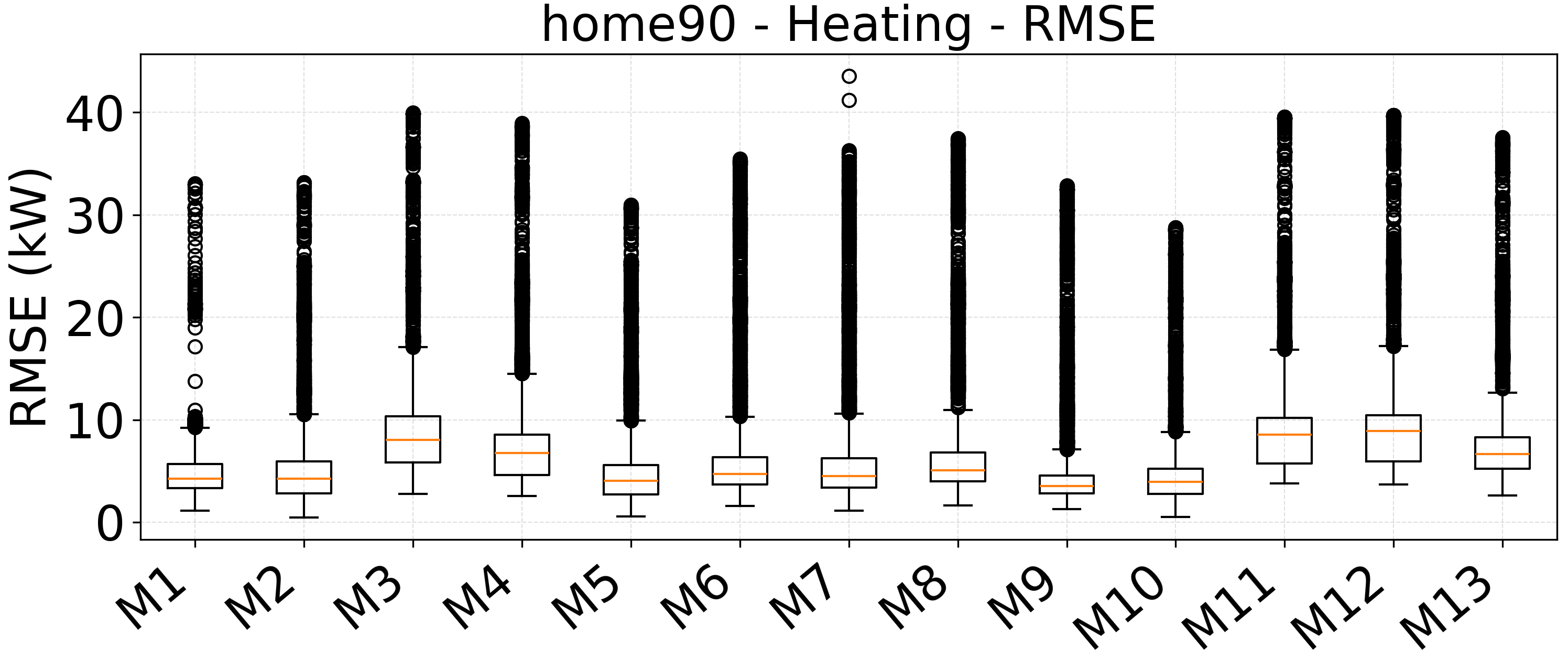}
        \label{fig:home90_RMSE_Heating}
    }
    \subfloat[]{
        \includegraphics[width=0.47\textwidth]{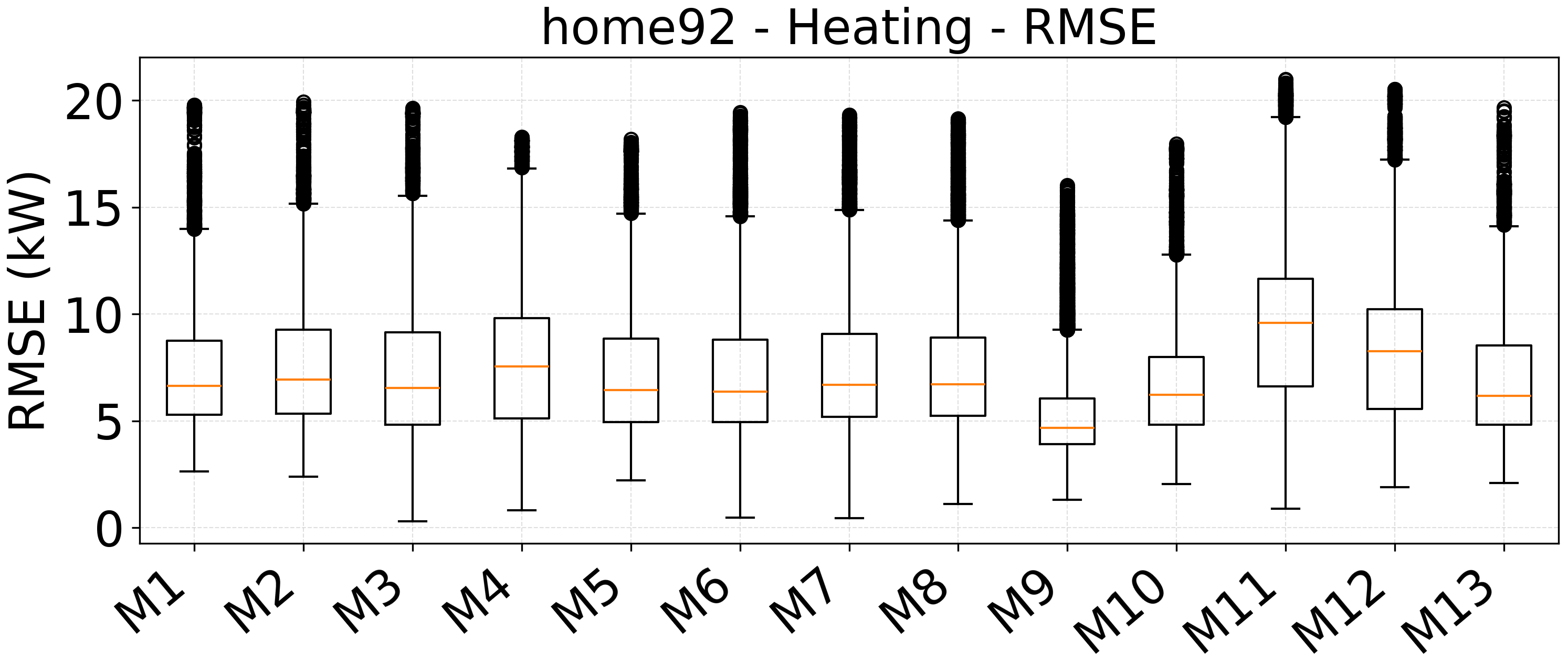}
        \label{fig:home92_RMSE_Heating}
    }\\
    \subfloat[]{
        \includegraphics[width=0.47\textwidth]{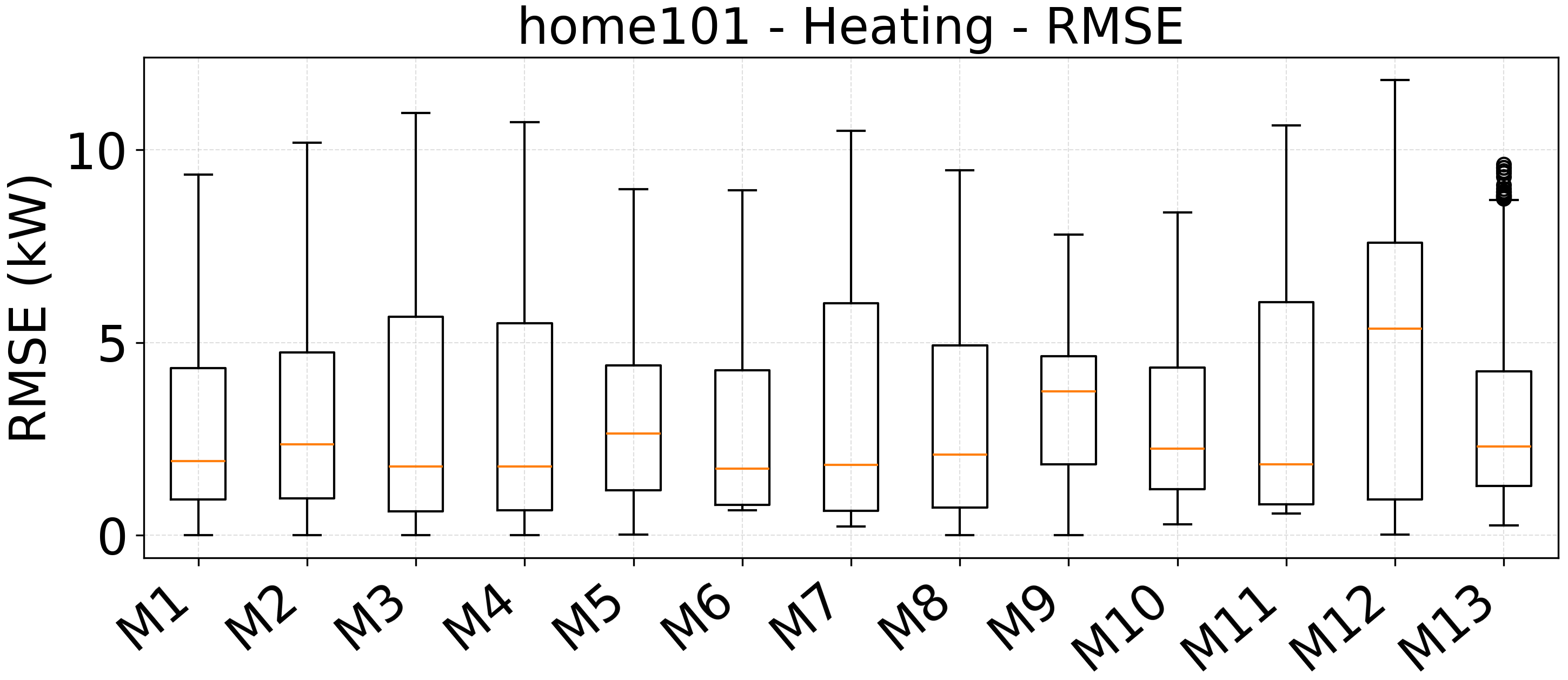}
        \label{fig:home101_RMSE_Heating}
    }
    \subfloat[]{
        \includegraphics[width=0.47\textwidth]{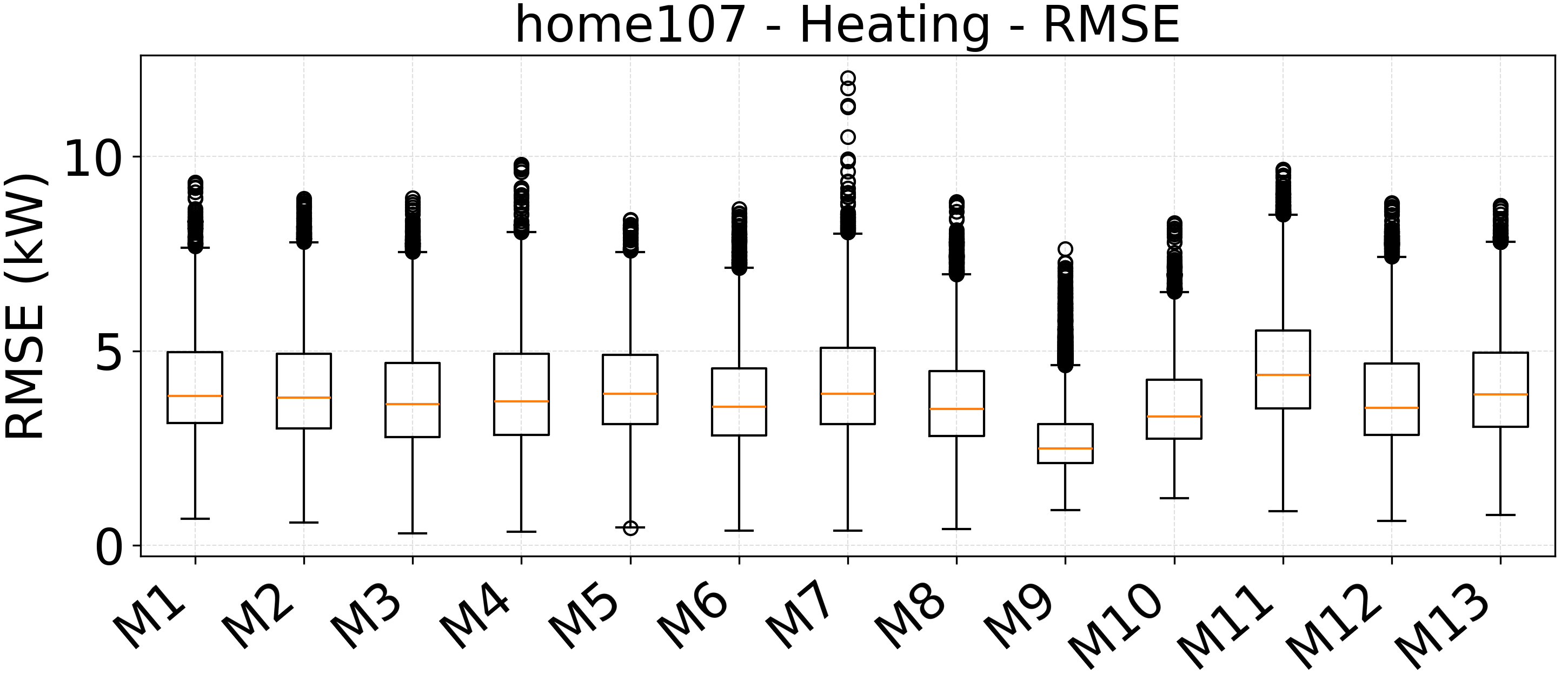}
        \label{fig:home107_RMSE_Heating}
    }\\
    \subfloat[]{
        \includegraphics[width=0.47\textwidth]{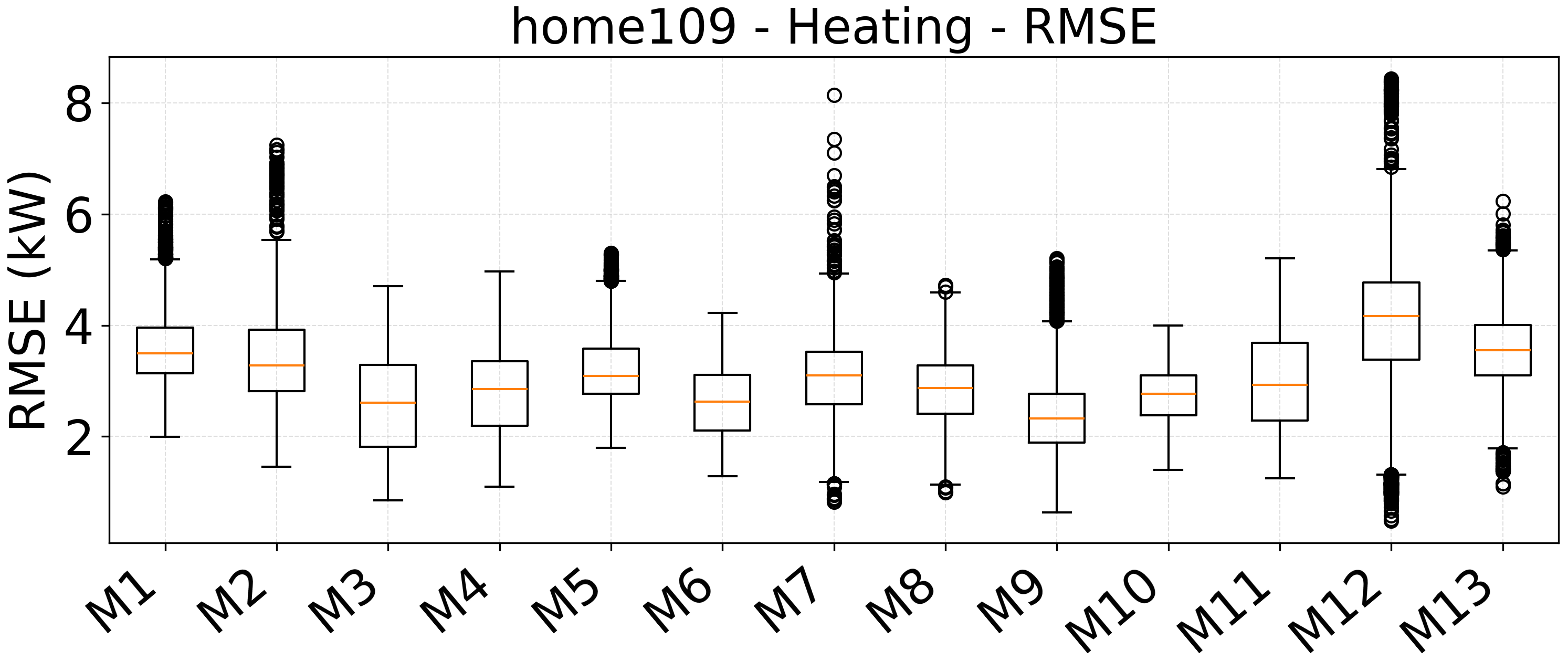}
        \label{fig:home109_RMSE_Heating}
    }
    \caption{\textrm{RMSE boxplots from the application of all heat forecasting models at home: (a) 68, (b) 81, (c) 82, (d) 88, (e) 90, (f) 92, (g) 101, (h) 107, and (i) 109.}}
    \label{fig:RMSE_heating_comparison}
\end{figure*}

\begin{figure*}[t]
    \centering
    \subfloat[]{
        \includegraphics[width=0.47\textwidth]{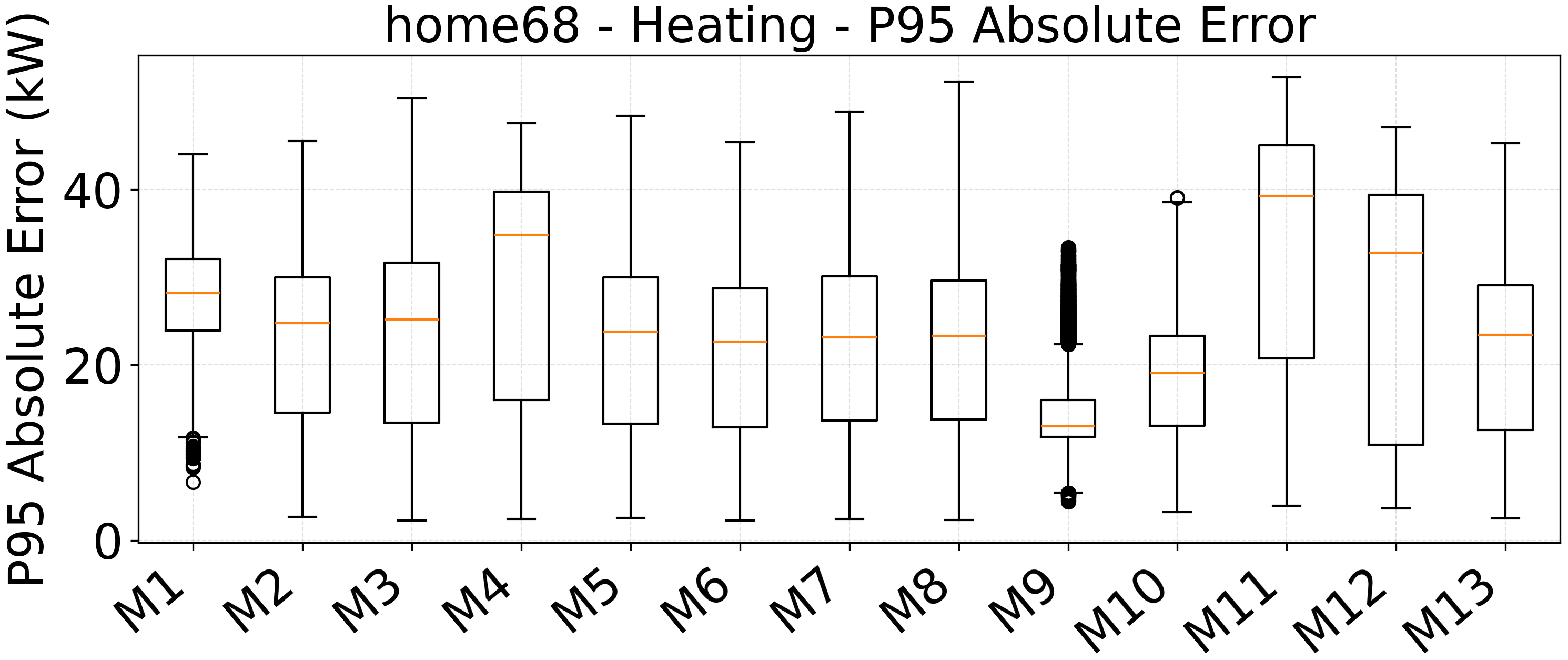}
        \label{fig:home68_P95_absolute_error_Heating}
    }
    \subfloat[]{
        \includegraphics[width=0.47\textwidth]{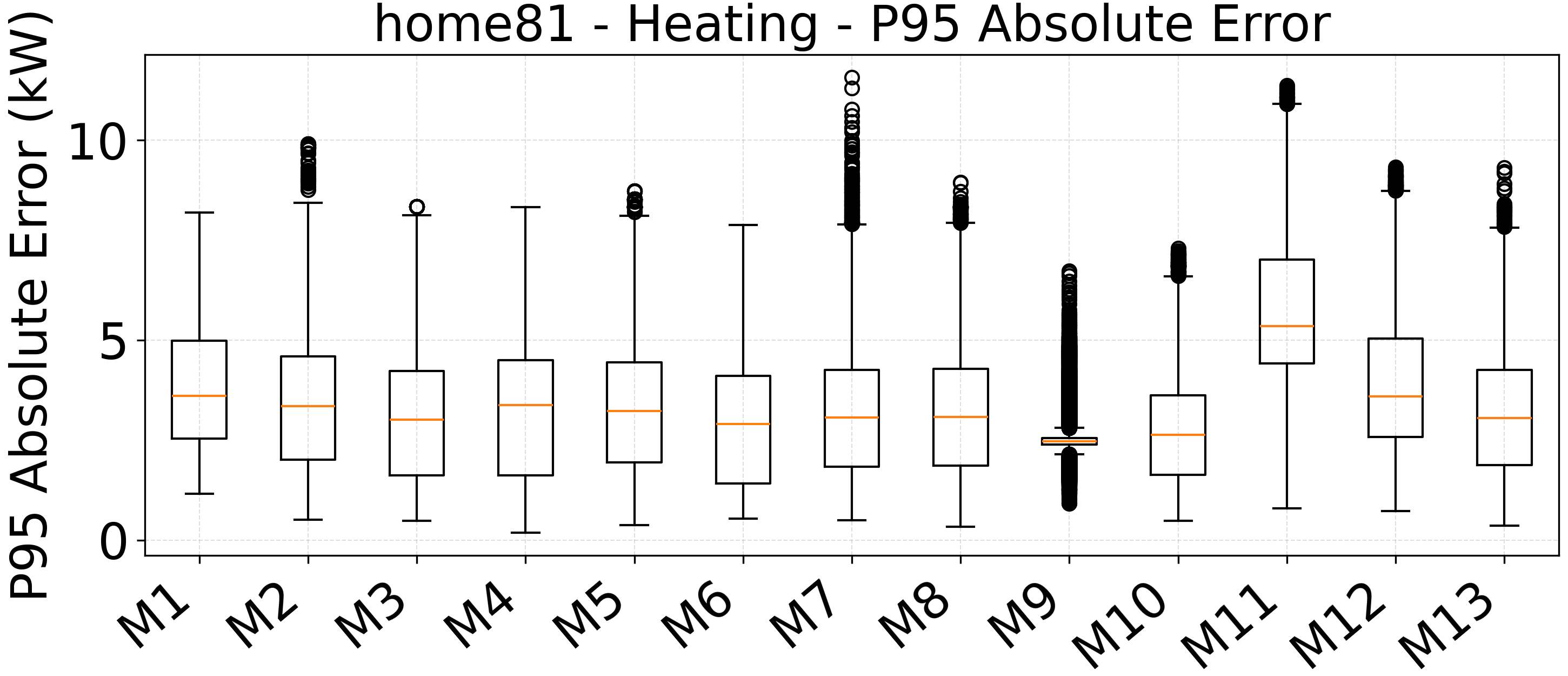}
        \label{fig:home81_P95_absolute_error_Heating}
    }\\
    \subfloat[]{
        \includegraphics[width=0.47\textwidth]{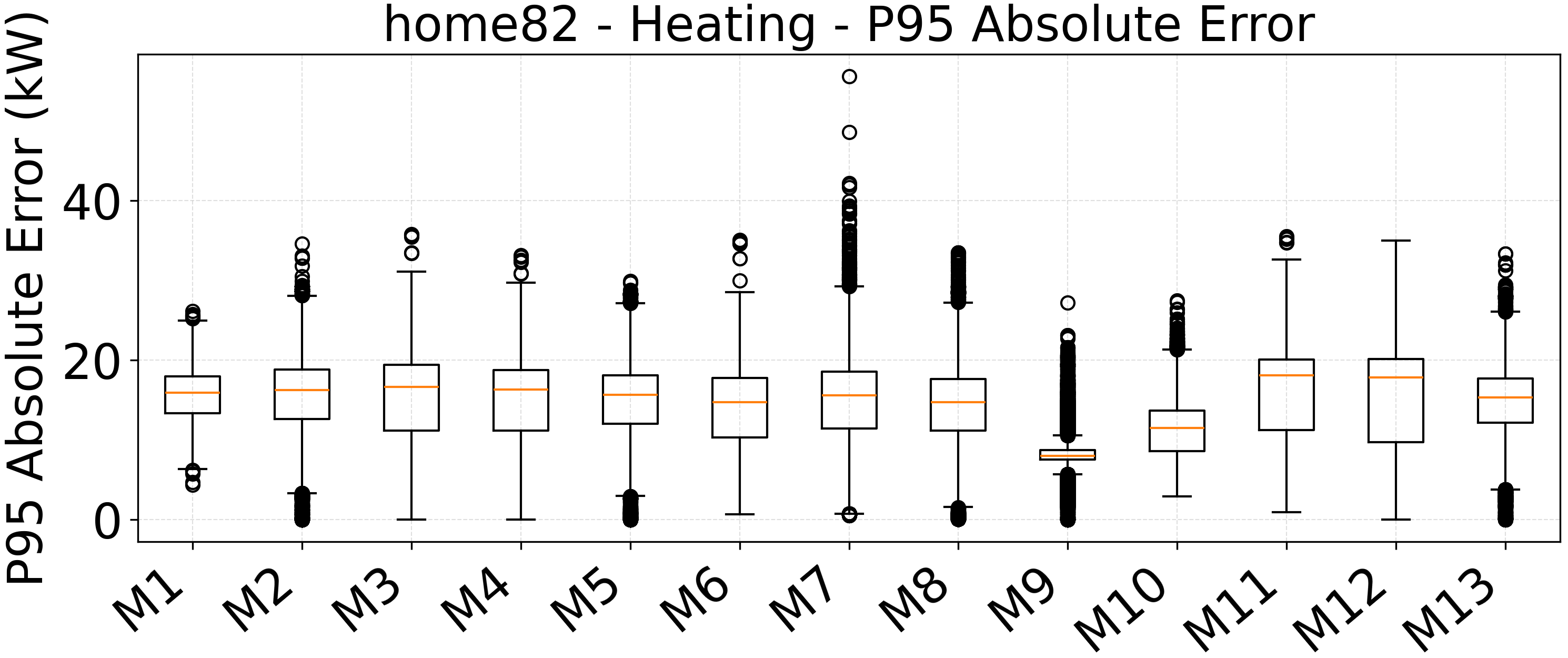}
        \label{fig:home82_P95_absolute_error_Heating}
    }
    \subfloat[]{
        \includegraphics[width=0.47\textwidth]{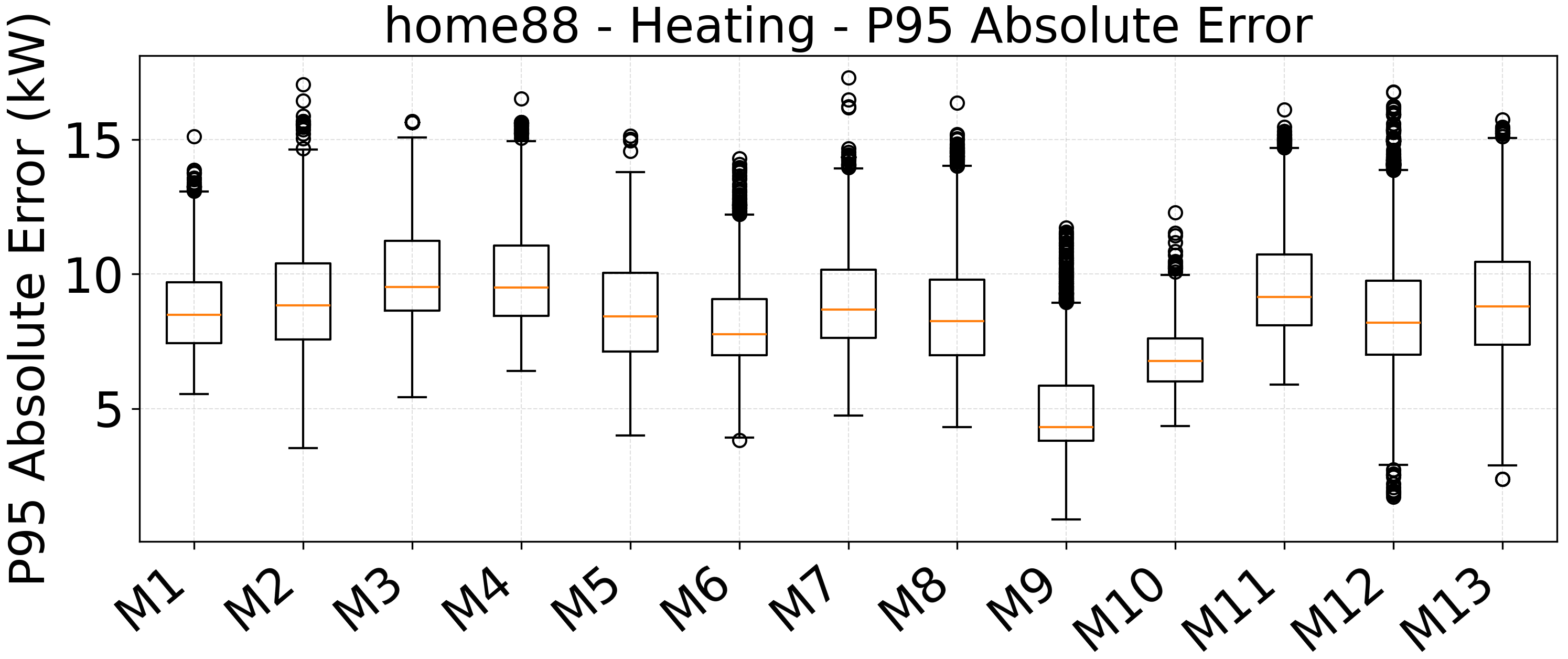}
        \label{fig:home88_P95_absolute_error_Heating}
    }\\
    \subfloat[]{
        \includegraphics[width=0.47\textwidth]{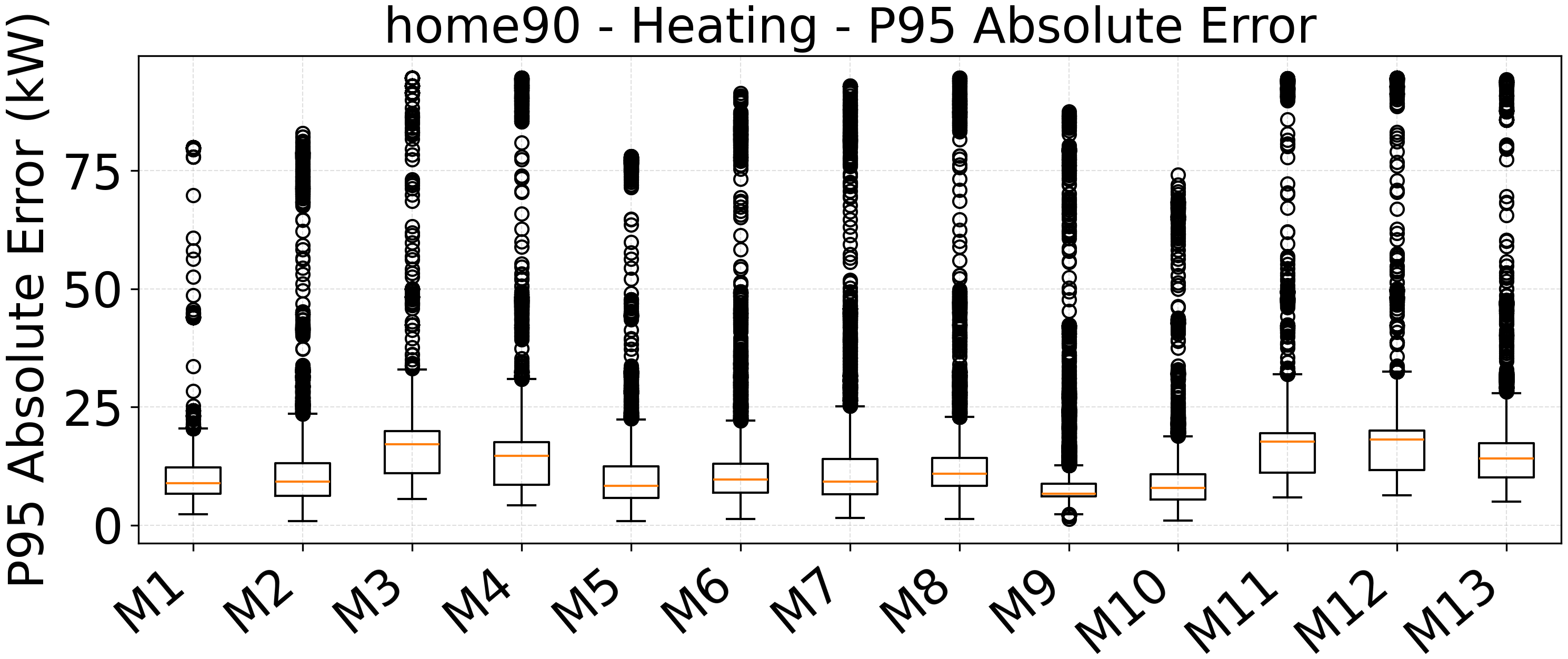}
        \label{fig:home90_P95_absolute_error_Heating}
    }
    \subfloat[]{
        \includegraphics[width=0.47\textwidth]{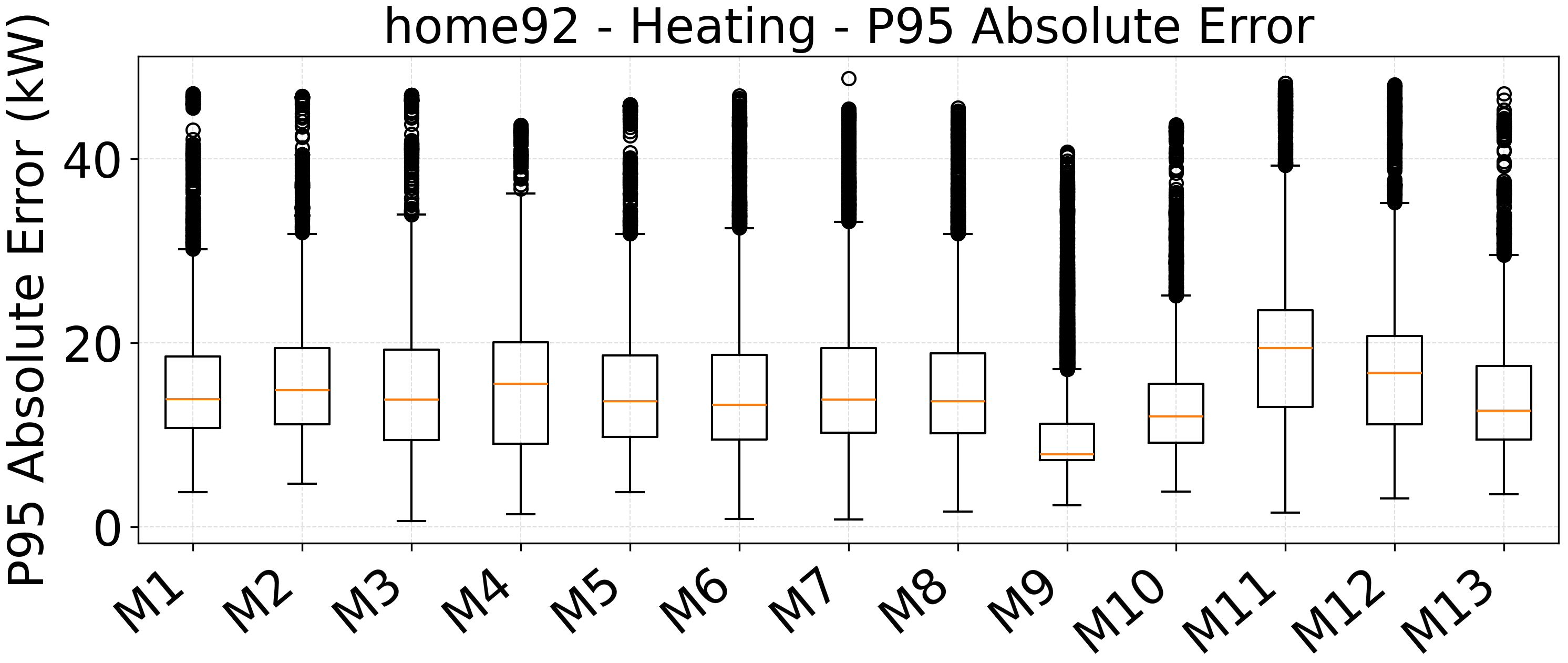}
        \label{fig:home92_P95_absolute_error_Heating}
    }\\
    \subfloat[]{
        \includegraphics[width=0.47\textwidth]{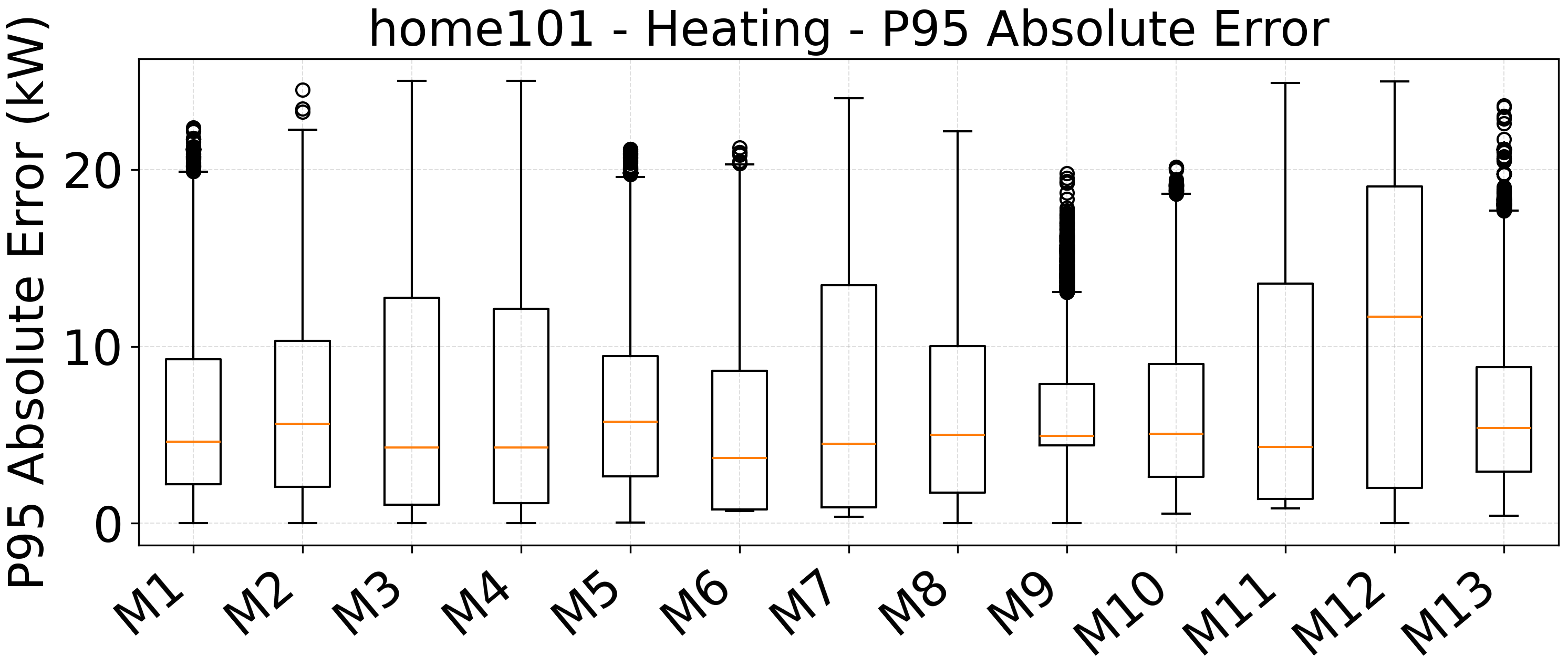}
        \label{fig:home101_ P95_absolute_error_Heating}
    }
    \subfloat[]{
        \includegraphics[width=0.47\textwidth]{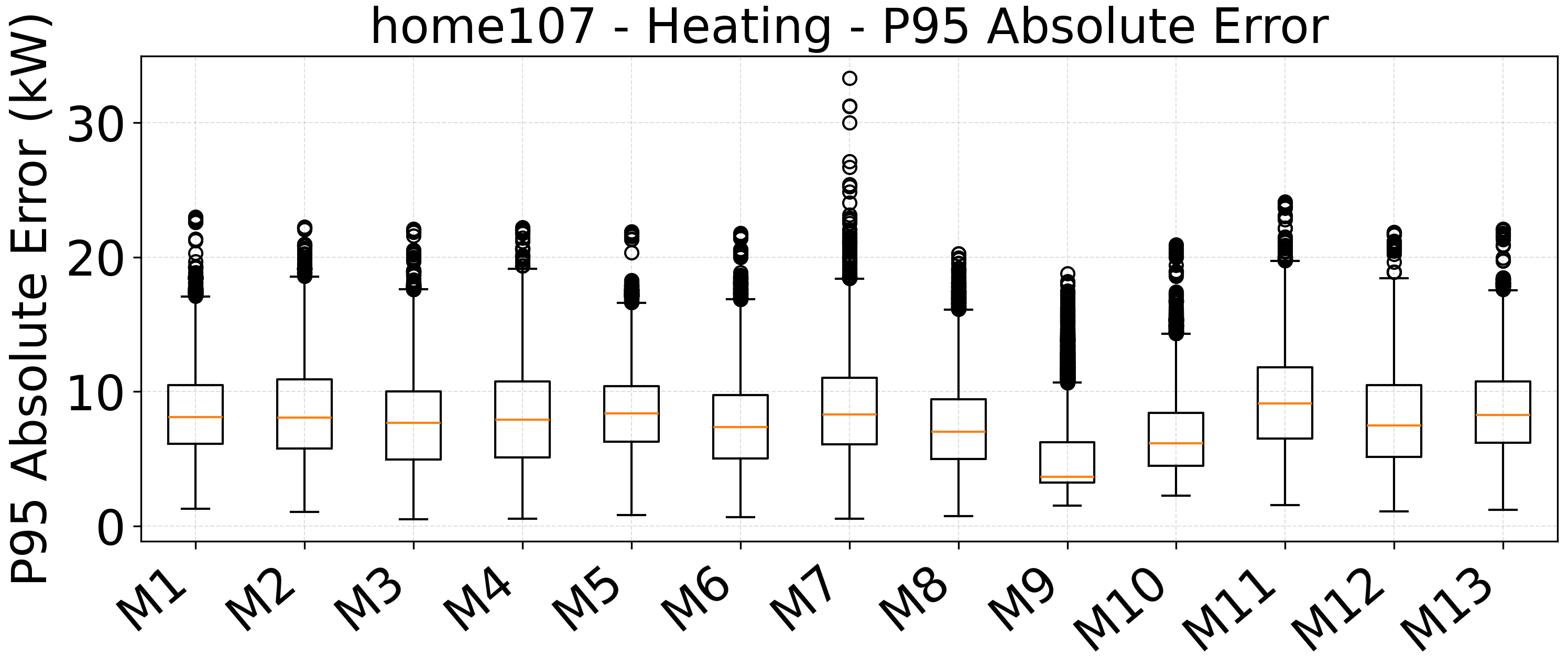}
        \label{fig:home107_P95_absolute_error_Heating}
    }\\
    \subfloat[]{
        \includegraphics[width=0.47\textwidth]{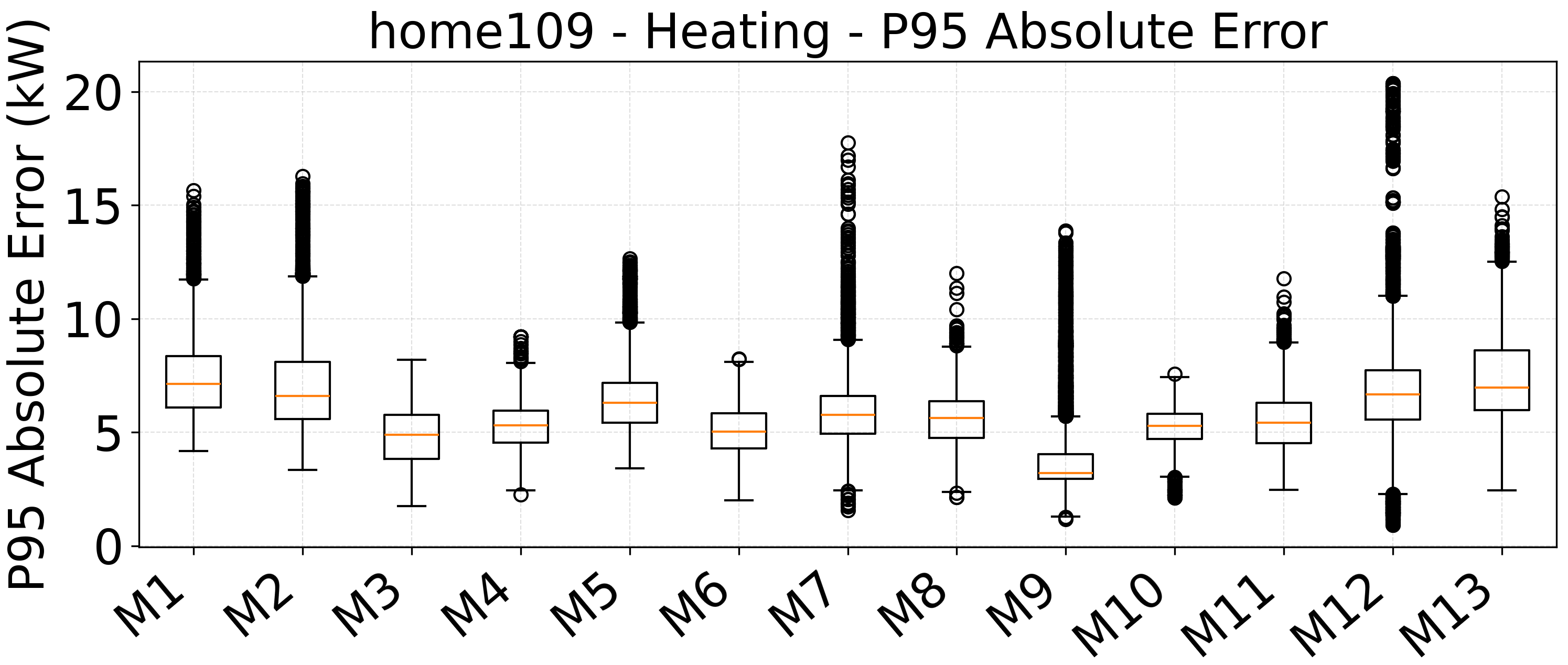}
        \label{fig:home109_P95_absolute_error_Heating}
    }
    \caption{\textrm{P95AE boxplots from the application of all heat forecasting models at home: (a) 68, (b) 81, (c) 82, (d) 88, (e) 90, (f) 92, (g) 101, (h) 107, and (i) 109.}}
    \label{fig:P95AE_heating_comparison}
\end{figure*}

\begin{figure*}[t]
    \centering
    \subfloat[]{
        \includegraphics[width=0.5\textwidth]{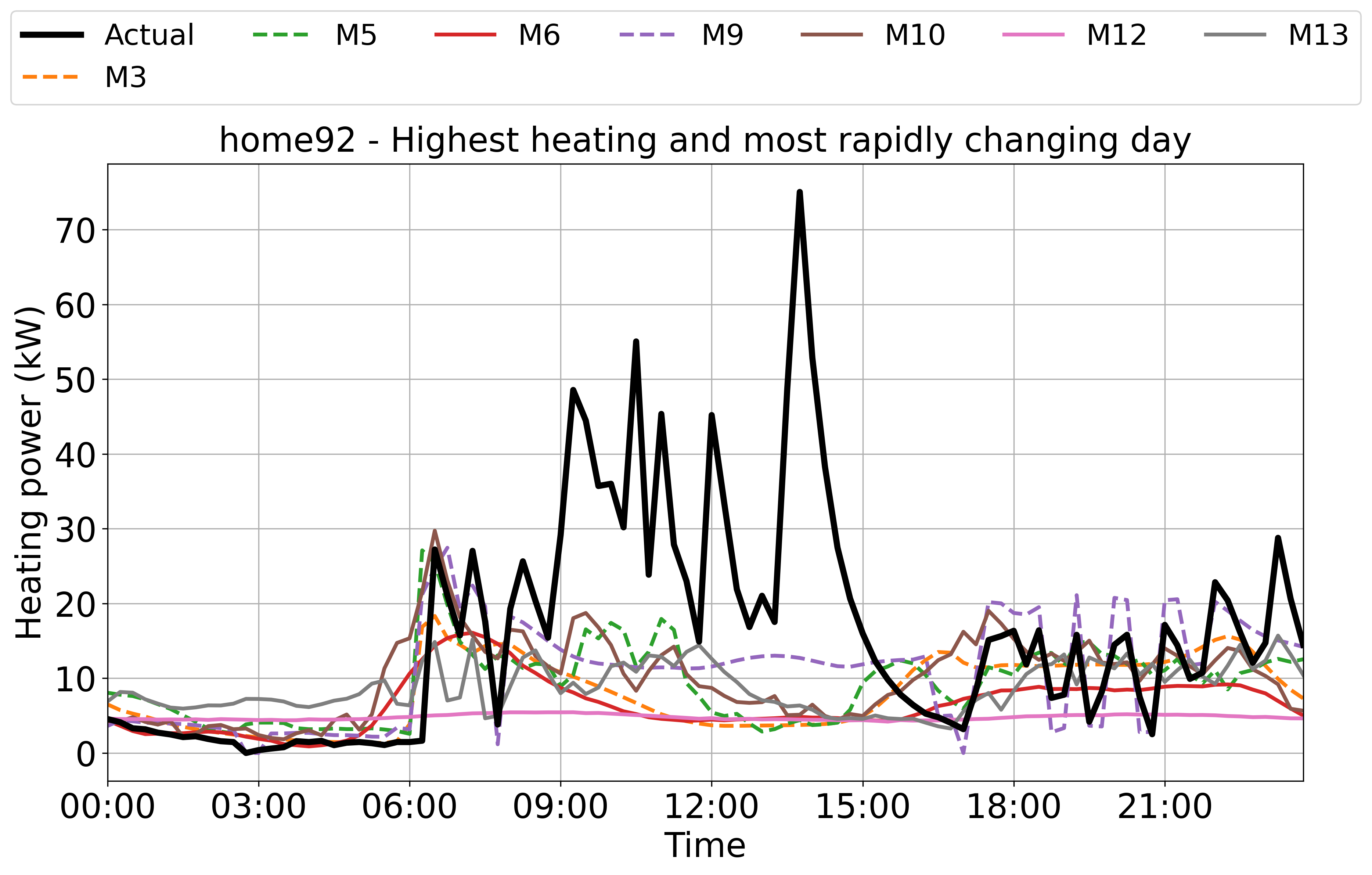}
        \label{fig:home92_Highest_Heating_Day}
    }
    \subfloat[]{
        \includegraphics[width=0.5\textwidth]{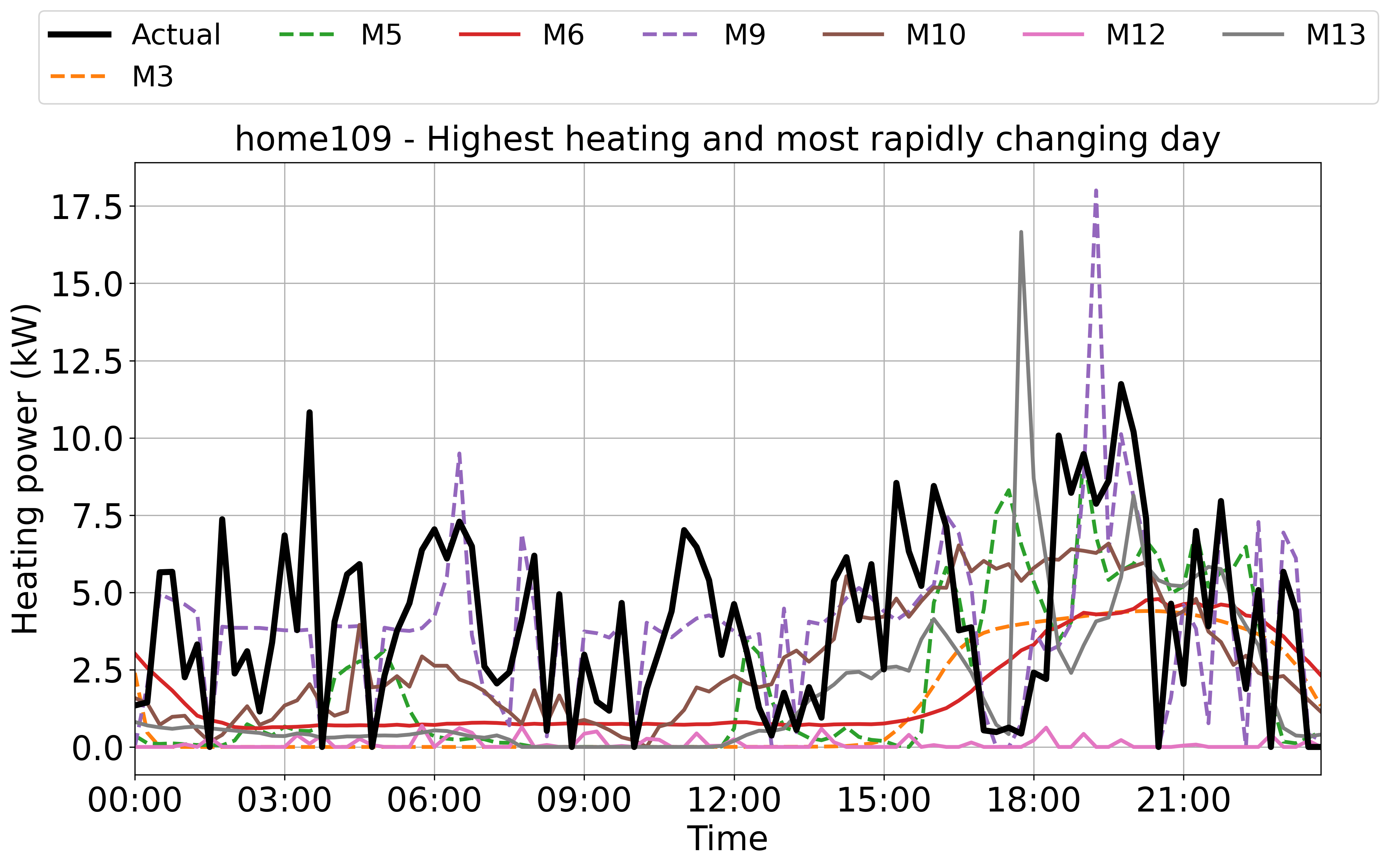}
        \label{fig:home109_Highest_Heating_Day}
    }
    \caption{\textrm{Highest heating demand day with most rapid changes for: (a) home 92 - 22/03/2018, and (b) home 109 - 10/05/2018.}}
    \label{fig:Examples_Heating}
\end{figure*}

The M12-model selects the top 6 input features based on the Pearson correlation coefficient. From the analysis, the historical loads h-1, h-7, w-1, w-7 and the weather parameter td are selected at all households. Moreover, the tt, dhi and hod show comparable correlation, as they are used at 8 buildings. The third group with lower influence on thermal loads includes rh and ghm, as they are selected at 5 households. The remaining parameters ff, gh, dni and dow present low correlation, since they are considered at 1-3 buildings. 

By using a RF feature importance ranking, the top 6 input features are selected by the M13-model. From the evaluation, the weather parameters ff, tt, td, rh and the historical load w-1 are selected at all households. High correlation with the thermal loads is observed for the parameters ghm and dow, which are used at 7 and 8 buildings, respectively. Besides them, the historical loads h-1, w-7 and the weather parameter dni present lower impact on the thermal loads with selection at 5-6 households. Finally, gh is considered at a building, while dhi and hod are not used at any household. 

In summary, the historical loads h-1, h-7, w-1 and w-7 present the highest frequency of selection by the forecasting models applied for the examined households. From the weather parameters, tt and td present clearly dominant influence on the thermal loads. Concerning the calendar parameters, hod shows high frequency of selection for most forecasting models. Furthermore, dow presents low correlation with the thermal loads. Moreover, the feature selection method can lead to contradictory results. Particularly, some features, e.g., ff, dow and dni, are rarely selected by most models, while the RF-based feature importance ranking of M13-model leads to frequent selection. Finally, among the solar parameters gh, dhi and ghm, there is not a clear trend about the most influential factor, since their selection varies on the model.

\subsection{Heating demand forecasts}
\label{heat_dem_for}

At this stage of evaluation, we calculate the metrics MAE, RMSE and P95AE for the day-ahead forecasts calculated at each timestep of the validation period. Next, for each home, we generate boxplots that show the distribution of the metrics for each method. 

In Fig.~\ref{fig:MAE_heating_comparison}, the MAE boxplots are illustrated, when applying the examined hybrid forecasting models for each home. First of all, the maximum MAE varies among the examined methods for each home considering the outliers of all boxplots. Nevertheless, it can be observed that the method M9-EMD-LSTM-Markov presents the lowest MAE values among almost all houses apart from the household 90. The model M9 also shows the lowest 50\% percentiles at most buildings apart from the houses 81 and 101. The next model, which presents most frequently the lowest MAE-percentiles is the M3-CLSTM-AR-MFFCLA for the buildings 92, 101, 107 and 109. Comparable performance is also noticed for the households 81 and 90 with the models M6-TE-BiSRU-MTL and M5-FTTrans-E-BL, respectively. The model M10-SVM-SA presents low MAE-values for the households 81, 82 and 88. Besides them, the M13-RF-ISSA-LSTM and M2-DwdAdam-ILSTM shows the second lowest MAE-percentiles for the buildings 68 and 88, correspondingly. Moreover, the hybrid method M12-SSA-CNN-SVM indicates high performance for the households 82, 88 and 107.

At the next step, the same process is repeated for the metric RMSE, as displayed in Fig.~\ref{fig:RMSE_heating_comparison}. From the maximum values of all methods including the outliers, M9 surpasses the other methods at six households, while M10 presents the highest performance at the remaining buildings. When comparing the 50\% RMSE-percentiles, M9 shows the lowest values at most households apart from 81 and 101, where M10 and M11 surpass the other methods. Moreover, M10 shows the second lowest 50\% RMSE-percentiles at most buildings taking into account a comparable performance for M6 at buildings 101 and 109. Additional models that achieve slightly higher 50\% RMSE-percentiles are M13, M3, and M5. 

\begin{figure*}[t]
    \centering
    \subfloat[]{
        \includegraphics[width=0.47\textwidth]{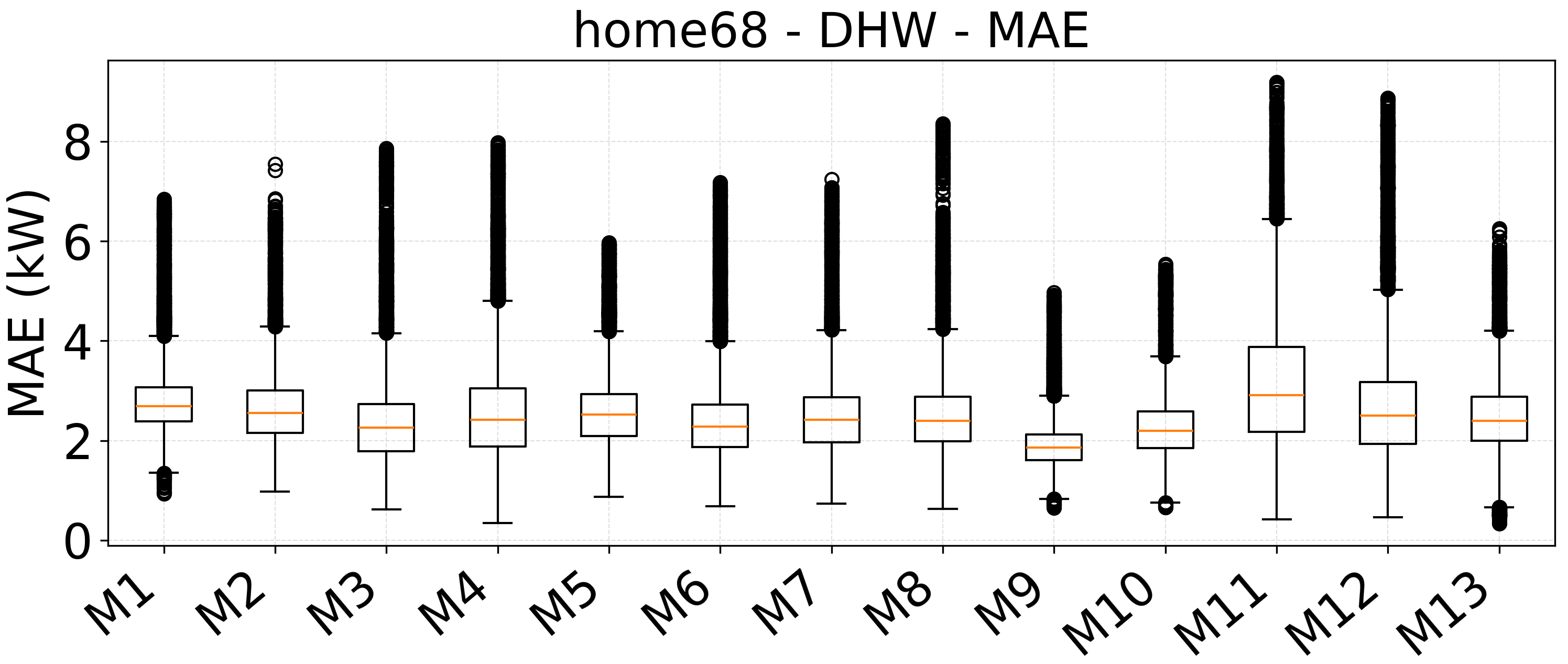}
        \label{fig:home68_MAE_DHW}
    }
    \subfloat[]{
        \includegraphics[width=0.47\textwidth]{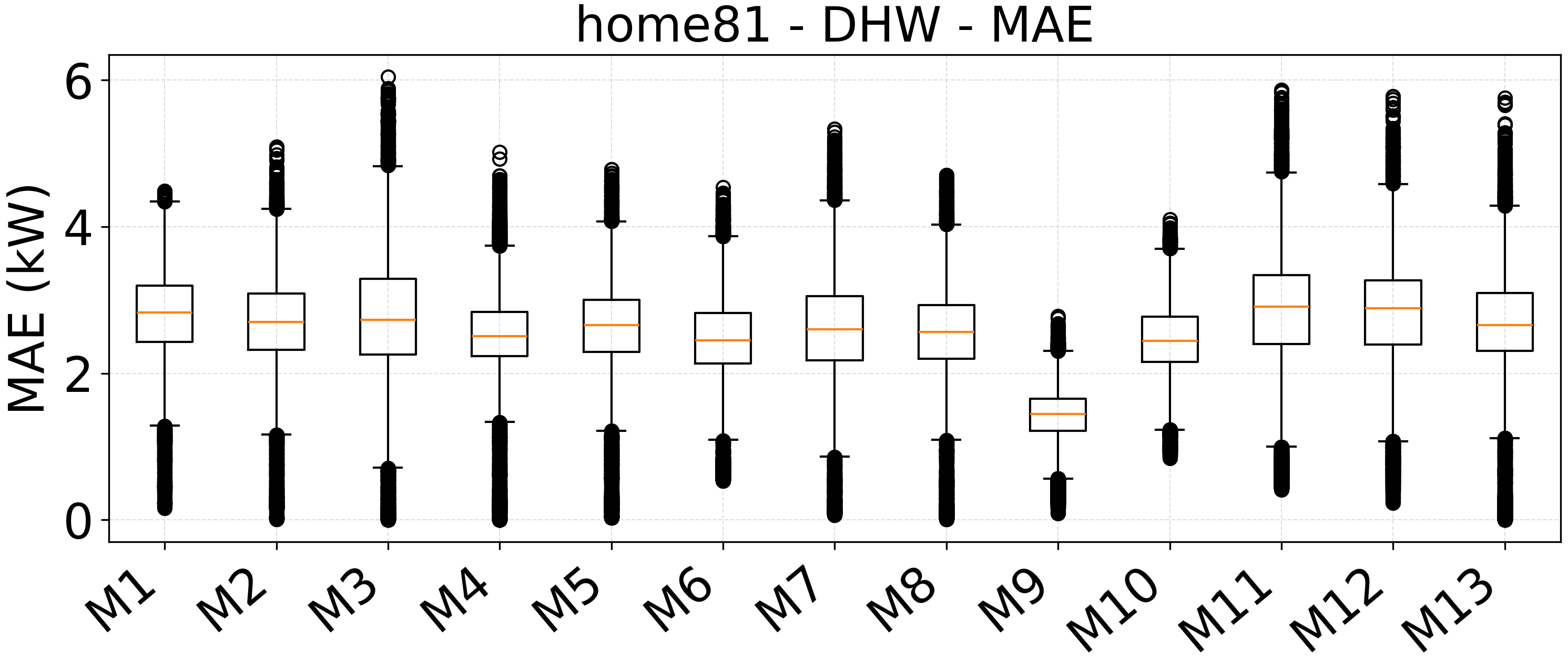}
        \label{fig:home81_MAE_DHW}
    }\\
    \subfloat[]{
        \includegraphics[width=0.47\textwidth]{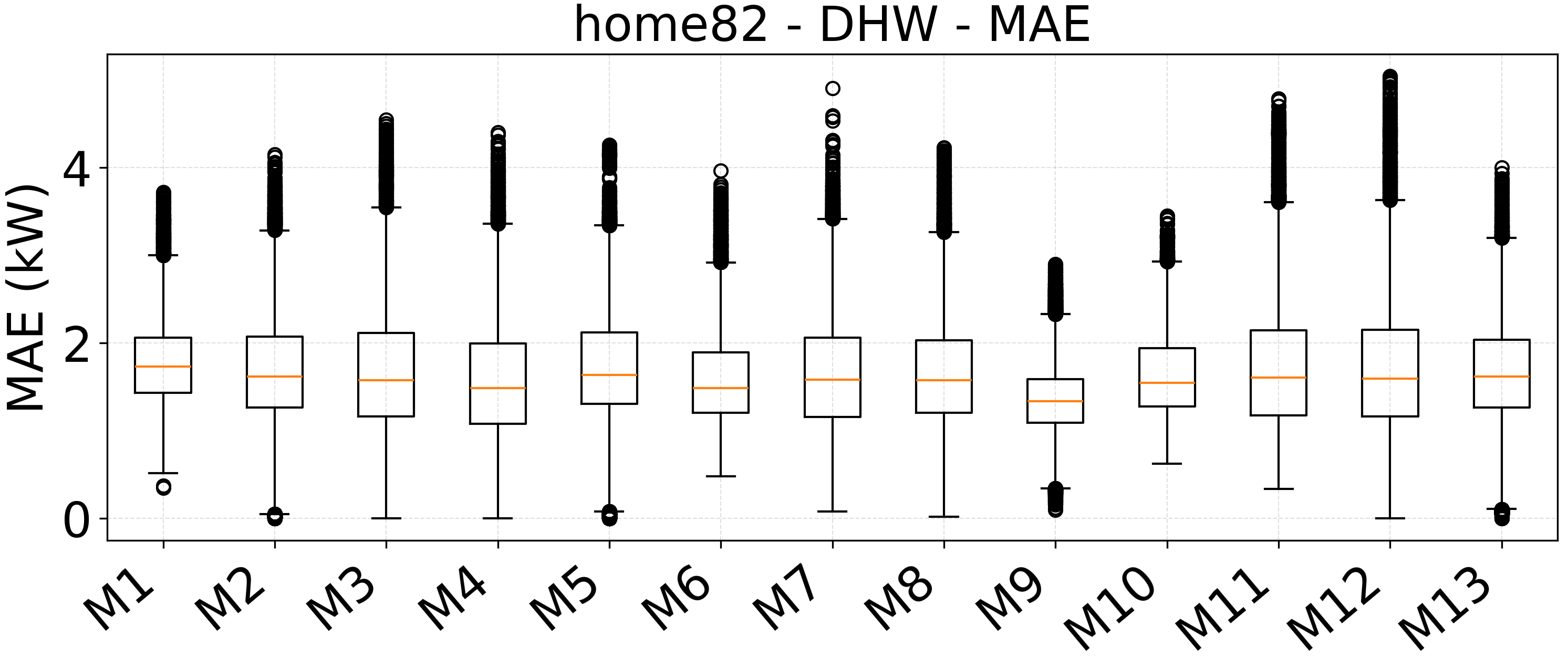}
        \label{fig:home82_MAE_DHW}
    }
    \subfloat[]{
        \includegraphics[width=0.47\textwidth]{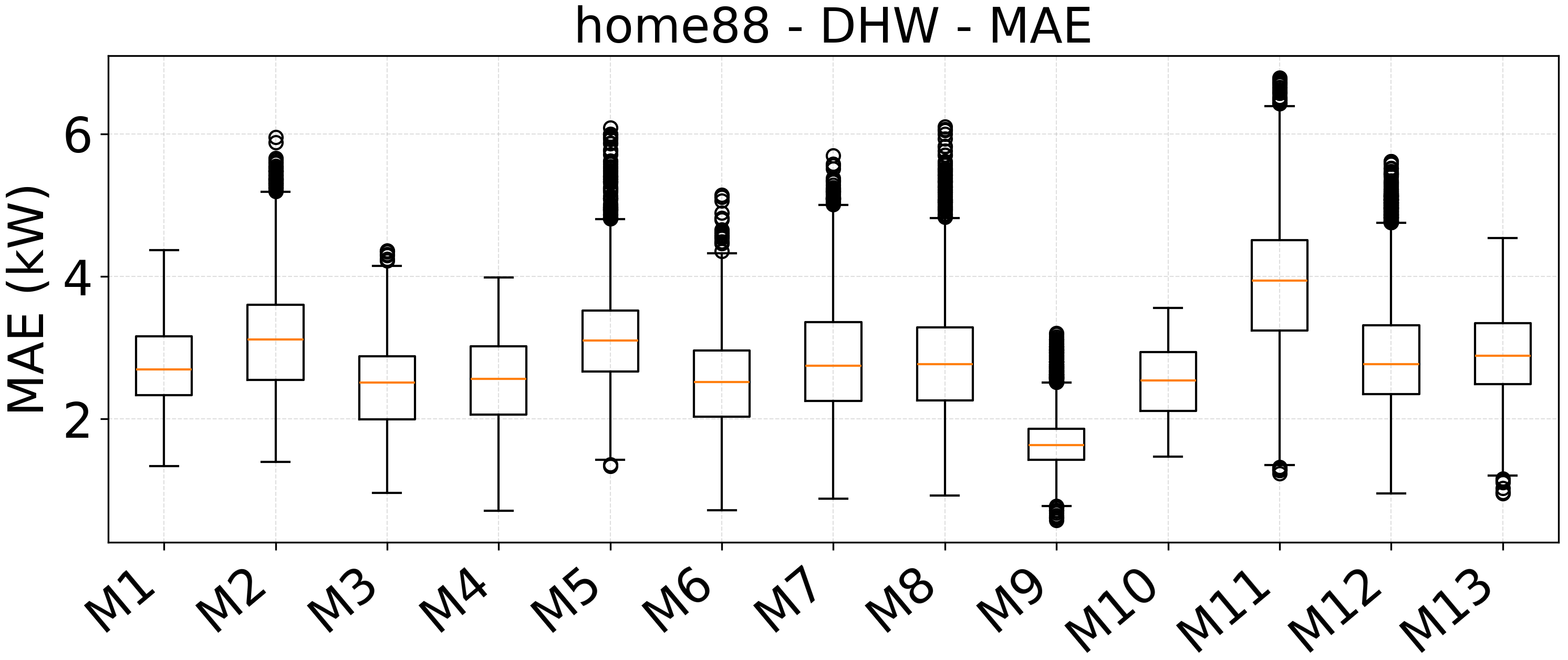}
        \label{fig:home88_MAE_DHW}
    }\\
    \subfloat[]{
        \includegraphics[width=0.47\textwidth]{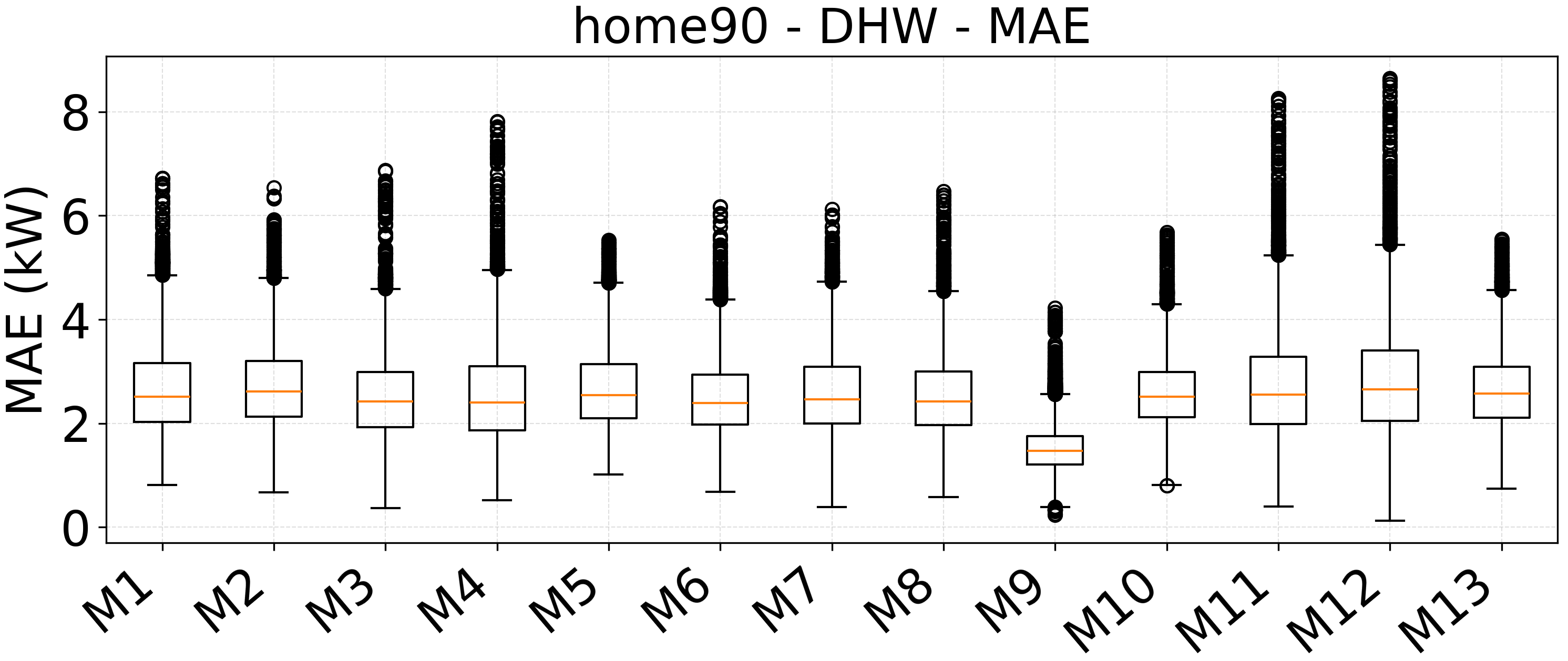}
        \label{fig:home90_MAE_DHW}
    }
    \subfloat[]{
        \includegraphics[width=0.47\textwidth]{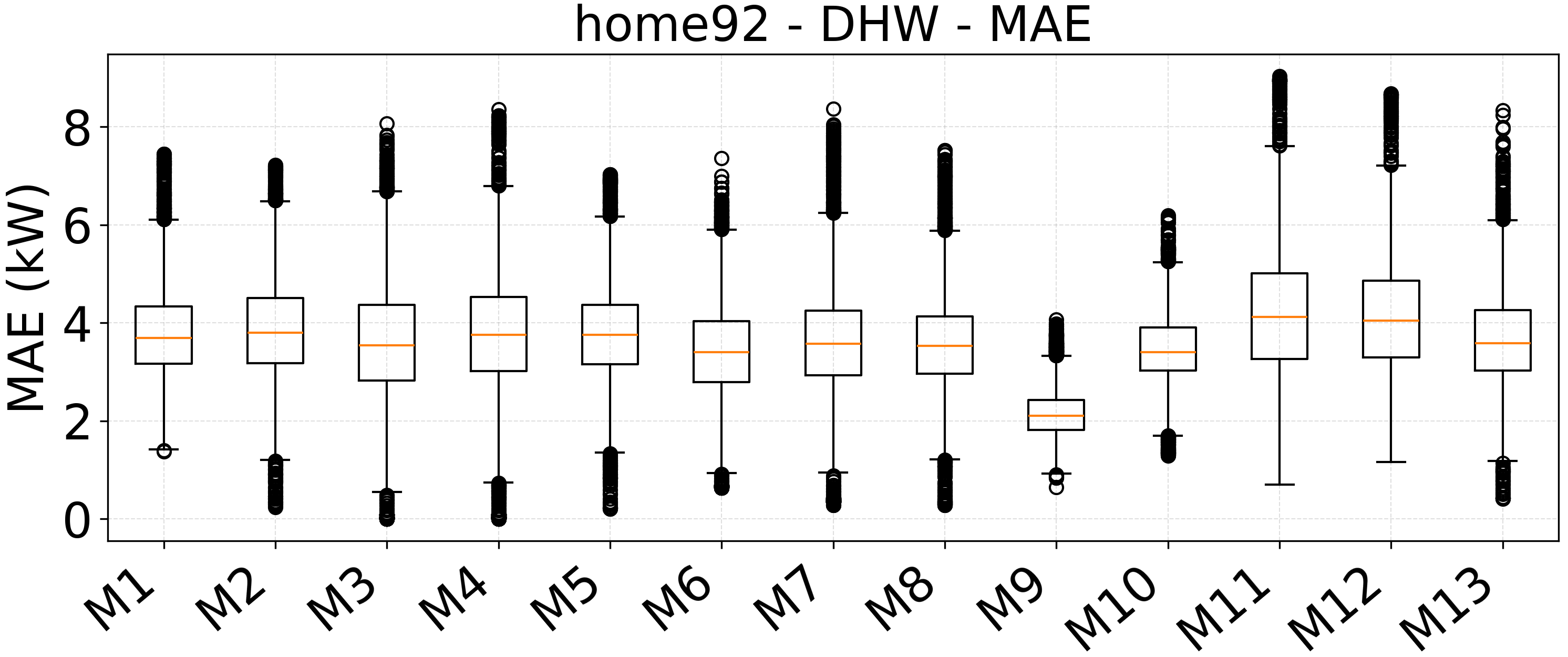}
        \label{fig:home92_MAE_DHW}
    }\\
    \subfloat[]{
        \includegraphics[width=0.47\textwidth]{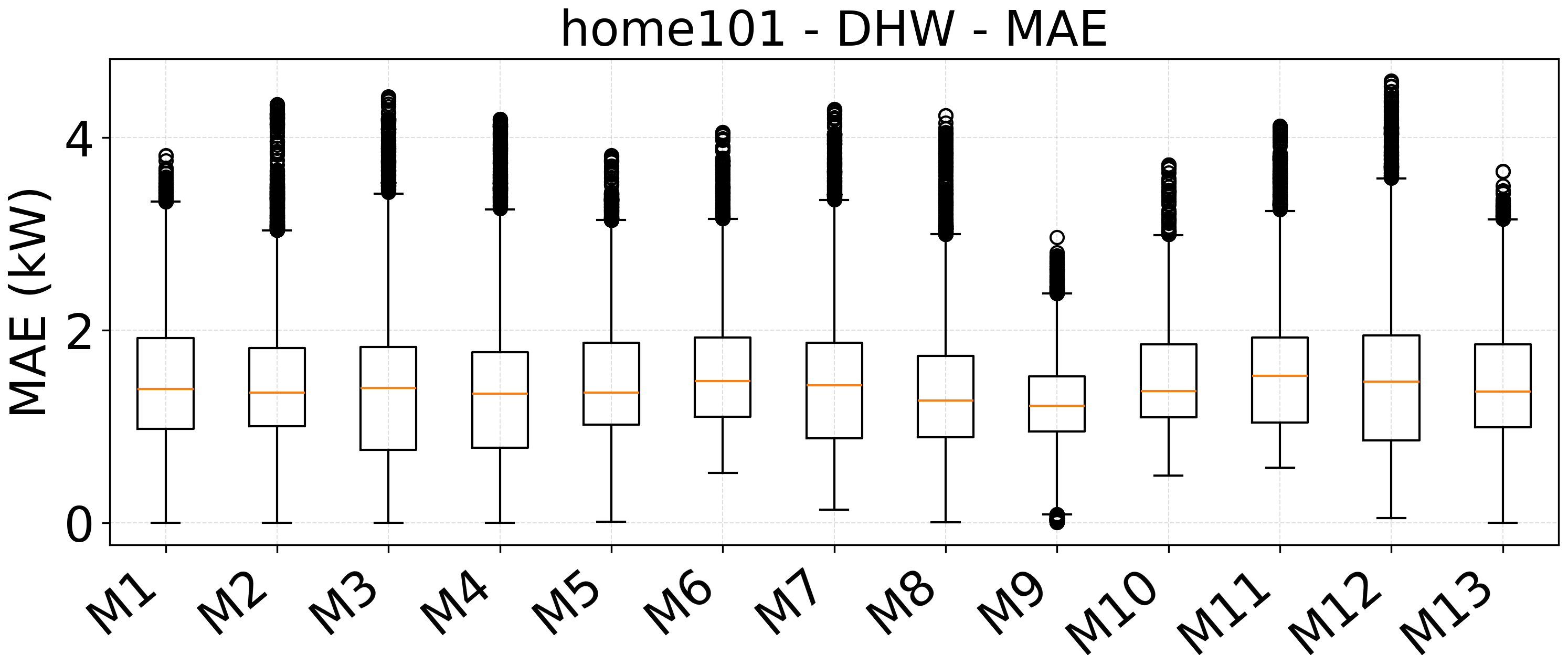}
        \label{fig:home101_MAE_DHW}
    }
    \subfloat[]{
        \includegraphics[width=0.47\textwidth]{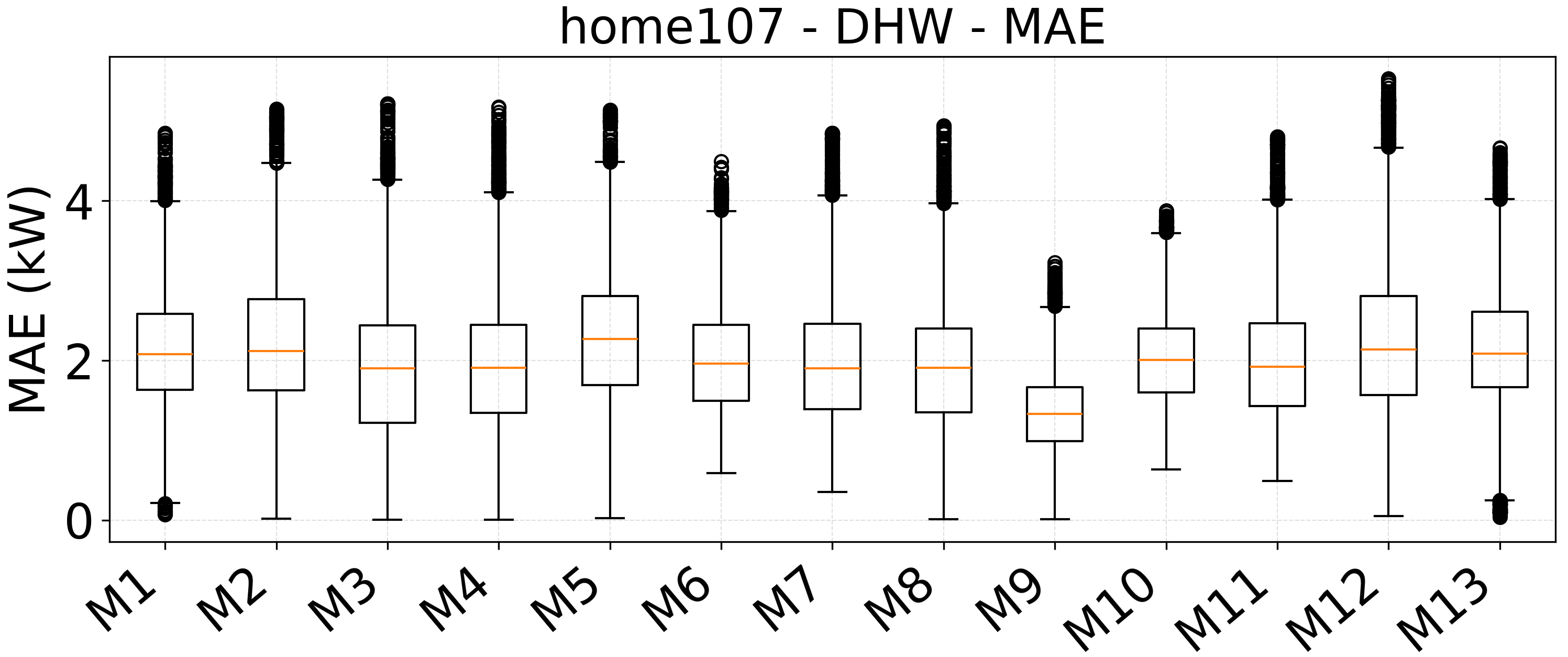}
        \label{fig:home107_MAE_DHW}
    }\\
    \subfloat[]{
        \includegraphics[width=0.47\textwidth]{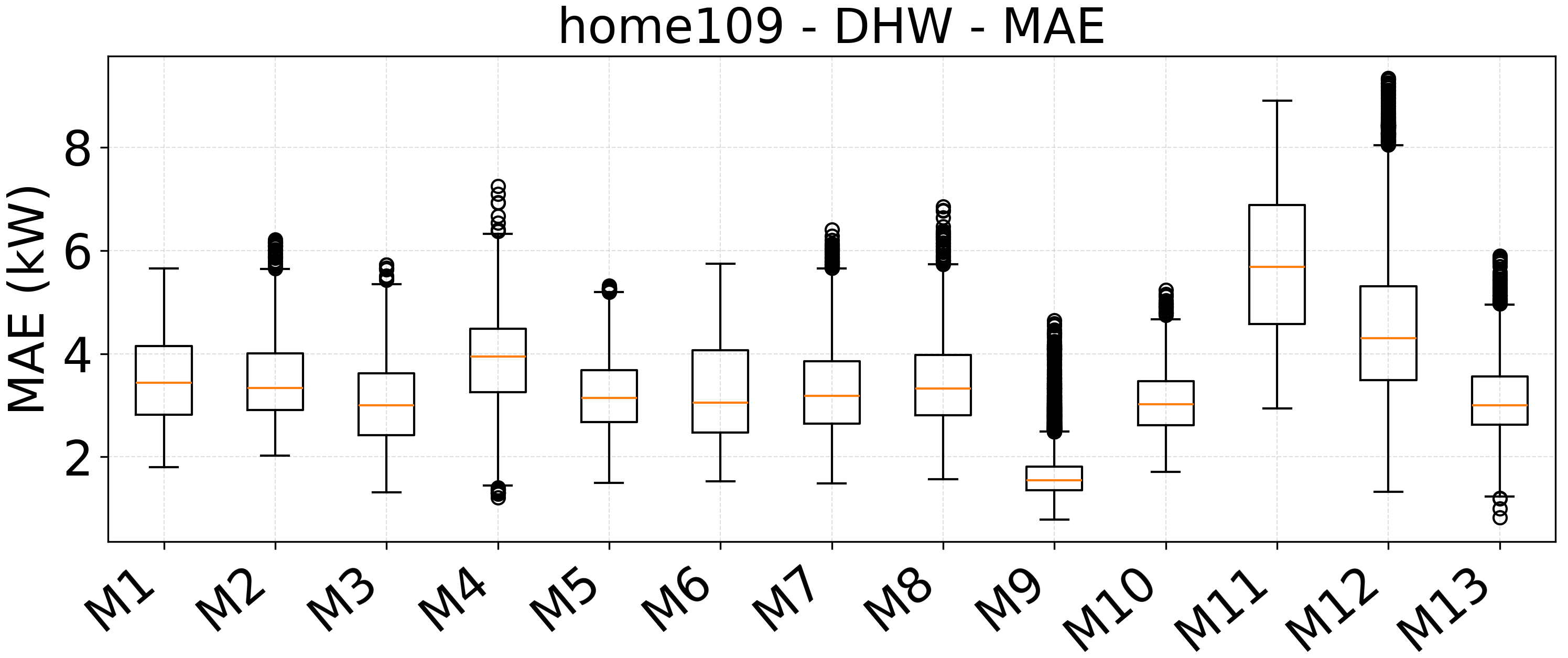}
        \label{fig:home109_MAE_DHW}
    }
    \caption{\textrm{MAE boxplots from the application of all DHW forecasting models at home: (a) 68, (b) 81, (c) 82, (d) 88, (e) 90, (f) 92, (g) 101, (h) 107, and (i) 109.}}
    \label{fig:MAE_DHW_comparison}
\end{figure*}

\begin{figure*}[t]
    \centering
    \subfloat[]{
        \includegraphics[width=0.47\textwidth]{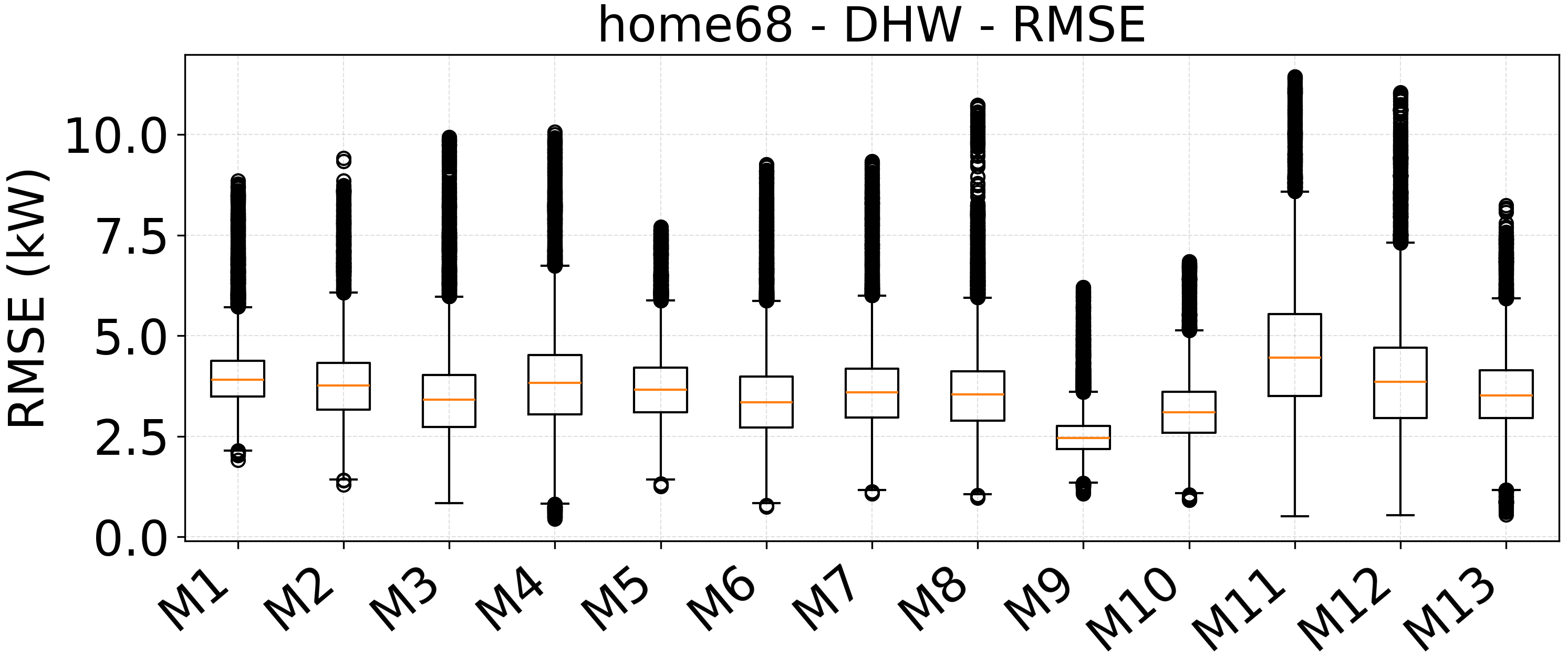}
        \label{fig:home68_RMSE_DHW}
    }
    \subfloat[]{
        \includegraphics[width=0.47\textwidth]{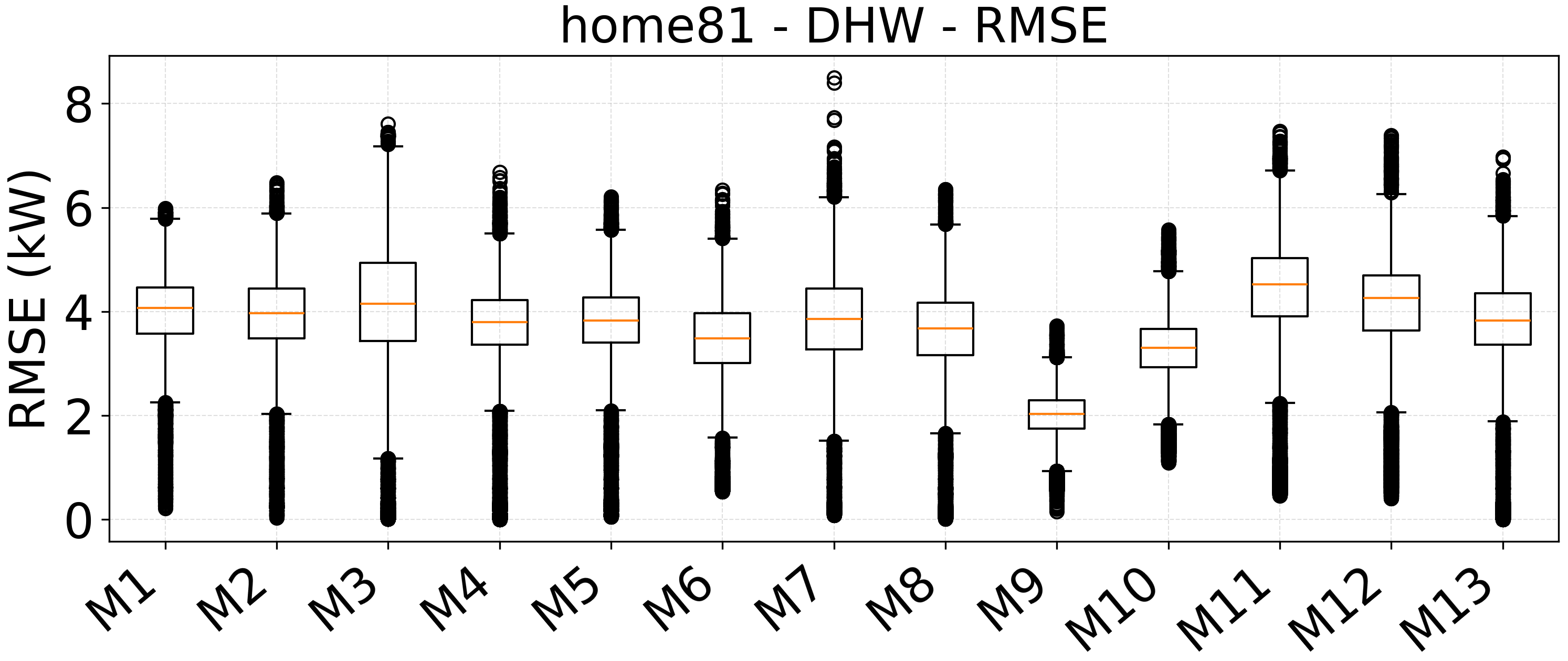}
        \label{fig:home81_RMSE_DHW}
    }\\
    \subfloat[]{
        \includegraphics[width=0.47\textwidth]{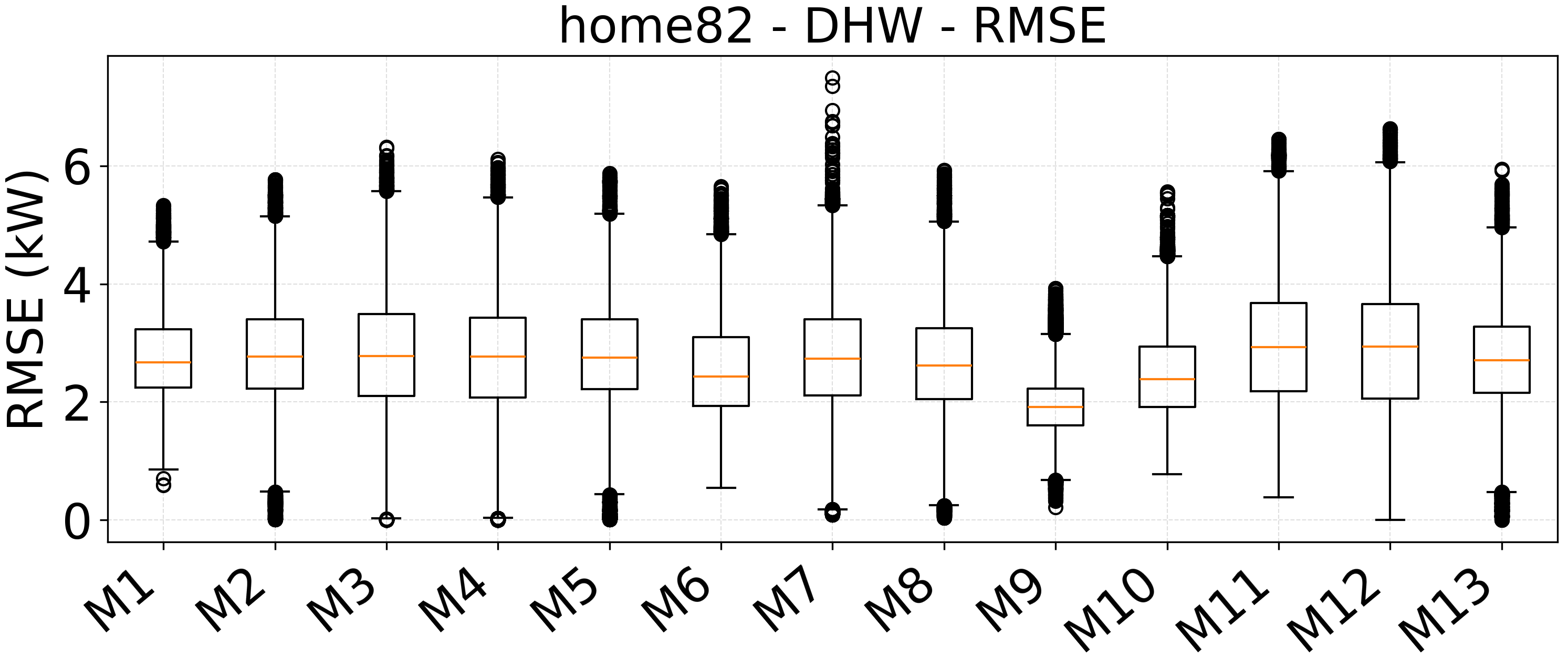}
        \label{fig:home82_RMSE_DHW}
    }
    \subfloat[]{
        \includegraphics[width=0.47\textwidth]{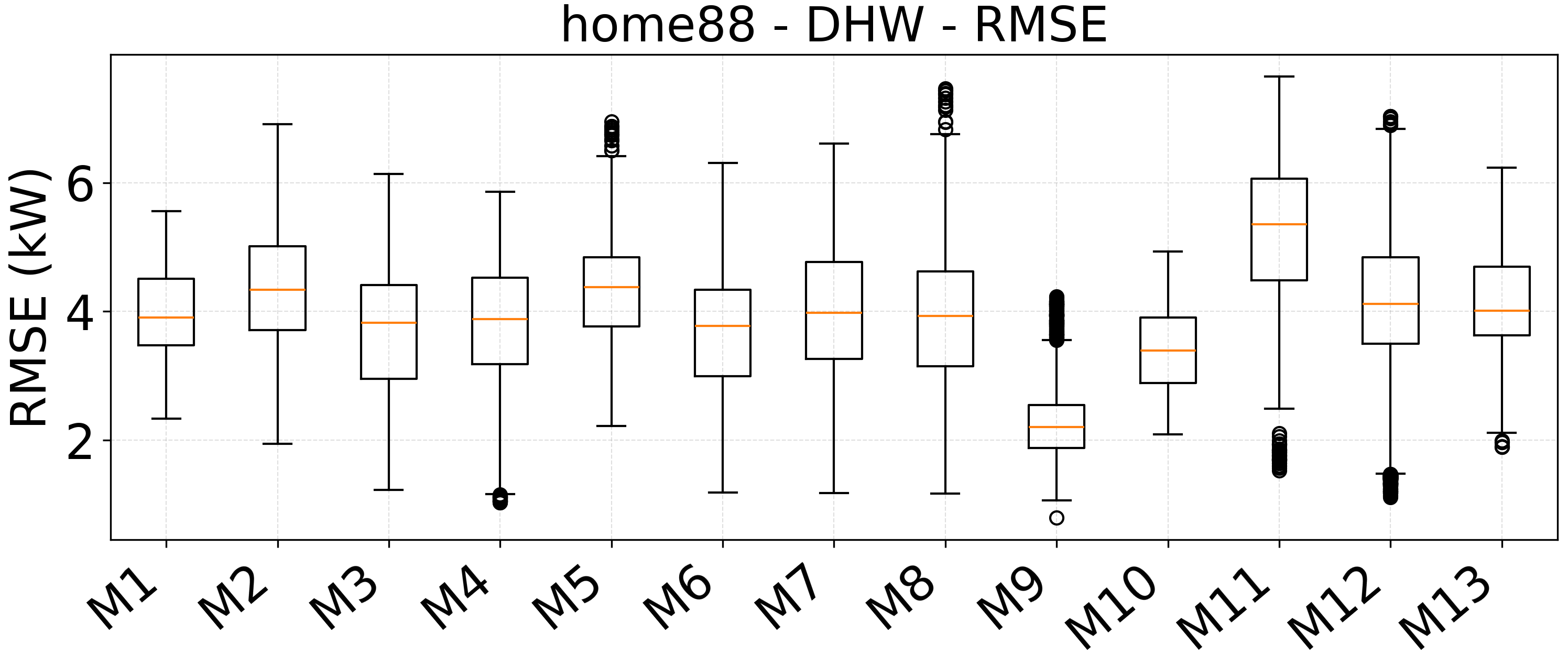}
        \label{fig:home88_RMSE_DHW}
    }\\
    \subfloat[]{
        \includegraphics[width=0.47\textwidth]{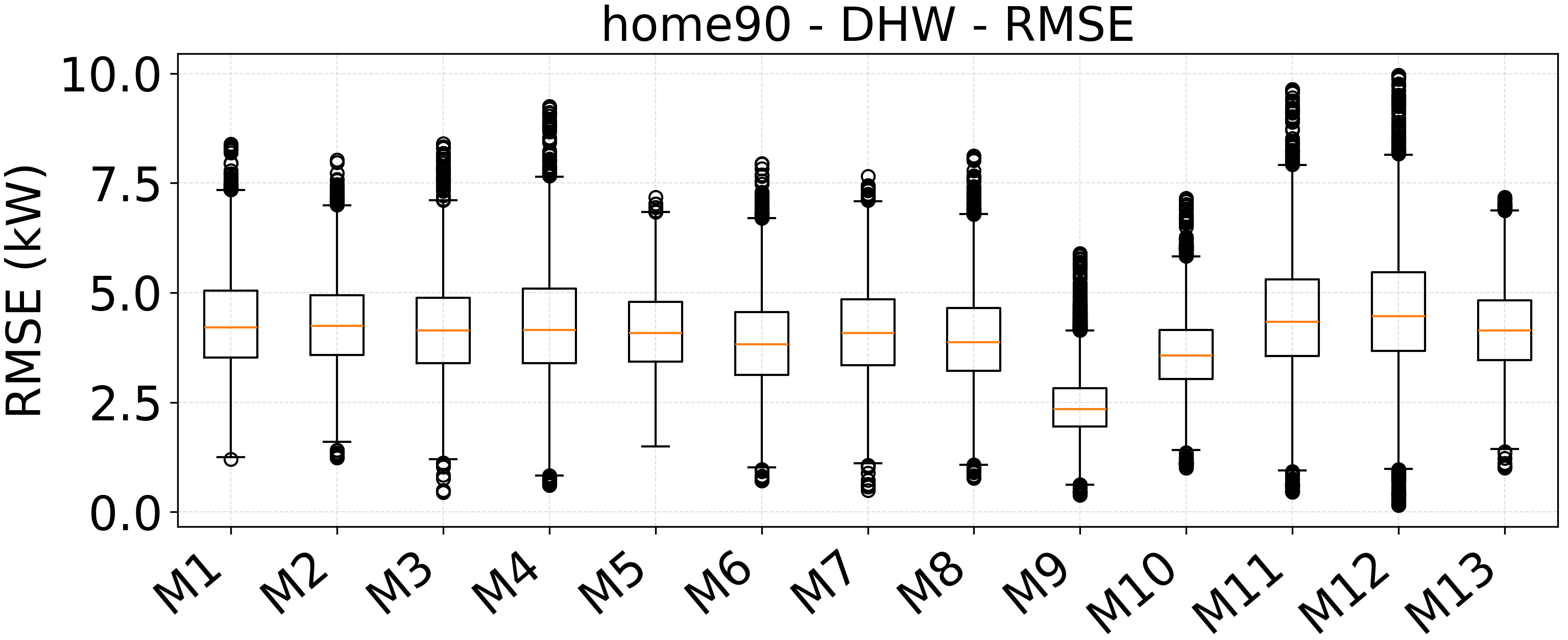}
        \label{fig:home90_RMSE_DHW}
    }
    \subfloat[]{
        \includegraphics[width=0.47\textwidth]{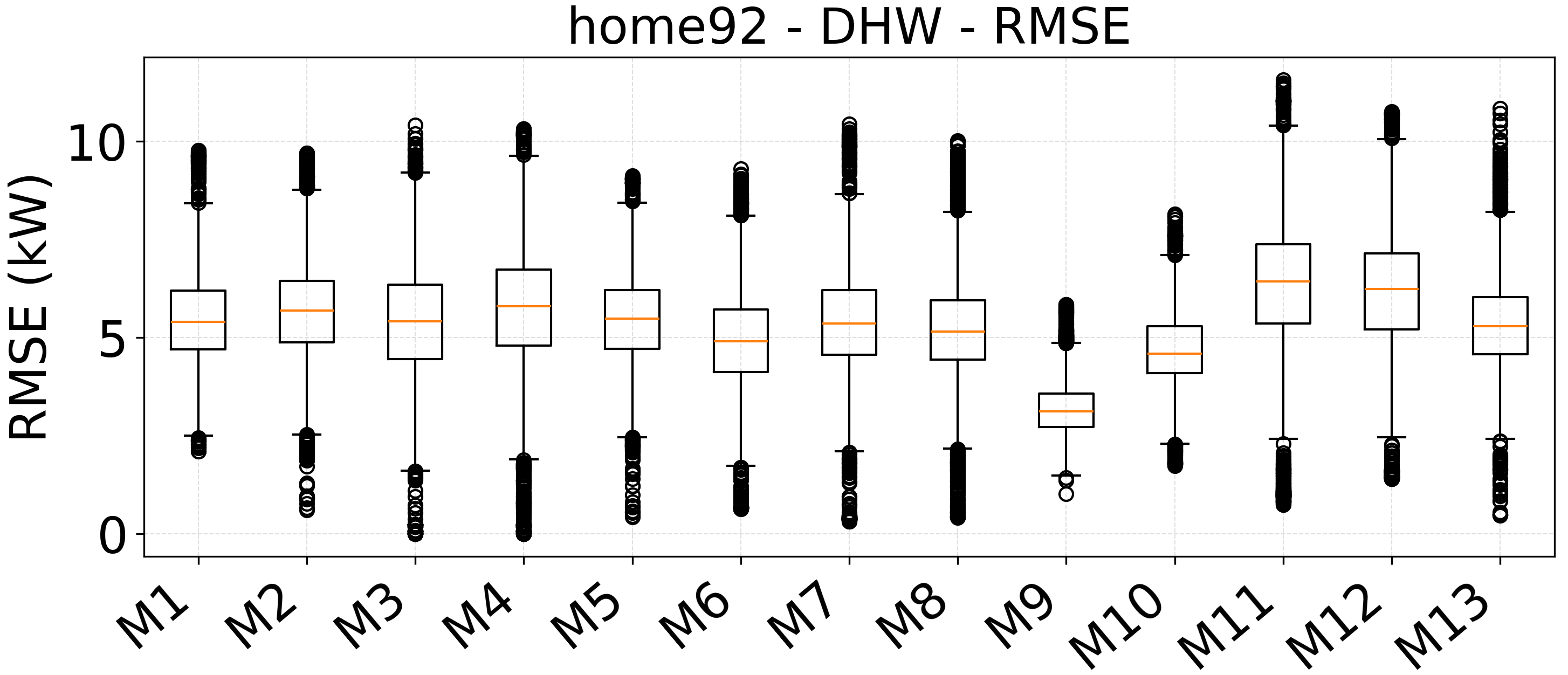}
        \label{fig:home92_RMSE_DHW}
    }\\
    \subfloat[]{
        \includegraphics[width=0.47\textwidth]{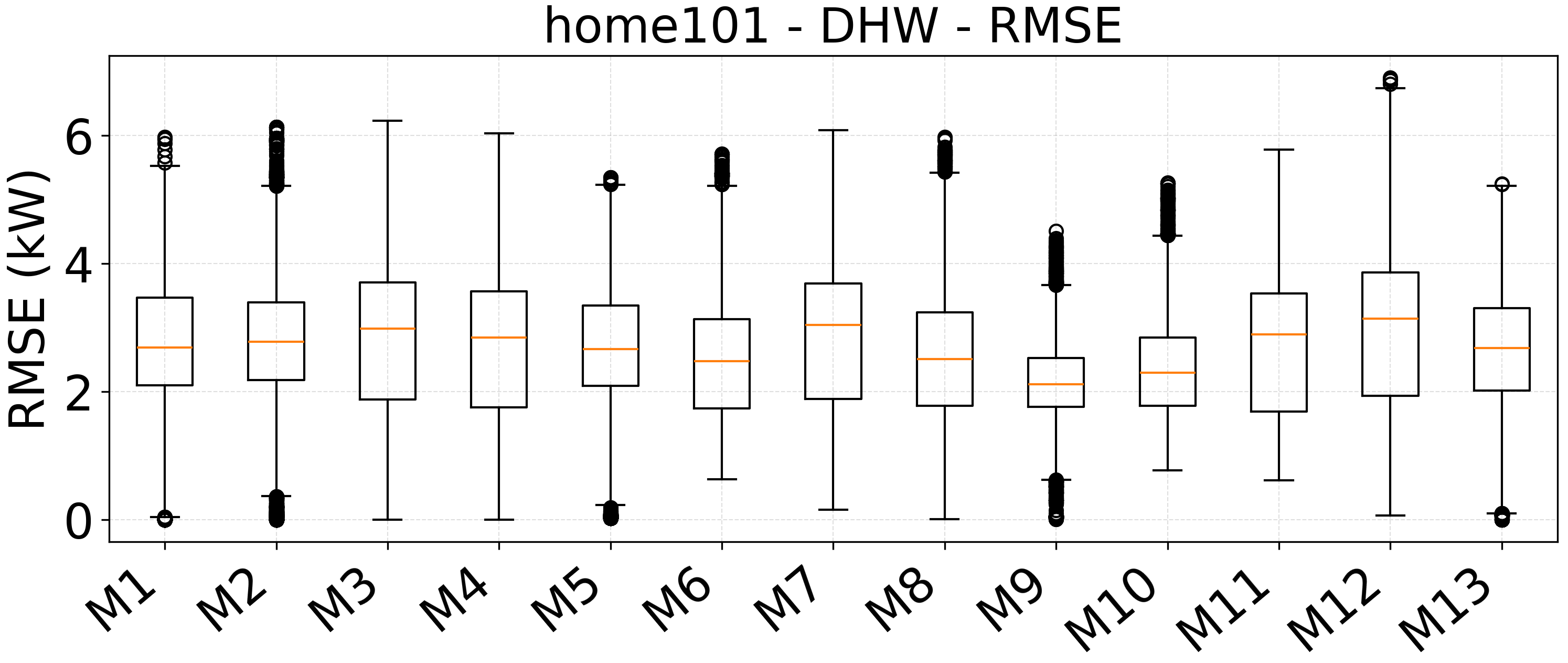}
        \label{fig:home101_RMSE_DHW}
    }
    \subfloat[]{
        \includegraphics[width=0.47\textwidth]{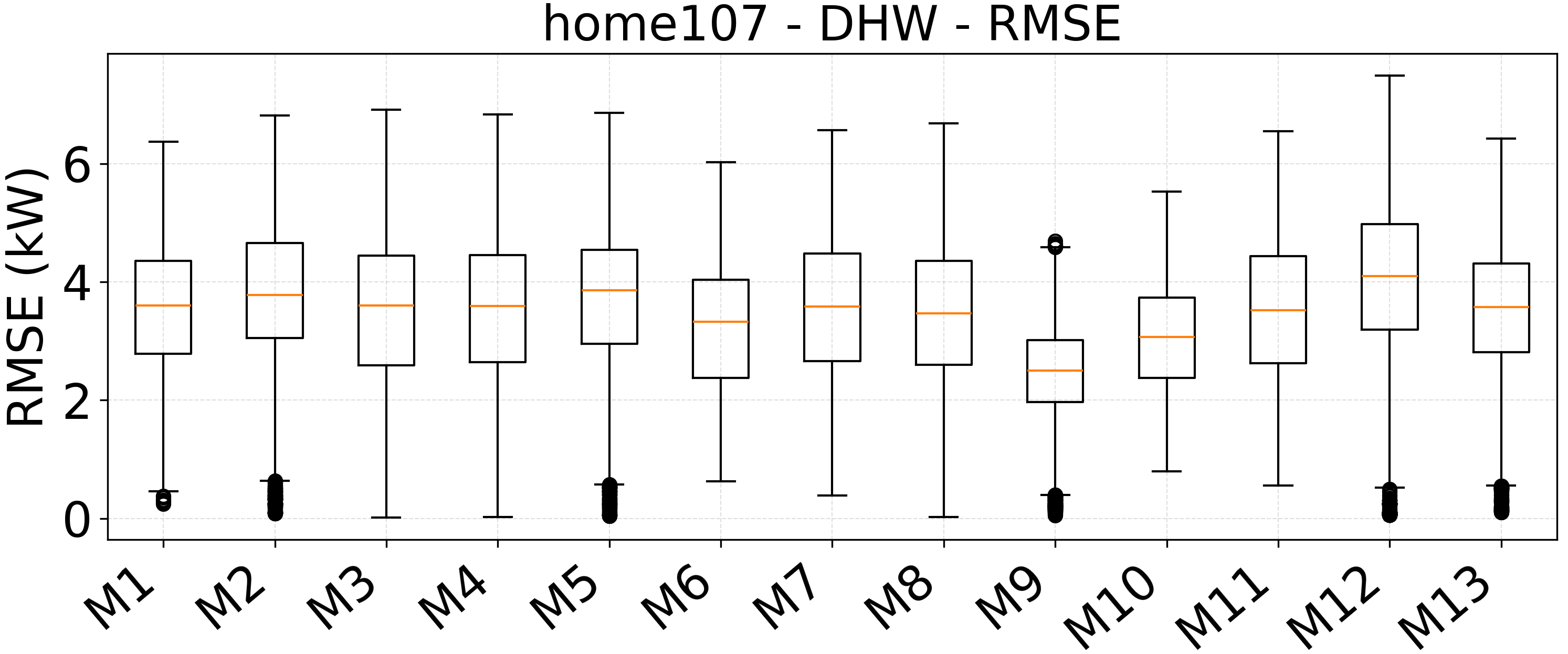}
        \label{fig:home107_RMSE_DHW}
    }\\
    \subfloat[]{
        \includegraphics[width=0.47\textwidth]{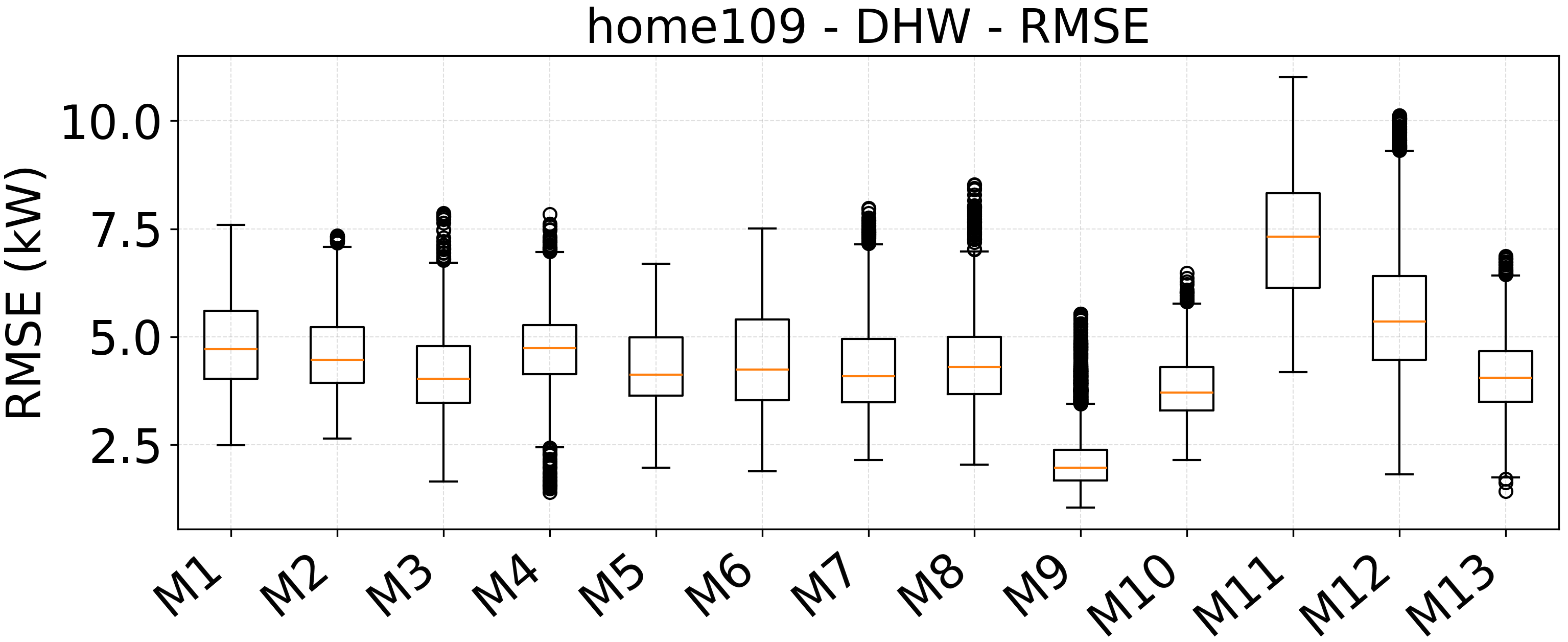}
        \label{fig:home109_RMSE_DHW}
    }
    \caption{\textrm{RMSE boxplots from the application of all DHW forecasting models at home: (a) 68, (b) 81, (c) 82, (d) 88, (e) 90, (f) 92, (g) 101, (h) 107, and (i) 109.}}
    \label{fig:RMSE_DHW_comparison}
\end{figure*}

\begin{figure*}[t]
    \centering
    \subfloat[]{
        \includegraphics[width=0.47\textwidth]{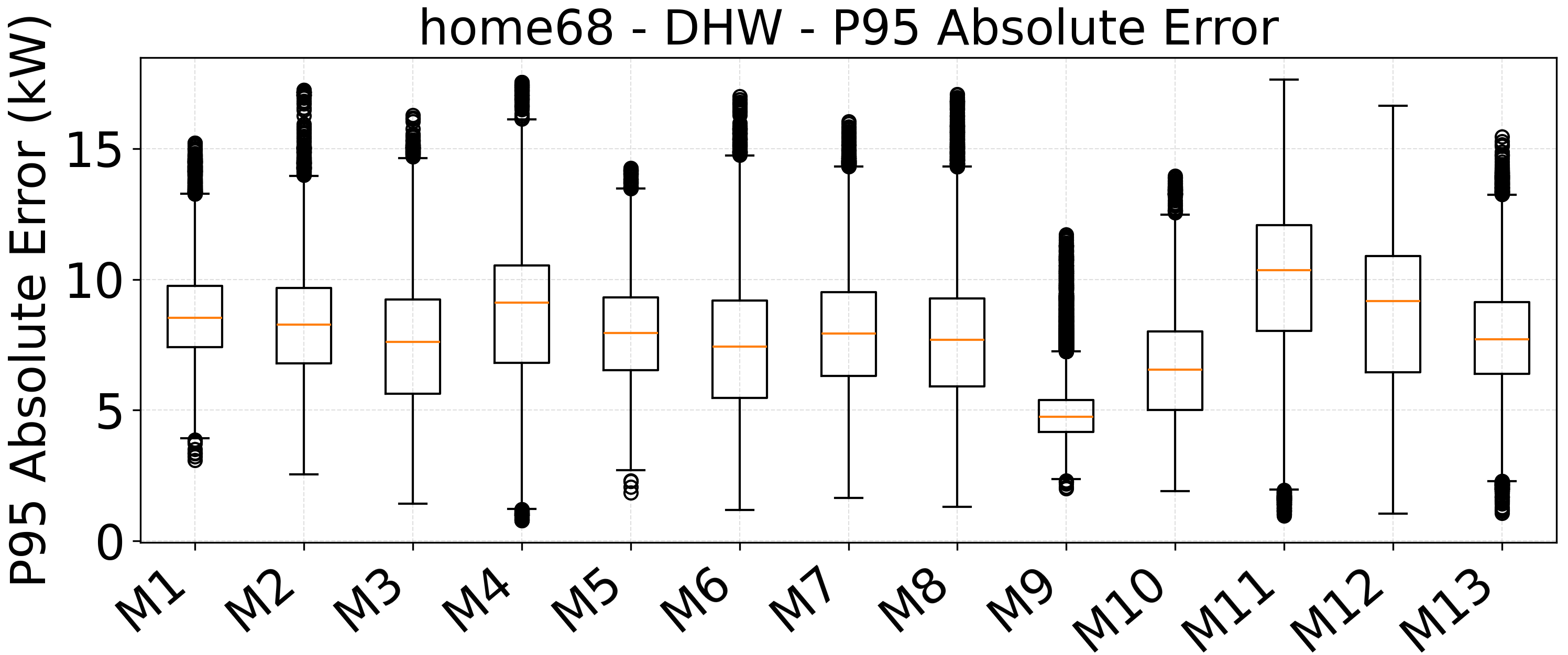}
        \label{fig:home68_P95_absolute_error_DHW}
    }
    \subfloat[]{
        \includegraphics[width=0.47\textwidth]{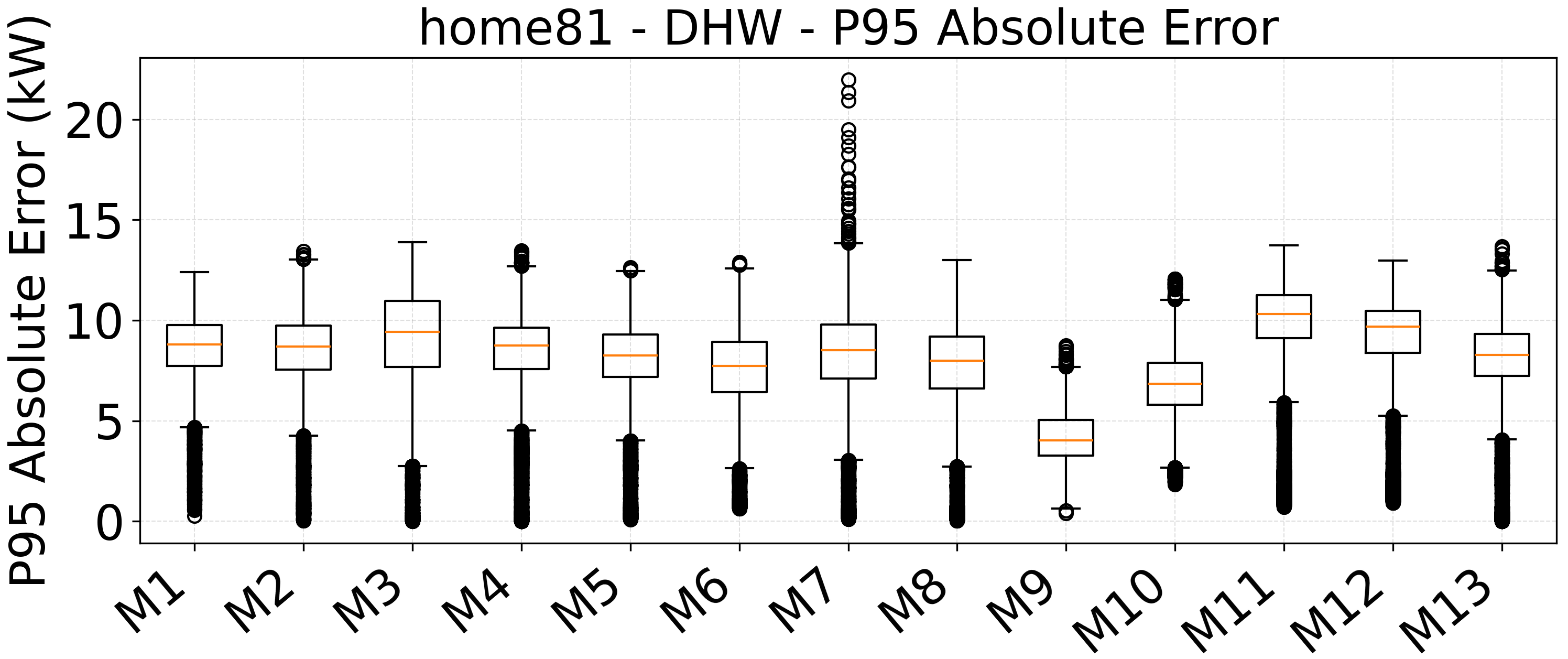}
        \label{fig:home81_P95_absolute_error_DHW}
    }\\
    \subfloat[]{
        \includegraphics[width=0.47\textwidth]{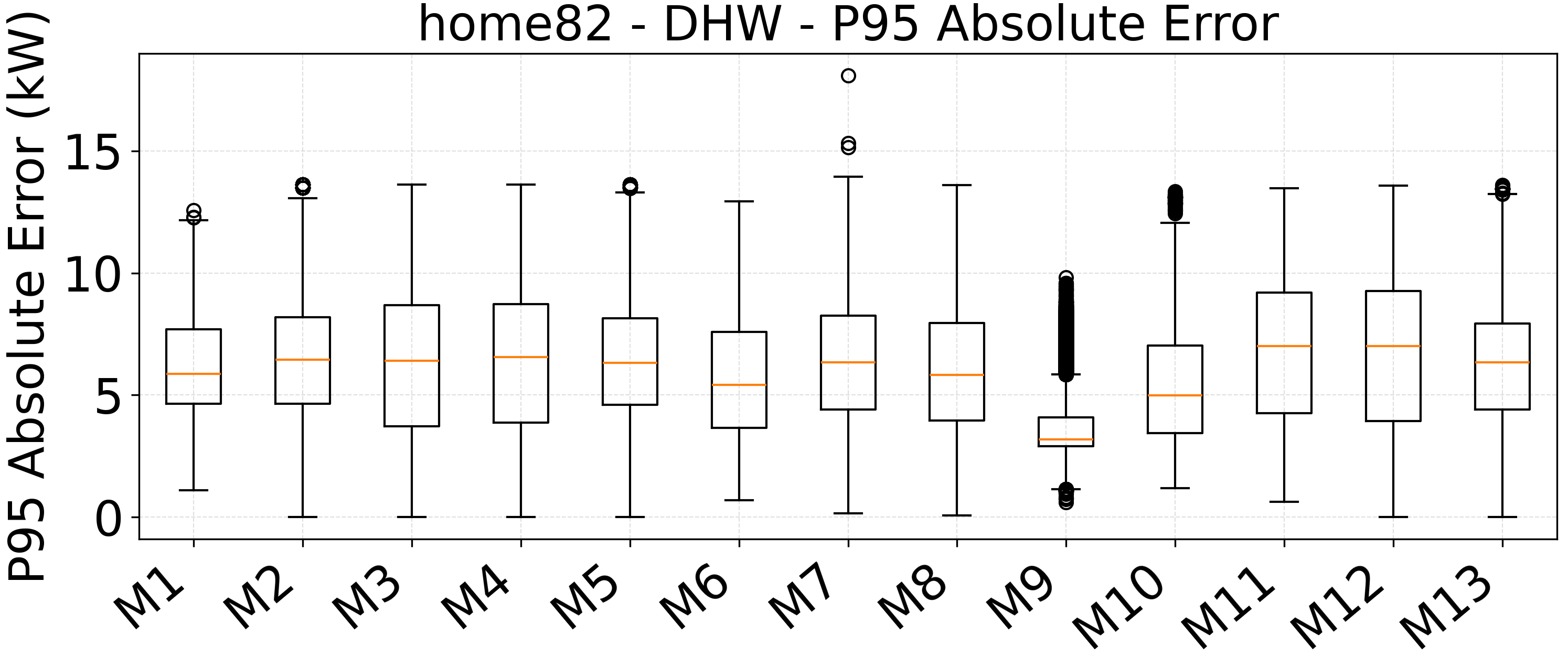}
        \label{fig:home82_P95_absolute_error_DHW}
    }
    \subfloat[]{
        \includegraphics[width=0.47\textwidth]{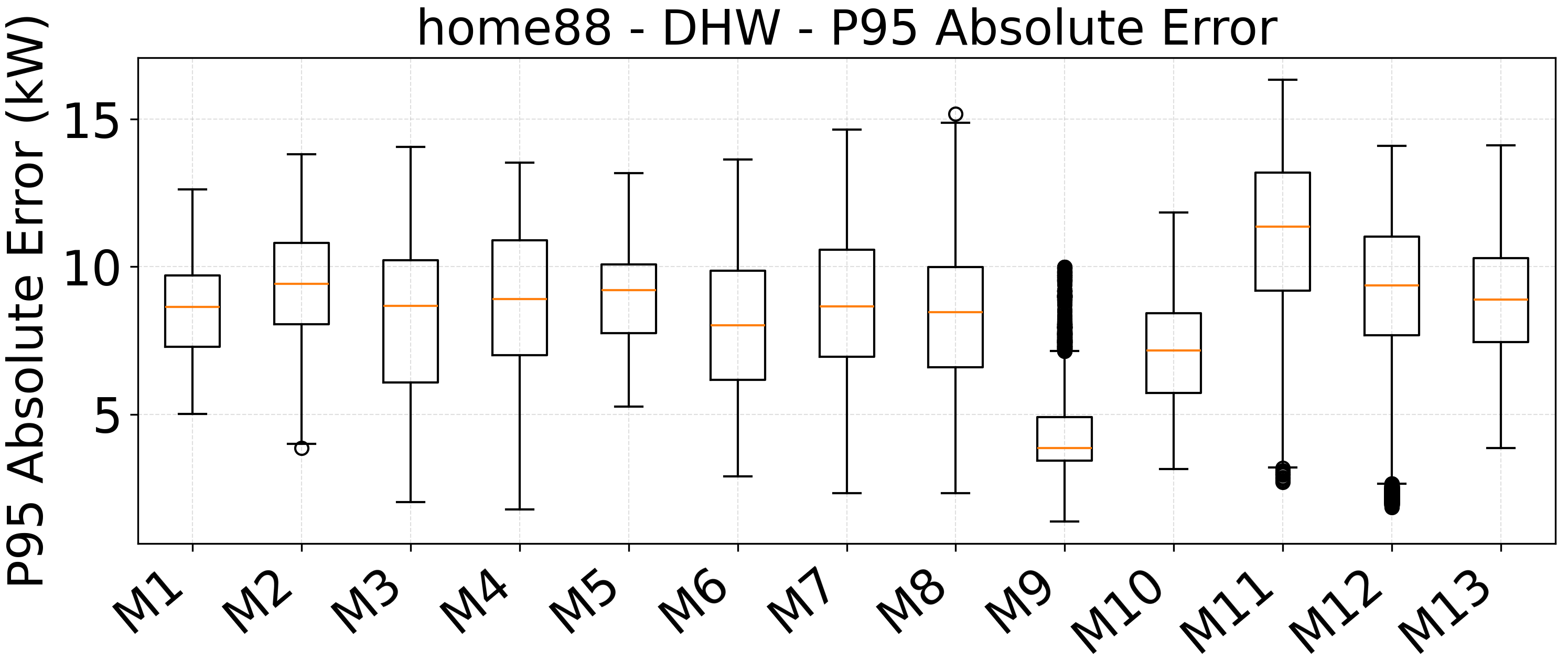}
        \label{fig:home88_P95_absolute_error_DHW}
    }\\
    \subfloat[]{
        \includegraphics[width=0.47\textwidth]{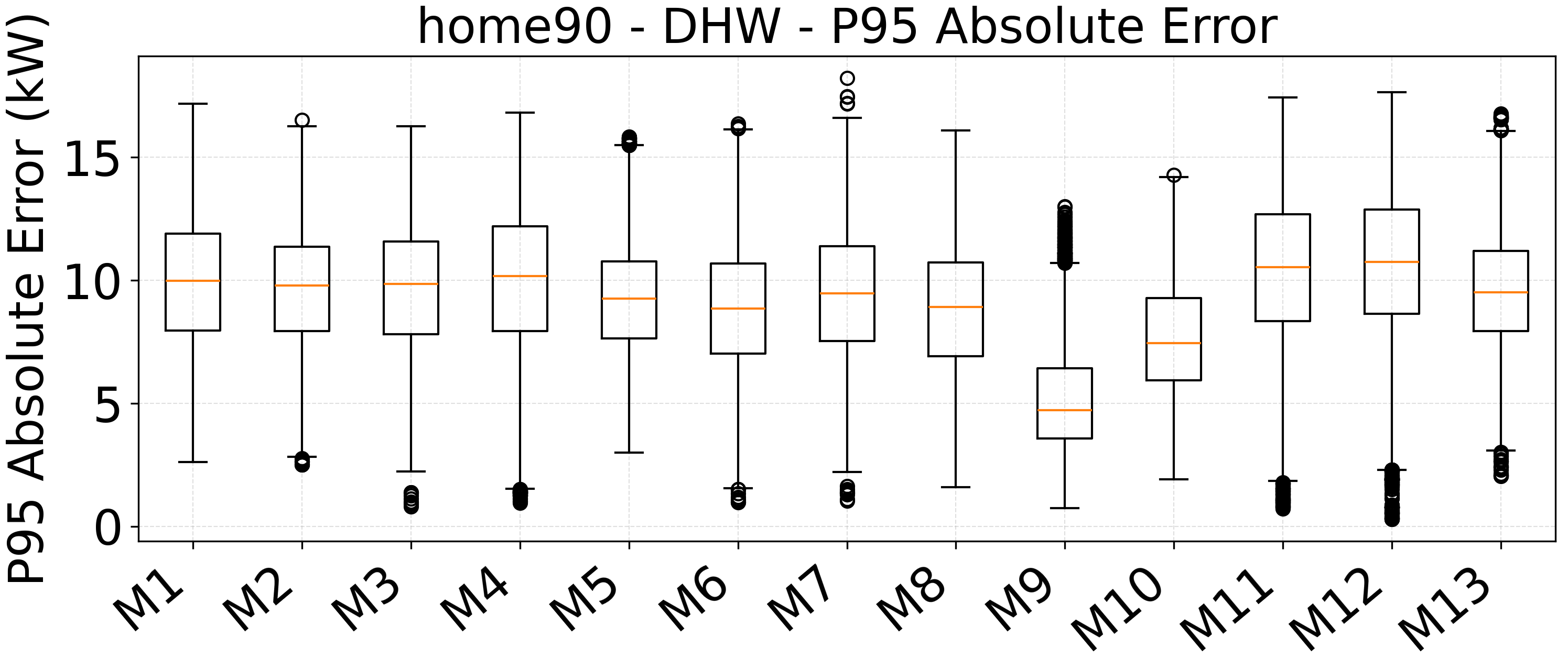}
        \label{fig:home90_P95_absolute_error_DHW}
    }
    \subfloat[]{
        \includegraphics[width=0.47\textwidth]{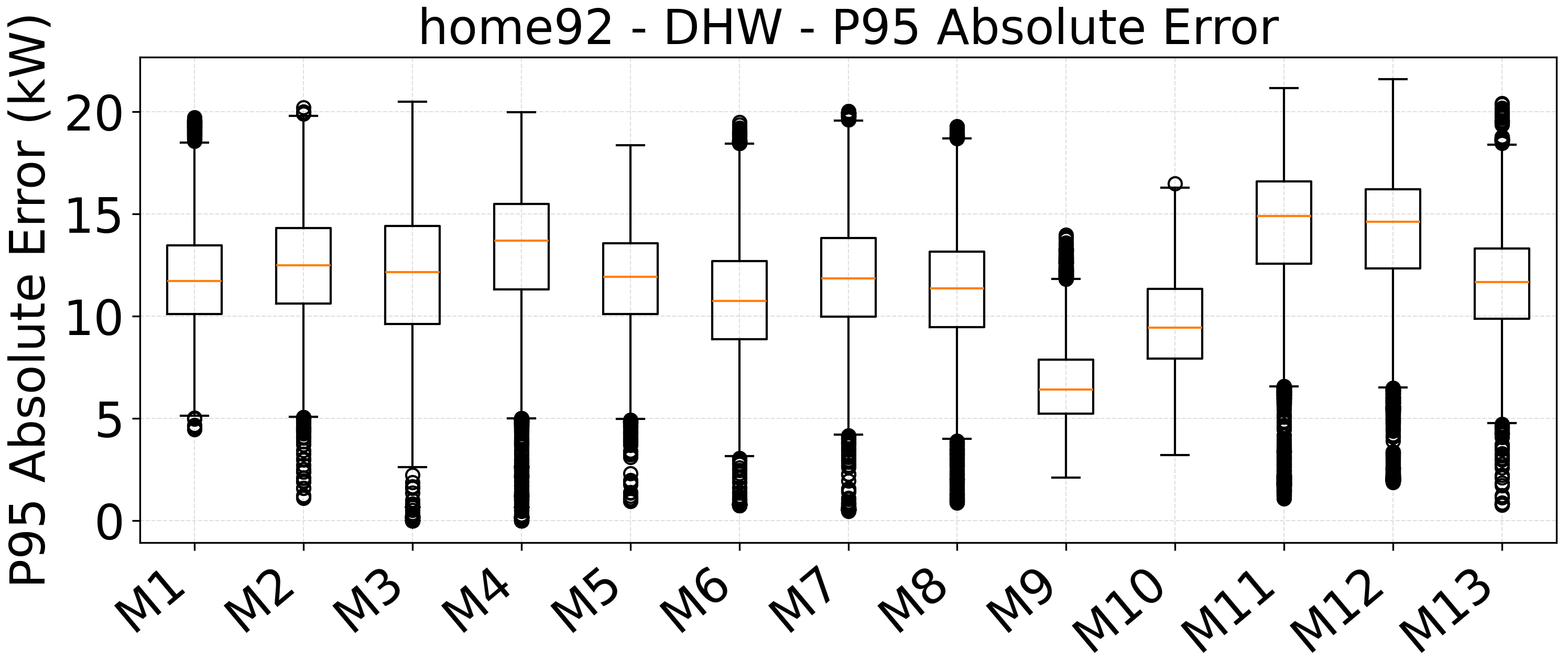}
        \label{fig:home92_P95_absolute_error_DHW}
    }\\
    \subfloat[]{
        \includegraphics[width=0.47\textwidth]{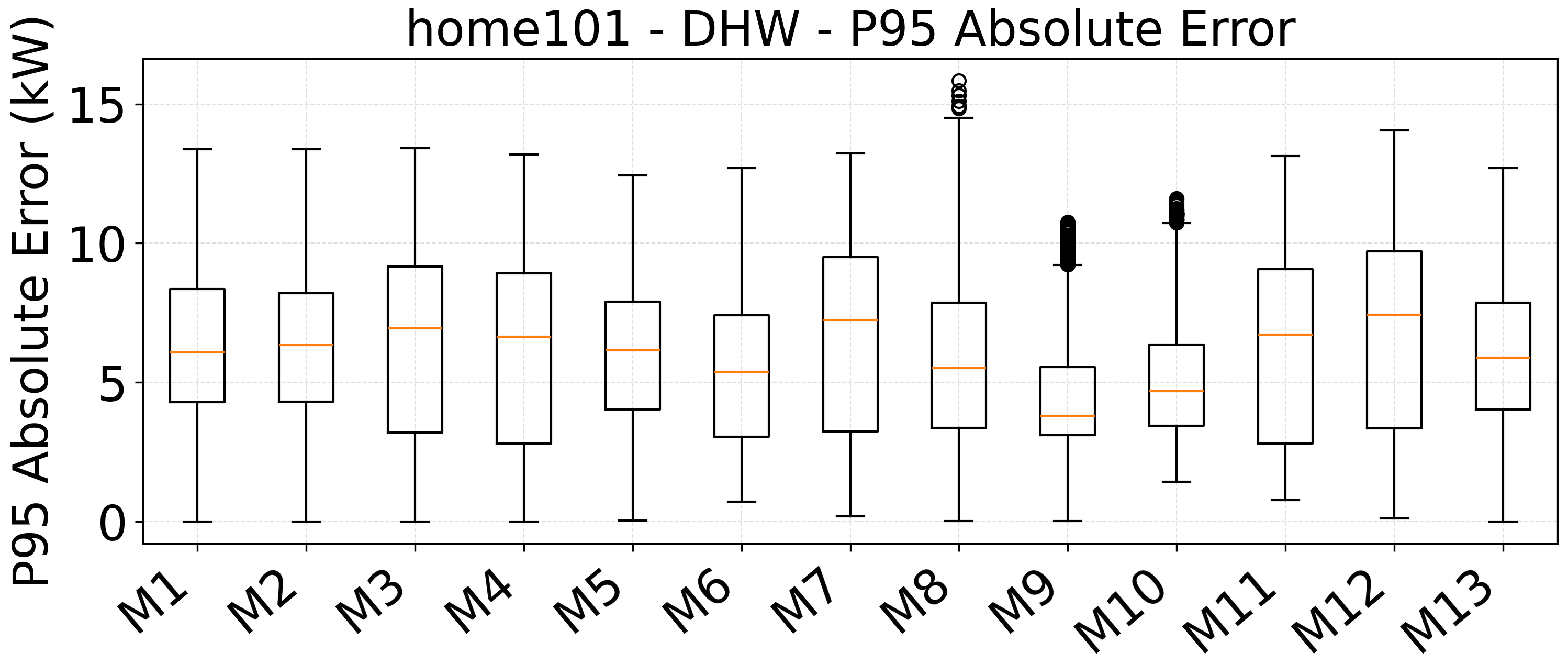}
        \label{fig:home101_P95_absolute_error_DHW}
    }
    \subfloat[]{
        \includegraphics[width=0.47\textwidth]{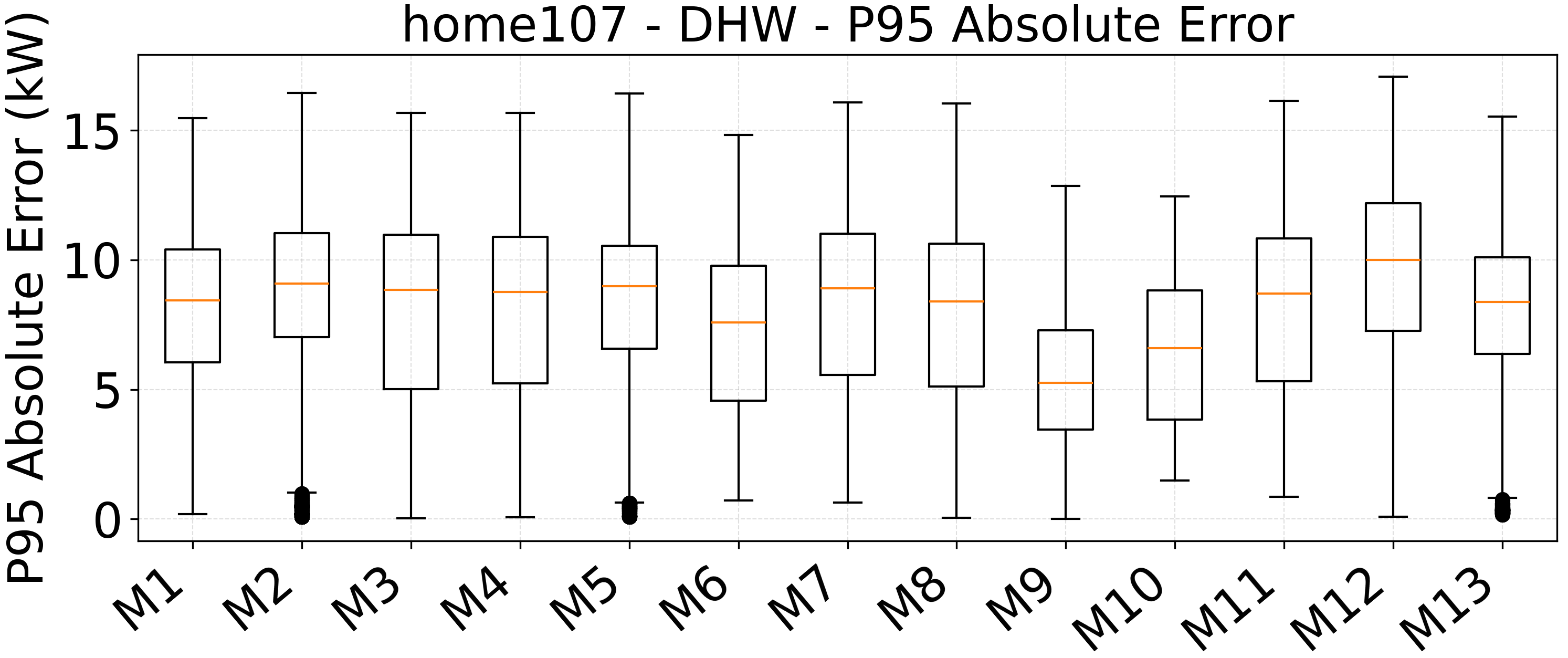}
        \label{fig:home107_P95_absolute_error_DHW}
    }\\
    \subfloat[]{
        \includegraphics[width=0.47\textwidth]{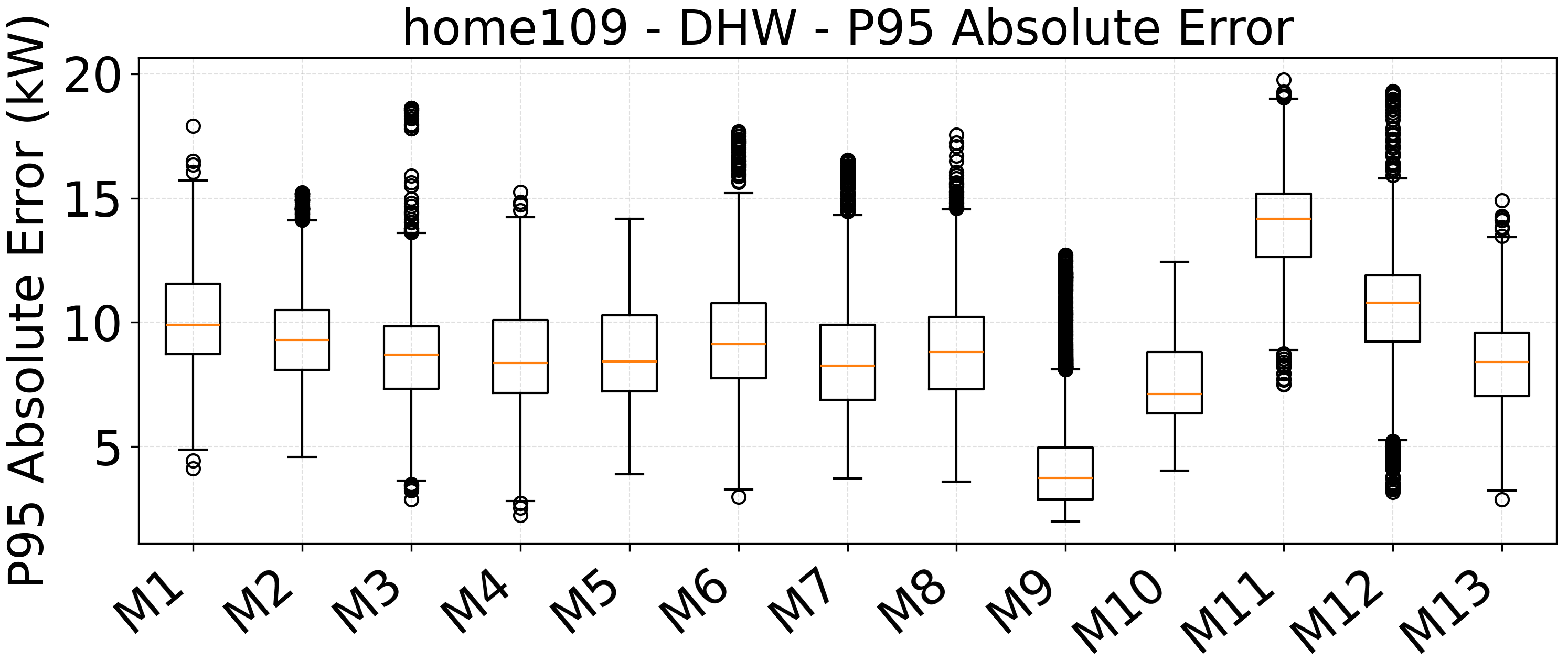}
        \label{fig:home109_P95_absolute_error_DHW}
    }
    \caption{\textrm{P95AE boxplots from the application of all DHW forecasting models at home: (a) 68, (b) 81, (c) 82, (d) 88, (e) 90, (f) 92, (g) 101, (h) 107, and (i) 109.}}
    \label{fig:P95_absolute_error_DHW_comparison}
\end{figure*}

\begin{figure*}[t]
    \centering
    \subfloat[]{
        \includegraphics[width=0.49\textwidth]{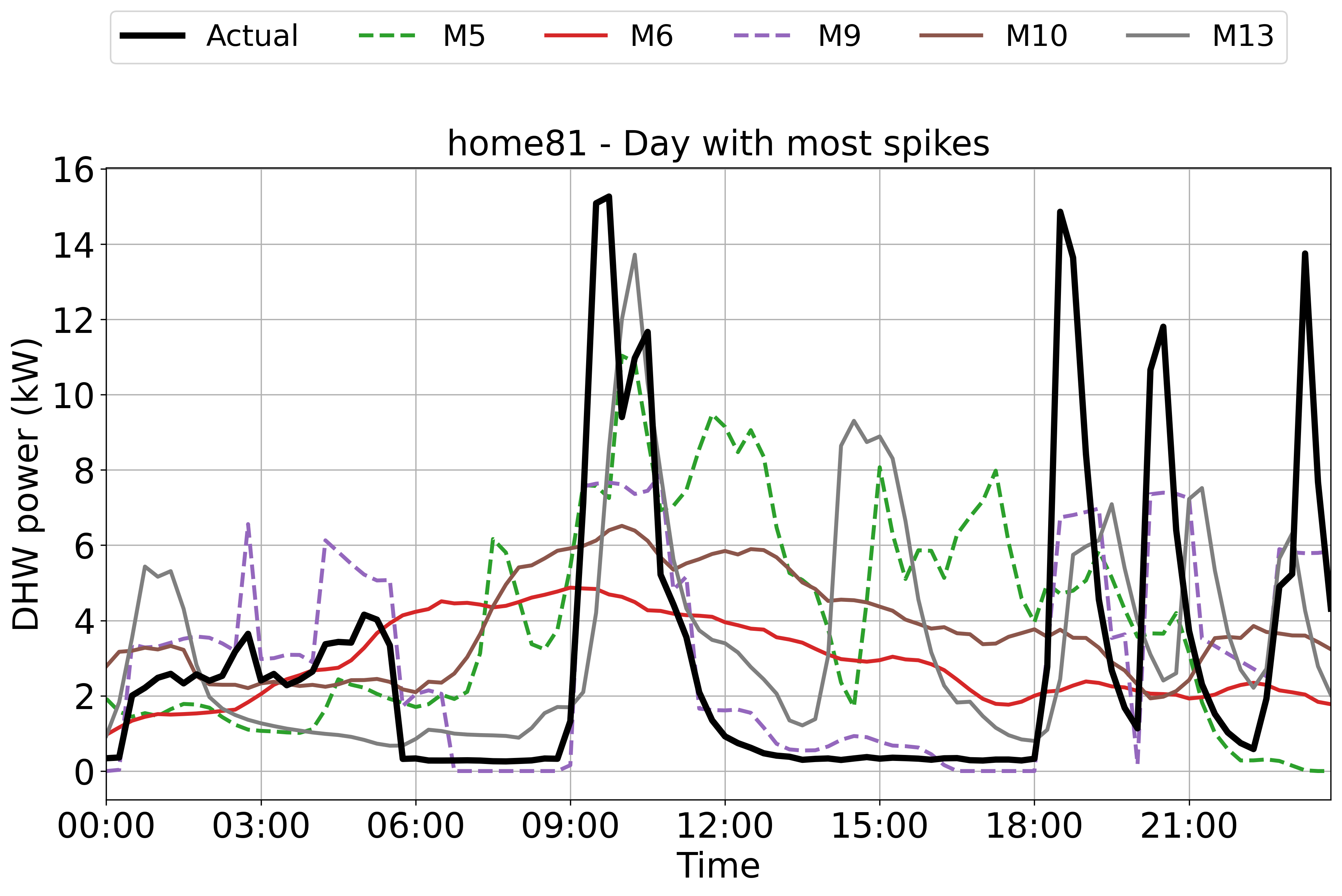}
        \label{fig:home81_Most_Spikes_DHW_Day}
    }
    \subfloat[]{
        \includegraphics[width=0.49\textwidth]{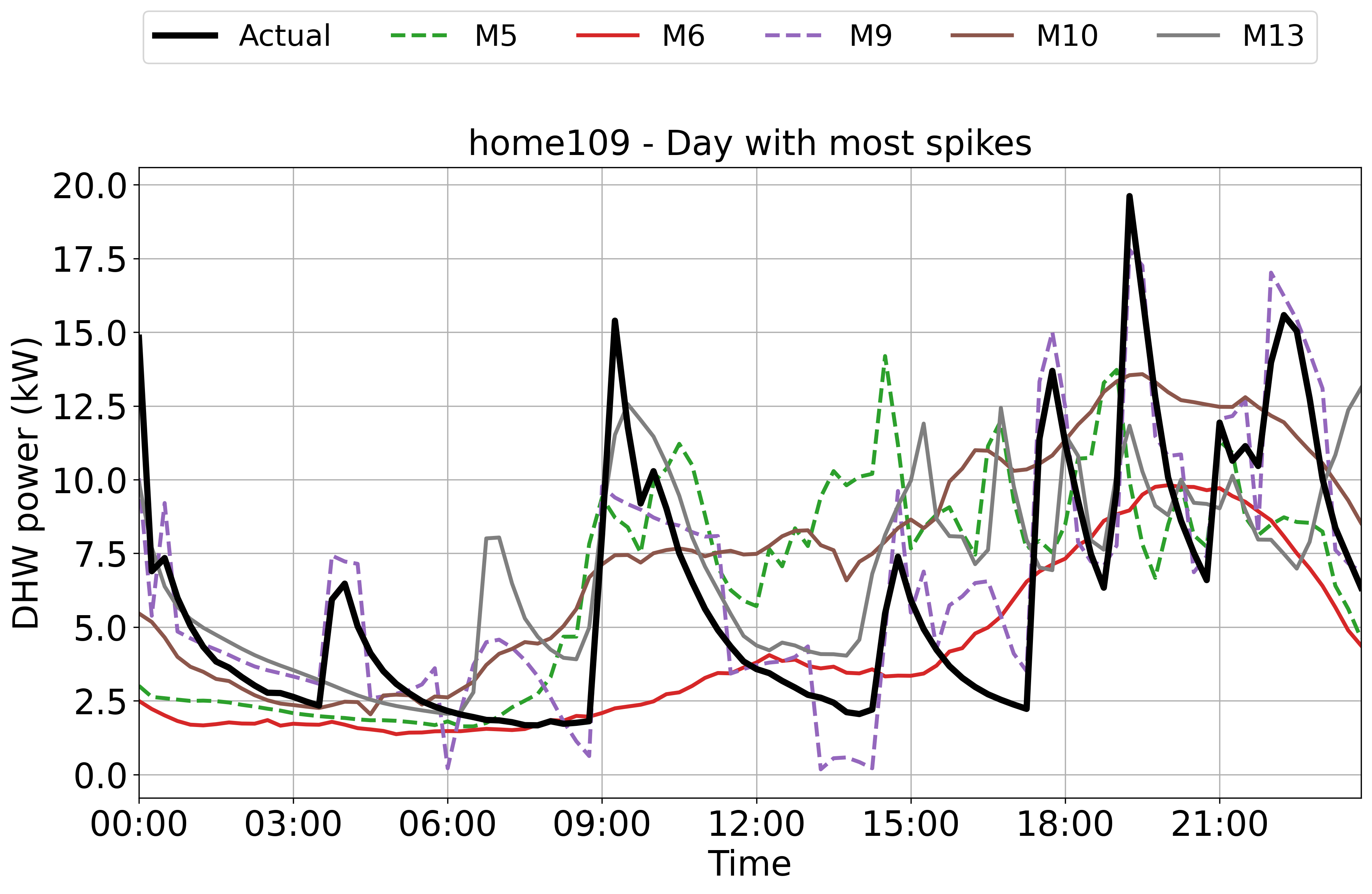}
        \label{fig:home109_Most_Spikes_DHW_Day}
    }\\
    \subfloat[]{
        \includegraphics[width=0.47\textwidth]{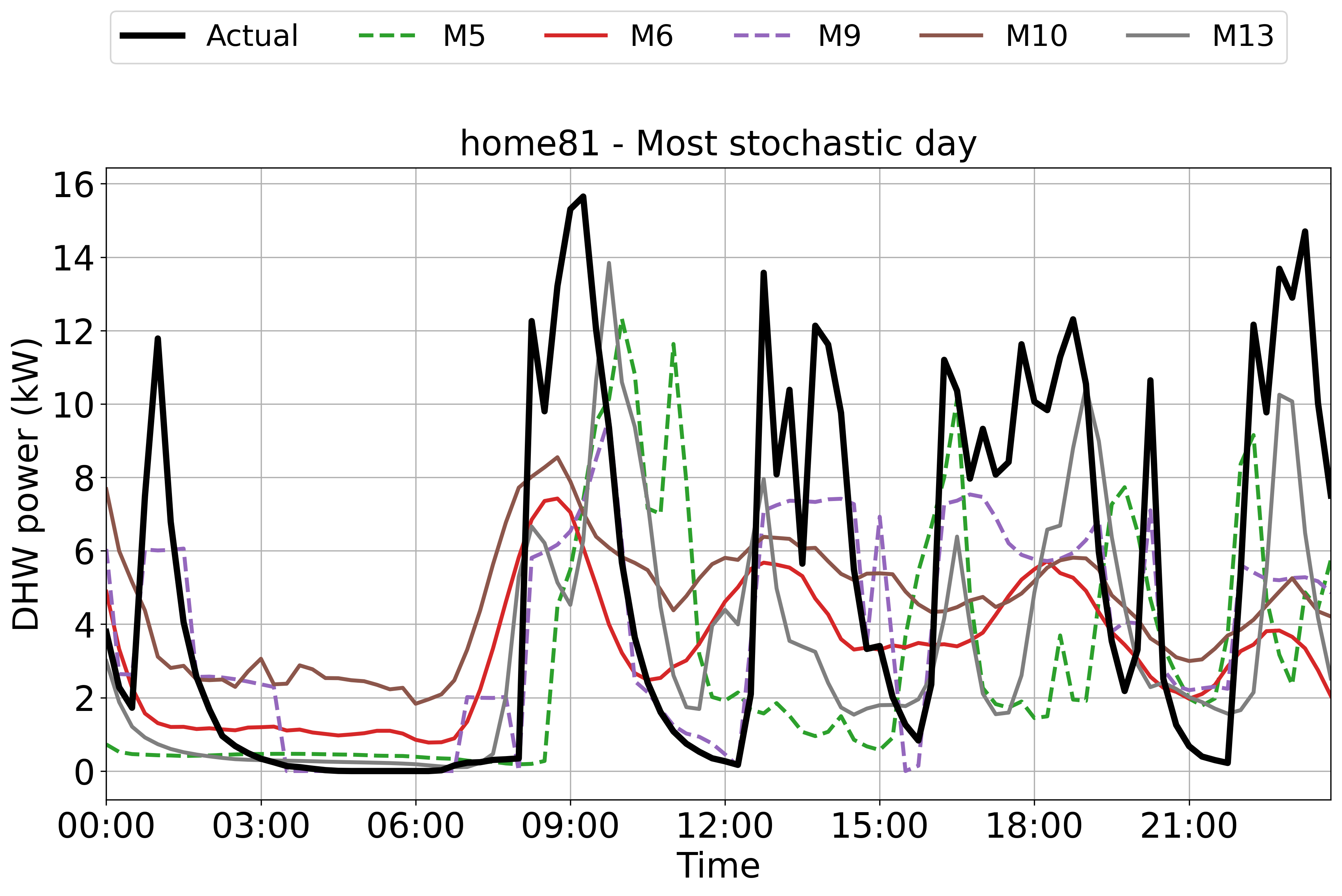}
        \label{fig:home81_Most_Stochastic_DHW_Day}
    }
    \subfloat[]{
        \includegraphics[width=0.47\textwidth]{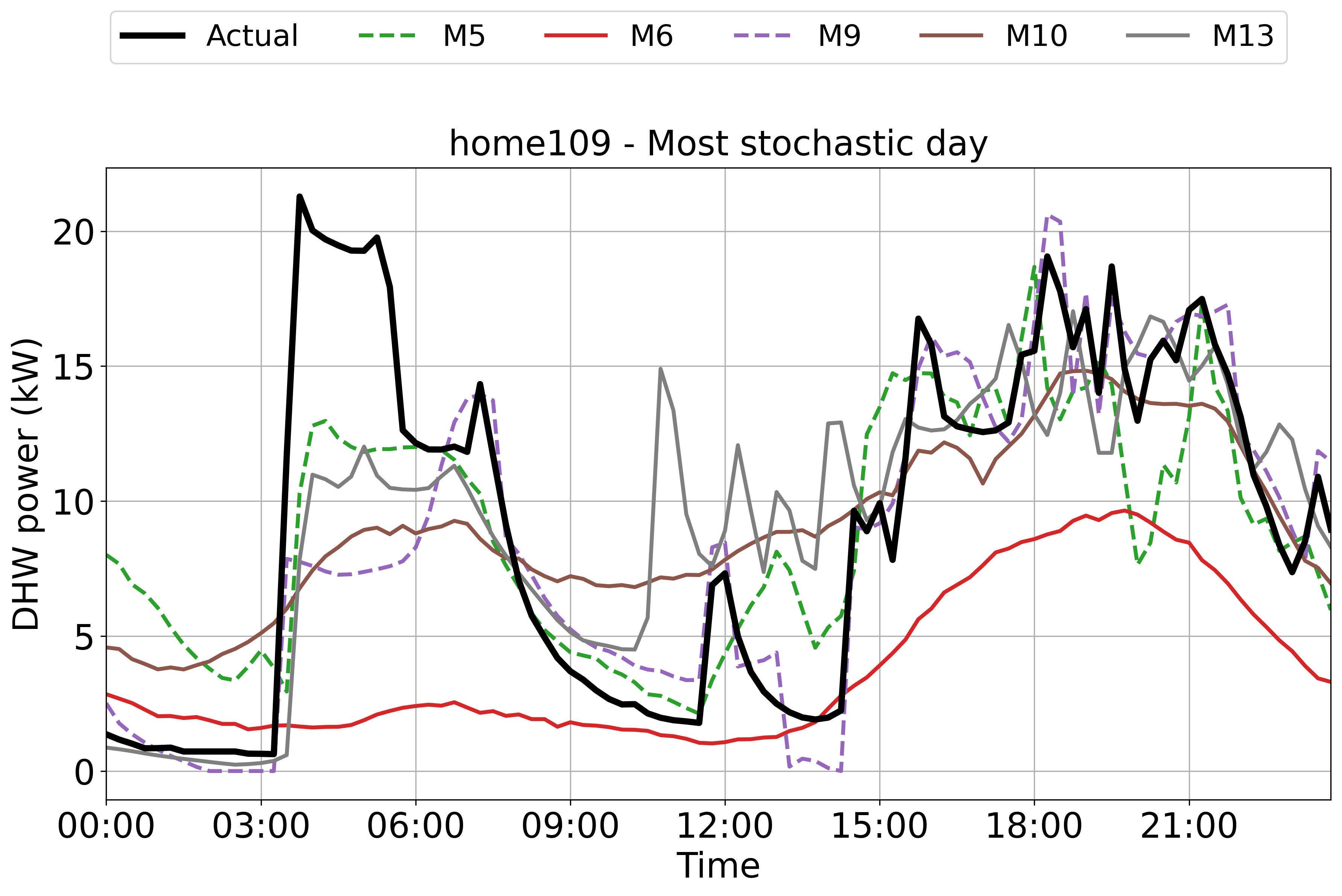}
        \label{fig:home109_Most_Stochastic_DHW_Day}
    }
    \caption{\textrm{Day with most DHW spikes for: (a) home 81 - 01/01/2018, and (b) home 109 - 21/04/2018, as well as most stochastic DHW day for: (c) home 81 - 22/12/2017, and (d) home 109 - 01/05/2018.}}
    \label{fig:Examples_DHW}
\end{figure*}

In terms of the metric P95AE, Fig.~\ref{fig:P95AE_heating_comparison} displays the distribution of P95AE boxplots of all methods for each home. As for the maximum values, it is clear that M9 outperforms the other methods for most buildings, while M10 and M1 present the lowest maximum P95AE values at three buildings. As for the 50\% P95AE-percentiles, the M9 also shows the highest performance at most buildings apart from home 101, where M6 surpasses the other buildings. The second lowest 50\% P95AE-percentiles are achieved by M10 at 7 buildings, as well as by M3 at households 101 and 109. Higher 50\% P95AE-percentiles are noticed for the models M5, M6 and M13.  

Next, the day-ahead heating demand forecasts are analyzed for two representative buildings. In Fig.~\ref{fig:Examples_Heating}, the day-ahead predictions at the beginning of the day, as generated by the models with the highest performance, are displayed for the days with the highest load and most rapid changes of the households 92 and 109. For the aforementioned buildings, the day with the highest heating demand coincides with the most rapidly changing day. As can be noticed in Fig.~\ref{fig:home92_Highest_Heating_Day}, the method M9 predicts accurately not only the load pattern, but also the local peaks. However, the high demand swells and peaks cannot be predicted by the selected hybrid models of home 92. Comparable performance on local peaks is also noticed for the model M5 with slightly higher deviations than M9. As shown in Fig.~\ref
{fig:home109_Highest_Heating_Day}, a similar trend is also observed for the household 109. Particularly, the method M9 predicts the load pattern, however, the majority of power peaks are underestimated. Besides that, the global daily peak is overestimated by M9. It can also be noticed that from the other methods the model M10 predicts partially the pattern, though the forecasts are conservative compared to the actual values. 

From the above analysis, it is clear that the model M9 outperforms the other hybrid methods when evaluating the boxplots of MAE, RMSE and P95AE. Moreover, M9 can predict accurately the load pattern, however, high demand swells and peaks are underestimated. In some cases, local peaks may be overestimated by M9. As for the other methods, the models M5 and M10 can also predict the pattern, though their forecasts mostly underestimate the actual values, especially with regard to the power peaks.  

\subsection{DHW demand forecasts}
\label{dhw_dem_for}

In the same manner to heating demand, the metrics MAE, RMSE and P95AE are calculated for the day-ahead DHW load forecasts at each timestep of the validation period. Next, for each home, the boxplots display the distribution of the metrics throughout the validation period for each model.

Fig.~\ref{fig:MAE_DHW_comparison} displays the MAE boxplots of the investigated hybrid methods for each building. In terms of the maximum MAE, it is evident that the M9 model achieves the lowest values including the outliers at all buildings. As for the 50\% MAE-percentiles, the model M9 outperforms the other methods at all households. The second lowest 50\% MAE-percentiles are achieved by M10 at most buildings, apart from the models M6 and M8 at households 82 and 101, respectively. Higher 50\% MAE-percentiles are achieved by the models M3, M4, M6 and M13. 

In Fig.~\ref{fig:RMSE_DHW_comparison}, the RMSE-boxplots of all methods are displayed for each home. Regarding the maximum RMSE, M9 achieves the lowest values at all households. Besides that, M9 outperforms the other methods with respect to the lowest 50\% RMSE-percentiles. The model M10 achieves the second lowest 50\% RMSE-percentiles at all buildings. Besides them, higher 50\% RMSE-percentiles are achieved by the models M6 and M13.

In terms of the P95AE-boxplots, Fig.~\ref{fig:P95_absolute_error_DHW_comparison} presents the distribution of P95AE errors of all methods for the day-ahead DHW load forecasts at each home. From the analysis, it is clear that the model M9 achieves the lowest maximum P95AE values at most buildings apart from the method M10 for the households 107 and 109. Regarding the lowest 50\% P95AE-percentiles, M9 outperforms the other methods at all households. The second lowest 50\% P95AE-percentiles are achieved by M10 at all buildings. Comparable performance with slightly higher 50\% P95AE-percentiles is observed for the models M5, M6 and M13. 

Taking into account the above results, the methods M1, M5, M6, M9, M10 and M13 are selected for further assessment. Particularly, the day-ahead DHW load forecasts of the selected methods are presented for the day with most spikes and the most stochastic day of the households 81 and 109. Since the day-ahead predictions are updated at each 15-min timestep, we evaluate the forecasts generated at the beginning of the examined day in Fig.~\ref{fig:Examples_DHW}.  

In Fig.~\ref{fig:home81_Most_Spikes_DHW_Day}, the day-ahead DHW predictions of home 81 are displayed for the day with most spikes. It is evident that the model M9 predicts the DHW pattern better than the other methods. However, the local power peaks are overestimated and the high swells and peaks are underestimated by M9. The model M13 also predicts the timestamp of various peaks, but underestimates their values.

In Fig.~\ref{fig:home109_Most_Spikes_DHW_Day}, the method M9 also achieves the highest performance, as the pattern and most power peaks are predicted with high accuracy. As for the other methods, the model M10 predicts the pattern with a smooth form, therefore, most spikes are not included in the forecast. 

As for the most stochastic days, Fig.~\ref{fig:home81_Most_Stochastic_DHW_Day} displays the day-ahead forecasts for the household 81. The power pattern is predicted by M9, however, all swells and hence, the peaks are underestimated. A similar behaviour is observed for the methods M6 and M10, which generate a smooth load curve underestimating the power peaks. In Fig.~\ref{fig:home109_Most_Stochastic_DHW_Day}, the model M9 predicts accurately not only the power pattern, but also all local peaks. However, the global daily DHW swell is underestimated. Regarding the other methods, M5 and M13 predict partially the load pattern and the local peaks, though the global swell is underestimated similarly to M9.

Overall, the method M9 outperforms the other forecasting models when comparing the boxplots of MAE, RMSE and P95AE. Furthermore, M9 predicts not only the power pattern but also the timestamp of both local and global peaks. Nevertheless, in several cases, the high power swells are underestimated, while local power peaks may be overestimated by M9. Concerning the other methods, the models M10 and M13 predict the pattern with a smooth form, therefore, many spikes are underestimated. 

\subsection{Discussion}
\label{tech}

Due to the various feature selection methods applied to the examined forecasting models, the selected input features are not always identical. The selection of different input features also leads to deviations in the calculated metrics among the methods. Though the selection of the same input features could set a benchmark framework for the comparison of the hybrid methods, the inputs are selected by the feature selection method included in each model. 

The weather forecasts are provided by a specific weather station of Meteonorm in Edinburgh, while the households are located in the metropolitan region of Edinburgh. Also, missing weather forecasts were filled according to the empirical distribution of existing weather forecasts. These assumptions can further deteriorate the performance of the examined forecasting models. 

As for the assessment criteria, multiple additional metrics can be calculated, however, from the selected ones, it is clear that the model EMD-LSTM-Markov outperforms the other hybrid methods in most cases. Other metrics, such as MAPE, and sMAPE have also been computed, nevertheless, due to the actual zero heating and DHW power demands, the extreme computed values did not allow the extraction of clear outcomes as with the MAE, RMSE and P95AE. 

Most of the examined forecasting models are evaluated in the literature for a shorter prediction horizon than one day. Therefore, the selected prediction horizon of 96-steps is challenging for the evaluated hybrid methods and can lead to higher accuracy errors compared to the literature. It should be highlighted that the goal of this paper is to assess the methods considering the same requirements. The common prediction horizon of 96-steps is a prerequisite for the examined forecasting models, which can be considered in the BEMS operation, in the context of a future work.

\section{Conclusions}
\label{sum}

In this paper, the comprehensive comparison of hybrid forecasting models includes combinations of black-box methods, data-driven methods combined with Metaheuristics or enhanced with noise reduction. From the analysis of 13 hybrid methods, it is concluded that the inputs can include multiple weather factors, typical calendar parameters, historical outputs, as well as features related to the building and HVAC system. The output features include various temperatures and demand variables. Furthermore, the advantages and disadvantages of each method are highlighted, however, no clear outperformance is concluded at this step. 

At the next stage, the hybrid methods are evaluated using a real energy dataset from Scottish households considering the generated day-ahead forecasts. First, the feature selection method of each model is applied and the input features with the highest correlation are selected. It is concluded that the historical outputs of the previous day and previous week, as well as the temperature parameters have the highest correlation with the actual heating and DHW demand. 

In the context of evaluating the day-ahead forecasts, the EMD-LSTM-Markov outperforms the other methods, when comparing the boxplots of MAE, RMSE and P95AE. Besides that, it is clear that the EMD-LSTM-Markov can predict more accurately the patterns of heating and DHW demands than the other models. Moreover, the EMD-LSTM-Markov can predict the local power peaks more accurately than the high demand swells and global peaks, which are underestimated. As for the other methods, the SVM-SA clearly achieves the second highest performance predicting the heating and DHW power patterns with a smooth form. Consequently, most local and global power peaks are underestimated. From the whole analysis, it is concluded that the EMD-LSTM-Markov can be considered for the prediction of both heating and DHW demands in the BEMS operation.

\section*{Declaration of competing interest}

The authors declare that they have no known competing financial interests or personal relationships that could have appeared to influence the work reported in this paper.

\section*{Acknowledgment}

The authors would like to thank the weather data company Meteonorm for the provision of historical weather forecasts, as well as the School of Informatics from the University of Edinburgh for the open-source database of residential thermal energy demand. Without the aforementioned contributions, the additional assessment of the preselected hybrid approaches would not be feasible. 

\begin{appendices}

\section{Model parameters}
\label{ap:Hyperpar}

\begin{table}[h]
\begin{center}
\caption{\textrm{PatchTCN-TST parameters}}
\label{tab:PatchTCN-TSTparams}
\begin{tabular*}{\columnwidth}{@{}LL@{}}
\toprule
\textrm{Parameter} & \textrm{Value} \\
\midrule
\textrm{Patch length} & $8$ \\
\textrm{Number of patches} & $12$ \\
\textrm{Transformer dimension} & $128$ \\
\textrm{Attention heads} & $4$ \\
\textrm{Batch size} & $8$ \\
\textrm{Training epochs} & $40$ \\
\textrm{TCN layers} & $3$ \\
\textrm{Transformer encoder layers} & $3$ \\
\textrm{GRU temporal decoder layers} & $2$ \\
\textrm{GRU hidden size} & $128$ \\
\textrm{Feedforward dimension} & $256$ \\
\textrm{Dropout rate} & $0.1$ \\
\textrm{Activation function} & \textrm{GELU} \\
\textrm{Learning rate} & $3\times10^{-4}$ \\
\textrm{Weight decay} & $10^{-4}$ \\
\textrm{Optimizer} & \textrm{AdamW} \\
\bottomrule
\end{tabular*}
\end{center}
\end{table}

\begin{table}[h]
\begin{center}
\caption{\textrm{DwdAdam-ILSTM parameters}}
\label{tab:DwdAdam-ILSTMparams}
\begin{tabular*}{\columnwidth}{@{}LL@{}}
\toprule
\textrm{Parameter} & \textrm{Value} \\
\midrule
\textrm{LSTM batch size} & $8$ \\
\textrm{LSTM hidden layers} & $1$ \\
\textrm{LSTM hidden units} & $64$ \\
\textrm{Training epochs} & $25$ \\
\textrm{Learning rate} & $10^{-3}$ \\
\textrm{Activation function} & \textrm{Softplus} \\
\textrm{Optimizer} & \textrm{Adam} \\
\textrm{Weight decay} & $0$ \\
\textrm{Gradient clipping} & $1.0$ \\
\bottomrule
\end{tabular*}
\end{center}
\end{table}

\begin{table}[h]
\begin{center}
\caption{\textrm{CLSTM-AR forecasting model parameters}}
\label{tab:CLSTMARparams}
\begin{tabular*}{\columnwidth}{@{}LL@{}}
\toprule
\textrm{Parameter} & \textrm{Value} \\
\midrule
\textrm{Batch size} & $32$ \\
\textrm{Training epochs} & $30$ \\
\textrm{Learning rate} & $10^{-3}$ \\
\textrm{Encoder input size} & \textrm{Selected features} \\
\textrm{Encoder hidden size} & $64$ \\
\textrm{Encoder layers} & $1$ \\
\textrm{Decoder input size} & $1$ \\
\textrm{Decoder hidden size} & $64$ \\
\textrm{Decoder layers} & $1$ \\
\textrm{Output head [inputs, outputs]} & $[128,64]$ \\
\textrm{Final output layer} & $[64,1]$ \\
\textrm{Teacher forcing} & \textrm{Scheduled sampling} \\
\textrm{Output activation} & \textrm{Softplus} \\
\textrm{Optimizer} & \textrm{Adam} \\
\bottomrule
\end{tabular*}
\end{center}
\end{table}

\begin{table}[h]
\begin{center}
\caption{\textrm{PCC-GRU parameters}}
\label{tab:PCC-GRUparams}
\begin{tabular*}{\columnwidth}{@{} LL@{} }
\toprule
\textrm{Parameter} & \textrm{Value} \\
\midrule
\textrm{Training batch size} & $32$ \\
\textrm{Training epochs} & $8$ \\
\textrm{GRU layers} & $1$ \\
\textrm{GRU hidden units} & $64$ \\
\textrm{Output dimension} & $96$ \\
\textrm{Activation function} & \textrm{Softplus()} \\
\textrm{Optimizer} & \textrm{Adam} \\
\textrm{Learning rate} & $0.005$ \\
\textrm{Target scaling factor} & $100$ \\
\bottomrule
\end{tabular*}
\end{center}
\end{table}

\begin{table}[h]
\begin{center}
\caption{\textrm{FTTrans-E-BL parameters}}
\label{tab:FTTrans-E-BLparams}
\begin{tabular*}{\columnwidth}{@{} LL@{} }
\toprule
\textrm{Parameter} & \textrm{Value} \\
\midrule
\textrm{Training epochs} & $30$ \\
\textrm{Training batch size} & $32$ \\
\textrm{Learning rate} & $5\times10^{-4}$ \\
\textrm{Input feature embedding dimension} & $64$ \\
\textrm{Transformer encoder layers} & $2$ \\
\textrm{Transformer attention heads} & $4$ \\
\textrm{Transformer model dimension} & $64$ \\
\textrm{Temporal decoder type} & \textrm{LSTM} \\
\textrm{Temporal decoder layers} & $2$ \\
\textrm{Temporal decoder hidden size} & $64$ \\
\textrm{Output network hidden size} & $64$ \\
\textrm{Activation function} & \textrm{ReLU()} \\
\textrm{Gradient clipping} & $1.0$ \\
\textrm{Optimizer} & \textrm{Adam} \\
\textrm{Output constraint} & \textrm{Non-negative clipping} \\
\bottomrule
\end{tabular*}
\end{center}
\end{table}

\begin{table}[h]
\begin{center}
\caption{\textrm{TE-BiSRU-MTL parameters}}
\label{tab:FTA-BiGRUparams}
\begin{tabular*}{\columnwidth}{@{} LL@{} }
\toprule
\textrm{Parameter} & \textrm{Value} \\
\midrule
\textrm{Copula entropy bins} & $20$ \\
\textrm{Relative entropy tolerance} & $0.15$ \\
\textrm{Bi-GRU hidden size} & $128$ \\
\textrm{Bi-GRU output size} & $256$ \\
\textrm{Training epochs} & $20$ \\
\textrm{Training batch size} & $32$ \\
\textrm{Learning rate} & $10^{-3}$ \\
\textrm{Optimizer} & \textrm{Adam} \\
\textrm{Output activation at inference} & \textrm{Softplus} \\
\textrm{Numerical stability constant}($\epsilon$) &
$10^{-3}$ \\
\bottomrule
\end{tabular*}
\end{center}
\end{table}

\begin{table}[h]
\begin{center}
\caption{\textrm{WTD-CNN-LSTM parameters}}
\label{tab:WTD-CNN-LSTMparams}
\begin{tabular*}{\columnwidth}{@{} LL@{} }
\toprule
\textrm{Parameter} & \textrm{Value} \\
\midrule
\textrm{Training epochs} & $100$ \\
\textrm{Training batch size} & $32$ \\
\textrm{Learning rate} & $10^{-3}$ \\
\textrm{Wavelet family} & \textrm{bior2.8} \\
\textrm{WTD decomposition level} & $3$ \\
\textrm{CNN layers} & $2$ \\
\textrm{CNN filters} & $128/128$ \\
\textrm{CNN kernel size} & $3$ \\
\textrm{CNN activation} & \textrm{ReLU()} \\
\textrm{LSTM layers} & $1$ \\
\textrm{LSTM hidden size} & $64$ \\
\textrm{Output activation} & \textrm{Softplus()} \\
\textrm{Optimizer} & \textrm{Adam} \\
\bottomrule
\end{tabular*}
\end{center}
\end{table}

\begin{table}[h]
\begin{center}
\caption{\textrm{aTCN-LSTM parameters}}
\label{tab:aTCN-LSTMparams}
\begin{tabular*}{\columnwidth}{@{}LL@{}}
\toprule
\textrm{Parameter} & \textrm{Value} \\
\midrule
\textrm{Correlation subsample} & $200$ \\
\textrm{Training batch size} & $32$ \\
\textrm{Inference batch size} & $16$ \\
\textrm{Training epochs} & $40$ \\
\textrm{Early stopping patience} & $6$ \\
\textrm{Learning rate} & $10^{-3}$ \\
\textrm{Optimizer} & \textrm{Adam} \\
\textrm{CEEMDAN downsample factor} & $4$ \\
\textrm{CEEMDAN noise realizations} & $20$ \\
\textrm{CEEMDAN maximum IMFs} & $6$ \\
\textrm{Sample entropy threshold} & $1$ \\
\textrm{Wavelet basis} & \textrm{bior2.8} \\
\textrm{WTD decomposition level} & $3$ \\
\textrm{GMTCN blocks} & $3$ parallel blocks \\
\textrm{GMTCN kernel sizes} & $[3,\;5,\;7]$ \\
\textrm{GMTCN dilation factors} & $[1,\;2,\;4]$ \\
\textrm{GMTCN output channels} & $64$ \\
\textrm{BiLSTM input size} & $64$ \\
\textrm{BiLSTM hidden size} & $64$ \\
\textrm{Attention embedding dimension} & $128$ \\
\textrm{Attention heads} & $2$ \\
\textrm{Output layer size} & $96$ \\
\bottomrule
\end{tabular*}
\end{center}
\end{table}

\begin{table}[h]
\begin{center}
\caption{\textrm{EMD-LSTM-Markov parameters}}
\label{tab:EMD-LSTM-Markovparams}
\begin{tabular*}{\columnwidth}{@{} LL@{} }
\toprule
\textrm{Parameter} & \textrm{Value} \\
\midrule
\textrm{Training epochs} & $30$ \\
\textrm{Learning rate} & $10^{-3}$ \\
\textrm{Training batch size} & $32$ \\
\textrm{Inference batch size} & $32$ \\
\textrm{Optimizer} & \textrm{Adam} \\
\textrm{Teacher forcing ratio} & $0.5$ \\
\textrm{Encoder architecture} & \textrm{Seq2Seq LSTM} \\
\textrm{Decoder architecture} & \textrm{Autoregressive LSTM} \\
\textrm{Number of LSTM layers} & $2$ \\
\textrm{LSTM hidden units} & $128$ \\
\textrm{LSTM activation functions} & \textrm{tanh, sigmoid}\\
\bottomrule
\end{tabular*}
\end{center}
\end{table}

\begin{table}[h]
\begin{center}
\caption{\textrm{SVM-SA parameters}}
\label{tab:SVM-SAparams}
\begin{tabular*}{\columnwidth}{@{} LL@{} }
\toprule
\textrm{Parameter} & \textrm{Value} \\
\midrule
\textrm{SA cycles} & $2$ \\
\textrm{SA trials per cycle} & $3$ \\
\textrm{SA subsample} & $1500$ \\
\textrm{LinearSVR maximum iterations} & $5000$ \\
\textrm{LinearSVR tolerance} & $10^{-3}$ \\
\textrm{Search space - $C$ range} & $[0.01, 10]$ \\
\textrm{Search space - $\epsilon$ range} & $[0.001, 0.2]$ \\
\textrm{Random seed} & $42$ \\
\bottomrule
\end{tabular*}
\end{center}
\end{table}

\begin{table}[h]
\begin{center}
\caption{\textrm{GA/iPSO-ANN parameters}}
\label{tab:iPSOANNparams}
\begin{tabular*}{\columnwidth}{@{} LL@{} }
\toprule
\textrm{Parameter} & \textrm{Value} \\
\midrule
\textrm{PCA number of components} & $150$ \\
\textrm{ANN hidden layers} & $1$ \\
\textrm{ANN hidden neurons} & $20$ \\
\textrm{ANN hidden activation function} & \textrm{ReLU} \\
\textrm{ANN output activation function} & \textrm{Softplus} \\
\textrm{ANN output layer} & $96$ \textrm{neurons} \\
\textrm{iPSO particles} & $30$ \\
\textrm{iPSO iterations} & $40$ \\
\textrm{iPSO inertia weight ($w$)} & $0.7$ \\
\textrm{iPSO cognitive coefficient ($c_1$)} & $1.5$ \\
\textrm{iPSO social coefficient ($c_2$)} & $1.5$ \\
\textrm{iPSO crossover rate} & $0.4$ \\
\textrm{iPSO mutation rate} & $0.1$ \\
\textrm{Time-series CV folds} & $3$ \\
\textrm{Minimum training rows} & $10$ \\
\textrm{Maximum weight value} & $5.0$ \\
\textrm{Maximum velocity value} & $1.0$ \\
\textrm{Maximum activation value} & $50.0$ \\
\bottomrule
\end{tabular*}
\end{center}
\end{table}

\begin{table}[h]
\begin{center}
\caption{\textrm{SSA-CNN-SVM parameters}}
\label{tab:SSA-CNN-SVMparams}
\begin{tabular*}{\columnwidth}{@{} LL@{} }
\toprule
\textrm{Parameter} & \textrm{Value} \\
\midrule
\textrm{Conv1 filters} & $32$ \\
\textrm{Conv2 filters} & $32$ \\
\textrm{Parallel convolution kernels} & \textrm{[3, 7]} \\
\textrm{Residual connection} & \textrm{Enabled} \\
\textrm{Dropout rate} & $0.2$ \\
\textrm{Adaptive pooling output length} & $24$ \\
\textrm{Deep feature dimension} & $128$ \\
\textrm{Optimizer} & \textrm{Adam} \\
\textrm{Learning rate} & $10^{-3}$ \\
\textrm{Training epochs} & $5$ \\
\textrm{Batch size} & $128$ \\
\bottomrule
\end{tabular*}
\end{center}
\end{table}

\begin{table}[h]
\begin{center}
\caption{\textrm{RF-ISSA-LSTM parameters}}
\label{tab:RF-ISSA-LSTMparams}
\begin{tabular*}{\columnwidth}{@{}LL@{}}
\toprule
\textrm{Parameter} & \textrm{Value} \\
\midrule
\textrm{ISSA population size} & $4$ \\
\textrm{ISSA candidates evaluated} & $4$ \\
\textrm{LSTM batch size} & $128$ \\
\textrm{Evaluation batch size} & $512$ \\
\textrm{Early stopping patience} & $8$ \\
\textrm{RF trees} & $120$ \\
\textrm{RF maximum depth} & $14$ \\
\textrm{Encoder architecture} &
\textrm{LSTM, 2 layers, dropout $=0.2$} \\
\textrm{Decoder architecture} &
\textrm{LSTM, 2 layers, dropout $=0.2$} \\
\textrm{LSTM hidden units} & $[64,\;192]$ \\
\textrm{LSTM epochs} & $[20,\;45]$ \\
\textrm{Learning rate} & $[0.0005,\;0.003]$ \\
\textrm{L2 regularization} & $[10^{-7},\;10^{-4}]$ \\
\textrm{Optimizer} & \textrm{AdamW} \\
\textrm{Scheduler factor} & $0.5$ \\
\textrm{Scheduler patience} & $3$ \\
\textrm{Teacher forcing ratio} & $0.70 \rightarrow 0.15$ \textrm{(scheduled decay)} \\
\textrm{Timestep embedding size} & $16$ \\
\textrm{Normalization layer} & \textrm{LayerNorm} \\
\textrm{Activation function} & \textrm{ReLU()} \\
\textrm{Gradient clipping} & $1.0$ \\
\bottomrule
\end{tabular*}
\end{center}
\end{table}

\end{appendices}

\printcredits

\bibliographystyle{unsrt}


\bibliography{example.bib}


\end{document}